%% file: 0arxiv.tex
\documentclass[conference,compsoc]{IEEEtran}
\IEEEoverridecommandlockouts
\usepackage{cite}
\usepackage{amsmath,amssymb,amsfonts}
\usepackage{algorithmic}
\usepackage{graphicx}
\usepackage{textcomp}
\usepackage{xcolor}
\usepackage{indentfirst}
\usepackage{tikz}
\usepackage{multirow}
\usepackage{listings}
\usepackage{lipsum} 
\usepackage{orcidlink}
\usepackage{booktabs}

\usepackage{filecontents}

\def\BibTeX{{\rm B\kern-.05em{\sc i\kern-.025em b}\kern-.08em
    T\kern-.1667em\lower.7ex\hbox{E}\kern-.125emX}}

\input{00macros}

\includeversion{arxiv}
\excludeversion{submission}

\begin{document}


\title{Do Concept Replacement Techniques Really Erase Unacceptable Concepts?}

\begin{arxiv}
\author{\IEEEauthorblockN{Anudeep Das}
\IEEEauthorblockA{\textit{University of Waterloo} \\
Waterloo, Canada \\
a38das@uwaterloo.ca}
\and
\IEEEauthorblockN{Gurjot Singh}
\IEEEauthorblockA{\textit{University of Waterloo} \\
Waterloo, Canada \\
g86singh@uwaterloo.ca}
\and
\IEEEauthorblockN{Prach Chantasantitam}
\IEEEauthorblockA{\textit{University of Waterloo} \\
Waterloo, Canada \\
pchantas@uwaterloo.ca}
\and
\IEEEauthorblockN{N. Asokan}
\IEEEauthorblockA{\textit{University of Waterloo} \\
Waterloo, Canada \\
asokan@acm.org}
}
\end{arxiv}

\maketitle
\pagestyle{plain}

\input{0abstract}
\input{01introduction}

\input{02background}
\input{03approach_recon}
\input{04defense_approach}
\input{05results_defense}

\input{07related_work}
\input{08future_work}
\input{09conclusion}
\begin{arxiv}
\input{10acks}
\end{arxiv}

\scriptsize
\bibliographystyle{plain}
\bibliography{jobnam}

\begin{arxiv}
\input{09appendix}
\end{arxiv}

\end{document}

%% file: 00macros.tex
\usepackage{tikz}
\usetikzlibrary{shapes,arrows}
\usetikzlibrary{backgrounds, positioning, fit}
\usepackage{dblfloatfix}
\usepackage{xurl}
\usepackage[inline]{enumitem}
\usepackage{graphics}
\usepackage[T1]{fontenc}
\usepackage{xspace}
\usepackage{multirow}
\usepackage[table]{colortbl}
\usepackage{csquotes}
\usepackage{versions}
\usepackage{pifont}
\usepackage[normalem]{ulem}
\usepackage[font=normalsize,justification=justified,labelsep=colon]{caption}

\usepackage{amssymb,amsmath,amsthm}
\newcommand{\dm}{$f$\xspace}

\newcommand{\dmcrt}{$f^{CRT}$\xspace}
\newcommand{\unet}{$\epsilon_\theta$\xspace}

\newcommand{\tti}{T2I\xspace}
\newcommand{\iti}{I2I\xspace}

\newcommand{\crt}{CRT\xspace}

\newcommand{\imgclean}{\texttt{$x$}\xspace}
\newcommand{\xin}{\texttt{$x_{in}$}\xspace}
\newcommand{\xxout}{\texttt{$x_{out}$}\xspace}
\newcommand{\xface}{\texttt{$x_{out}^f$}\xspace}
\newcommand{\xback}{\texttt{$x_{out}^b$}\xspace}
\newcommand{\espresso}{\text{C}}
\newcommand{\psrc}{\texttt{$p$}\xspace}

\newcommand{\cacc}{c^{a}\xspace}

\newcommand{\cunacc}{c^{u}\xspace}

\newcommand{\imgunacc}{\texttt{$x^{u}$}\xspace}
\newcommand{\imgacc}{\texttt{$x^{a}$}\xspace}

\newcommand{\methodfont}[1]{\textsc{#1}}
\newcommand{\method}{\methodfont{AntiMirror}\xspace}
\newcommand{\sdedit}{\methodfont{SDEdit}\xspace}
\newcommand{\moderator}{\methodfont{Mod}\xspace}
\newcommand{\mace}{\methodfont{MACE}\xspace}
\newcommand{\uce}{\methodfont{UCE}\xspace}
\newcommand{\none}{\methodfont{None}\xspace}
\newcommand{\park}{\methodfont{Park et al.}\xspace}
\newcommand{\lietal}{\methodfont{Li et al.}\xspace}

\newcommand{\metricfont}[1]{\texttt{#1}}
\newcommand{\lpips}{\metricfont{LPIPS}\xspace}
\newcommand{\recon}{\metricfont{Recon}\xspace}
\newcommand{\clipscore}{\metricfont{CLIP}\xspace}
\newcommand{\gcdd}{\metricfont{GCD}\xspace}

\usepackage{xcolor}
\definecolor{mynicegreen}{RGB}{171,195,47}
\definecolor{mynicered}{RGB}{255,154,154}
\definecolor{myniceblue}{RGB}{153,153,255}

%% file: 0abstract.tex
\begin{abstract}

Generative models, particularly diffusion-based text-to-image (\tti) models, have demonstrated astounding success. However, aligning them to avoid generating content with \emph{unacceptable concepts}\textemdash e.g., offensive or copyrighted content, or celebrity likenesses\textemdash remains a significant challenge. Concept replacement techniques (\crt{s}) aim to address this challenge, often  by trying to ``\emph{erase}'' unacceptable concepts from models.

Recently, model providers have started offering \emph{image editing} services which accept an image and a text prompt as input, to produce an image altered as specified by the prompt. These are known as image-to-image (\iti) models. 

In this paper, we first use an \iti model to empirically demonstrate that today's state-of-the-art \crt{s} \emph{do not in fact erase unacceptable concepts}. Existing \crt{s} are thus likely to be ineffective in emerging \iti scenarios, despite their effectiveness in \tti settings, highlighting the need to understand this discrepancy between \tti and \iti settings. 

Next, we argue that a good \crt, while replacing unacceptable concepts, should preserve other concepts specified in the inputs to generative models. We call this \emph{fidelity}.  Prior work on \crt{s} have neglected fidelity in the case of unacceptable concepts. 

Finally, we propose the use of \emph{targeted image-editing techniques} to achieve both effectiveness and fidelity. We present such a technique, \method, and demonstrate its viability.

\end{abstract}

%% file: 01introduction.tex
\section{Introduction}
Text-to-image (\tti) diffusion models such as DALLE3~\cite{openai2023dalle3}, Imagen~\cite{saharia2022photorealistic}, and Stable Diffusion~\cite{stable-diffusion} have demonstrated exceptional capabilities in generating high-quality images. Their effectiveness largely stems from training on extensive unfiltered text-image datasets gathered from the Internet. As a result, these models can generate images with \emph{unacceptable concepts}, such as offensive content, copyrighted material, or unauthorized likenesses of celebrities. Content with unacceptable concepts pose ethical, privacy, and legal challenges.

Concept replacement techniques (\crt{s}) have emerged as a way to address this problem. They aim to modify 
the diffusion generation pipeline to prevent unacceptable concepts from being produced.
Many \crt{s} focus on modifying model weights to ensure that a specific concept cannot be generated thereafter, by aiming to \emph{erase} the unacceptable concept from the model~\cite{lu2024mace,gandikota2024unified}. These methods have been shown to be effective in \tti models~\cite{wang2024moderator,lu2024mace, gandikota2024unified}.

Recently, \tti model providers have started offering \emph{image editing} as a service\footnote{\url{https://help.openai.com/en/articles/9055440-editing-your-images-with-chatgpt-images}}. 
It works by accepting an image and a text prompt (describing edits) as input and generating an edited image. They are known as \emph{image-to-image (\iti)} models. Naturally, \crt{s} are needed for \iti models too. Since the same underlying image-generation model is used in both \tti and \iti contexts, existing \crt{s} capable of erasing unacceptable concepts from models should apply in both contexts.

If a model has truly erased a concept, then it should, under no circumstance, produce content depicting that concept. We show that when given an image with an unacceptable concept as input,  image-generation models modified by existing \crt{s} do \emph{reconstruct} it, showing that the \crt{s} did not succeed in erasing the concept. This suggests that existing \crt{s} are unlikely to be effective in \iti settings, despite their effectiveness in the \tti setting.



An alternative \crt approach is 
to detect the unacceptable concept in the output image (e.g., by Espresso~\cite{das2024espressorobustconceptfiltering}), and edit it out using an \emph{editing method} like \sdedit~\cite{meng2022sdedit}. 

A good \crt, while replacing an unacceptable concept, should preserve all other concepts present in its inputs. We call this \emph{fidelity}. Prior work on \crt{s} have neglected fidelity in the case of unacceptable concepts, focusing only on the effectiveness of replacing them. 
Since \sdedit edits the entire image, it will not preserve fidelity.
Effectiveness without fidelity is not useful in concept-replacement settings because fidelity is essential for ensuring overall usability. For instance, if a user inadvertently induces the generation of unacceptable concepts, a model that selectively replaces those concepts in the output is more desirable than one that blocks the entire response due to unacceptability.


We propose a new \crt approach based on \emph{targeted editing}: identify \emph{key characteristics that define the unacceptability of a concept}, and focus on modifying only those characteristics. It is not clear how to do this in general, however, we demonstrate its feasibility in the case of one type of unacceptable concept: unauthorized likenesses of individuals.

Prominent public figures have expressed disapproval of their likenesses being generated by image generation models~\cite{bloomberg2024deepfakes,globe2024pope}. The key characteristics in this case are the facial features that correspond to a person's identity~\cite{jose2020face,clare2018face,rezaei2021face}. We present a targeted editing technique (\autoref{sec:approach_defense}) that changes only these facial features, thereby achieving better fidelity than \sdedit while retaining similar or better effectiveness in replacing celebrity likenesses.

We claim the following contributions: we
\begin{enumerate}[topsep=8pt]
    \item show that state-of-the-art \crt{s} that are effective in the \tti setting \textbf{do not} \textbf{erase unacceptable concepts}, thus limiting their effectiveness in the \iti setting, (\autoref{sec:erasing_experiment})
    \item argue that \crt{s} need to preserve \textbf{fidelity} in addition to being effective, (\autoref{sec:approach_defense}), and
    \item propose \method\footnote{We will open-source the code.}, a new \crt based on targeted image editing that balances fidelity and effectiveness for celebrity likenesses. (\autoref{sec:arch})
\end{enumerate}

%% file: 02background.tex
\section{Background}\label{sec:background}


\input{figures/figure_dm_pipeline}

\subsection{Diffusion Models}\label{back-diffusion-math}

Denoising Diffusion Probabilistic Models (DDPMs)~\cite{HoJA20}, otherwise known as \emph{diffusion models}, are a class of generative models that learn to synthesize data by reversing a gradual noising process. They consist of two processes: a \textit{forward diffusion process} that adds noise to the data, and a \textit{reverse diffusion process} that learns to reconstruct the data from noise.

\subsubsection{Forward Diffusion Process}

The forward process defines a Markov chain that progressively perturbs a data sample (usually an image; $x_0 \in \mathcal{X}$) by adding Gaussian noise over $T$ discrete time steps:

\begin{equation}
q(x_t \mid x_{t-1}) = \mathcal{N}(x_t; \sqrt{1 - \beta_t} x_{t-1}, \beta_t I),
\end{equation}

where $\beta_t \in (0, 1)$ is a predefined variance schedule. This process results in $x_T$ being approximately Gaussian noise.

Due to the linear Gaussian structure, we can marginalize the process to directly sample $x_t$ from $x_0$:

\begin{equation}
q(x_t \mid x_0) = \mathcal{N}(x_t; \sqrt{\bar{\alpha}_t} x_0, (1 - \bar{\alpha}_t)I),
\end{equation}

where $\alpha_t = 1 - \beta_t$ and $\bar{\alpha}_t = \prod_{s=1}^{t} \alpha_s$. Thus, sampling can be performed as:

\begin{equation}
x_t = \sqrt{\bar{\alpha}_t} x_0 + \sqrt{1 - \bar{\alpha}_t} {\epsilon}, \quad {\epsilon} \sim \mathcal{N}(0, I).
\end{equation}

\subsubsection{Reverse Diffusion Process}

The reverse process attempts to learn the denoising transitions $p_\theta(x_{t-1} \mid x_t)$ to recover the original data. These transitions are parameterized as Gaussians:

\begin{equation}
p_\theta(x_{t-1} \mid x_t) = \mathcal{N}(x_{t-1}; {\mu}_\theta(x_t, t), \Sigma_\theta(x_t, t)).
\end{equation}

In practice, the model ${\epsilon}_\theta(x_t, t)$, called the \emph{UNet}~\cite{ronneberger2015unet}, is trained to predict the noise ${\epsilon}$ added at each step. Using this prediction, the mean can be rewritten as:

\begin{equation}
{\mu}_\theta(x_t, t) = \frac{1}{\sqrt{\alpha_t}} \left( x_t - \frac{\beta_t}{\sqrt{1 - \bar{\alpha}_t}} \cdot {\epsilon}_\theta(x_t, t) \right).
\end{equation}

The model then samples $x_{t-1}$ as:

\begin{equation}
x_{t-1} = {\mu}_\theta(x_t, t) + \sigma_t {\epsilon}, \quad {\epsilon} \sim \mathcal{N}(0, I),
\end{equation}

where $\sigma_t$ may be fixed or learned. Prior work keeps this fixed to stabilize UNet training~\cite{HoJA20}.

\subsubsection{Training Objective}

The model is trained to minimize the discrepancy between the true noise ${\epsilon}$ and the predicted noise ${\epsilon}_\theta$ using the objective:

\begin{equation}
\mathcal{L}_{\text{original}} = \mathbb{E}_{x_0, {\epsilon}\sim \mathcal{N}(0,1), t} \left[ \left\| {\epsilon} - {\epsilon}_\theta\left( \sqrt{\bar{\alpha}_t} x_0 + \sqrt{1 - \bar{\alpha}_t} {\epsilon}, t \right) \right\|^2 \right].
\end{equation}

This loss function is equivalent to denoising score matching and is sufficient for generating high-quality samples~\cite{Sohl-DicksteinW15}. This loss function can be further simplified into 

\begin{equation}
\mathcal{L}_{\text{simple}} = \mathbb{E}_{ {\epsilon}\sim \mathcal{N}(0,1), t} \left[ \left\| {\epsilon} - {\epsilon}_\theta(x_t,t) \right\|^2 \right].
\end{equation}

However, this objective is modified for the text-to-image (\tti) and image-to-image (\iti) settings.

\subsubsection{Summary}\label{subsubsec:diffusion-summary}
By training the UNet, \unet, it is able to model $p_\theta(x_{t-1} \mid x_t)$ and approximate the space of images $\mathcal{X}$ starting from Gaussian noise. Crucially, this Gaussian noise can be random, or derived from an input image. In the latter case, the UNet has awareness of the original input image from which the Gaussian noise was derived.

\subsection{\tti Models}\label{back-diffusion}

In the previous section, we explained the forward and reverse diffusion processes, and how the UNet, \unet, of a diffusion model is involved. In practice, for the \tti setting (and \iti setting, which we will explain in ~\autoref{inversion-methods}), the diffusion model, which we now refer to as \dm, also includes text in the input.
Formally, \dm: $\mathcal{X} \times \mathcal{P} \rightarrow \mathcal{X}$, transforms an image $\xin \in \mathcal{X}$ to an image $\xxout \in \mathcal{X}$, with the goal that $\xin = \xxout$. $\mathcal{P}$ represents the space of text prompts with the input $\psrc \in \mathcal{P}$ describing the image $\xin$. The diffusion model contains two main components~\cite{basu2024localizing}; a pre-trained text encoder (usually a \clipscore text encoder~\cite{radford2021learning}), $\phi_p$, and the UNet, \unet. The \clipscore text encoder is used to generate the text embedding of \psrc, which will be used as a conditioning score, $\texttt{c} = \phi_p(\psrc)$, during the generation process. The UNet, as explained previously, is responsible for generating the noise during the reverse process, thus it is the primary generative component of the diffusion model. However, now, the UNet also takes the conditioning score as input as well.

Hence, training a diffusion model involves training the UNet by minimizing
\begin{align} 
\mathcal{L} = \mathbb{E}_{\epsilon \sim \mathcal{N}(0,1), t}[||\epsilon - \epsilon_\theta(\imgclean_t,\texttt{c},t)||_2^2 \label{eqn:baseloss}
\end{align}
for each timestep from $t$, 
and intermediate noised image $\imgclean_t$. 
The UNet noise output, $\epsilon_\theta(\imgclean_t,\texttt{c},t)$, is then used in a sampling algorithm, like DDPM sampling~\cite{NEURIPS2020_DDPM}, to generate the final image. During inference, in the \tti setting, \dm generates images starting from random noise and the text prompt $p$, thus an input image is not provided to \dm.

Stable Diffusion models are known as Latent Diffusion models (LDMs)~\cite{stable-diffusion} as they operate in the latent space of a pre-trained variational autoencoder (VAE) rather than on the images directly, thus being more computationally efficient. Formally, $\imgclean_{in}$ is first passed through the VAE encoder $\mathcal{E}$ to produce the latent $z_{in}$, and the objective function in Equation~\ref{eqn:baseloss} changes to
\[
\mathbb{E}_{\epsilon \sim \mathcal{N}(0,1), t} \big[ ||\epsilon - \epsilon_\theta(z_t,\texttt{c},t)||_2^2.
\]
The output is $\imgclean_{out} = \mathcal{D}(z_0)$, where $\mathcal{D}$ is the VAE decoder. The complete LDM generation process is illustrated in ~\autoref{fig:dm_pipeline}. We summarize commonly used notations in Table~\ref{tab:notations}.
\input{tables/tab_notations}

\subsection{\iti Models}\label{inversion-methods} 
A primary application of \iti models is image editing, where the user supplies an input image, \xin, and an editing prompt, \psrc, describing the desired changes, and the model generates the edited result. Diffusion-based \iti models are the current state-of-the-art. These models rely on inversion.

 Inversion is the process of finding the initial, noisy Gaussian latent $z_T$ (or $x_T$ in the case of a standard diffusion model), that would reconstruct $z_0$ ($x_0$) when passed through a sampling algorithm. This latent is also called the \textit{inverted latent}. Thus, in the \iti setting, \dm generates images starting from the inverted latent— Gaussian noise derived from the input image through the inversion process— and an editing prompt, \psrc. We focus on the state-of-the-art inversion method DDPM Inversion~\cite{huberman2024ddpm}.


DDPM inversion is a stochastic process and serves as a technique to invert the stochastic DDPM sampling algorithm:
\begin{align} 
    z_{t-1} = \hat\mu_t(z_t) + \sigma_t\epsilon_t,\label{eqn:ddpmsampling}
\end{align}

where $z_t$ represents the VAE latents in an LDM, $\epsilon_t \sim \mathcal{N}(0,I)$, and $\bar\alpha_t = \prod_{s=1}^t \alpha_s$, where $\alpha_t = 1 - \beta_t$, as previously.
The noise scale $\sigma_t$ is given by:
\begin{align} 
\sigma_t = \eta\beta_t(1-\alpha_{t-1})/(1-\alpha_t).\label{eqn:sigma}
\end{align}
The predicted mean $\hat\mu_t(z_t)$ is computed as  
\begin{multline}
    \hat\mu_t(z_t) = \sqrt{\frac{\bar\alpha_{t-1}}{\bar\alpha_t}} 
    \left( z_t - \sqrt{1 - \bar\alpha_t} \cdot \epsilon_\theta(z_t, t, \texttt{c}) \right) \\
    + \sqrt{1 - \bar\alpha_{t-1} - \sigma_t^2} \cdot \epsilon_\theta(z_t, t, \texttt{c}). \label{eq:muhat}
\end{multline}

 Later, we will use an \iti model to see how well it can \emph{reconstruct} an image. For reconstruction, the conditioning score $\texttt{c}$ is set to $\varnothing$, which corresponds to the \clipscore text embedding of the empty editing prompt, $``"$. For DDPM sampling, we set $\eta = 1$. DDPM inversion involves generating $z_t$ for $t=1, \dots,  T$ such that they strongly imprint with the random noise values $\{\epsilon_t\}_{t=1}^T$~\cite{huberman2024ddpm}. After inversion, the sampling algorithm is used for reconstruction, thus defining the complete diffusion-based \iti pipeline for image reconstruction.

Some \iti models are not built upon diffusion models, hence they do not use diffusion inversion. \sdedit~\cite{meng2022sdedit} is one such method; like a diffusion model, it relies on iteratively adding noise, and then de-noising. However, the underlying stochastic differential equation (SDE) that formulates this generative process is different. In addition, SDE-based \iti models typically add Gaussian noise from timestep $t=0$ to $t=1$, with the reverse process removing this noise. \sdedit does not require noise addition for all timesteps up to $t=1$. This helps preserve structural details of the original input image, while still facilitating editing. This is desirable for preserving ``fidelity" (a notion that we will return to in~\autoref{sec:approach_defense}). Diffusion-based \iti methods have generally replaced \sdedit, however, it is still used when fidelity is important. 


\subsection{Concept Replacement Techniques}\label{concept-removal-techniques}
Despite their usefulness, diffusion models raise concerns as they can generate images that may contain unacceptable concepts. Concept removal techniques (\crt{s}) aim to prevent the generation of such images. Many \crt{s} do this by attempting to \emph{erase} the corresponding unacceptable concept $\cunacc$ from the weights of the original, unaligned model, \dm. We are specifically concerned with \crt{s} that modify the weights of the UNet, \unet, since this is a common backbone in both \tti and \iti models.
We denote \dm protected by a \crt as the aligned model, \dmcrt. 

In this work, we focus on three \crt{s}: Moderator (\moderator)~\cite{wang2024moderator}, Unified Concept-Editing (\uce)~\cite{gandikota2024unified}, and Mass Concept Erasure (\mace)~\cite{lu2024mace}. 

\moderator~\cite{wang2024moderator} uses task vectors~\cite{ilharco2023editing} to remove the weights corresponding to generating $\imgunacc$, thus causing $\cunacc$ to be replaced. Specifically, they optimize $\theta_{new} = \theta - scale \times \tau_u$, where $\theta_{new}$ are the new weights of the UNet, and $\tau_u$ is the task vector corresponding to $\cunacc$. This task vector is calculated by first overfitting \unet such that it is more prone to generating the unacceptable concept, and using this to identify and modify the weights responsible for this unacceptable generation. 

\uce~\cite{gandikota2024unified} modifies the cross-attention layers of \unet to minimize the influence of $\cunacc$, without changing other concepts. Their optimization minimizes $W$ in:
\[\begin{split}
    \mathcal{L}_{\uce} = \sum_{\cunacc \in \mathcal{C}^{u}, \cacc \in \mathcal{C}^{a}}||W\times\cunacc-W^*\times\cacc||_2^2 \\ 
    +  \sum_{c \in S}||W\times c - W^*\times c||_2^2
\end{split}\]
where $W\in\theta$, $W^*\in\theta^*$ are the parameters of the new and frozen cross-attention layers in \unet, and frozen original UNet\footnote{A frozen version of a model is identical to the original (unfrozen) model, however, its parameters are not updated during optimization.} $\epsilon_{\theta^*}$, $\mathcal{C}^{u}$ and $\mathcal{C}^{a}$ are the spaces of pre-defined unacceptable and acceptable concepts, and $S$ is a set of concepts for which to preserve utility. 

\mace~\cite{lu2024mace} also modifies the UNet's cross-attention layers, but they minimize the following optimization with respect to $W$:

\[\begin{split}
    \mathcal{L}_{\mace} = \sum_{i=1}^{n}||W^*\times e_i^f-W\times e_i^g||_2^2 \\ 
    +  \lambda\sum_{c \in S}||W^*\times c - W\times c||_2^2
\end{split}\]

Here, $e_i^f$ represents the embedding of a word that co-occurs with $\cunacc$, while $e_i^g$ is the embedding of the same word when $\cunacc$ is replaced with a related term (e.g. a celebrity's gender, if $\cunacc$ is a celebrity). Here $\lambda\in\mathbb{R}^+$ is a hyperparameter.

\subsection{Notion of "Erasing" in \crt{s}}
Since the UNet, \unet, is a common backbone in both \tti and \iti settings, a \crt that attempts to erase the concept from the the weights of \unet should prevent unacceptable concept generation in both settings, regardless of the input. This is because, as explained in \autoref{subsubsec:diffusion-summary}, the UNet of the diffusion model (1) has awareness of the original image from which Gaussian noise was derived (which in the \iti setting, refers to the Gaussian noise from inversion), and (2) it is trained to approximate the space of images starting from this noise. Thus, erasing the weights corresponding to unacceptable concept generation should prevent unacceptable images from being generated even in reconstruction, where an unacceptable input image is provided. If unacceptable concepts continue to be generated, then unacceptable images persist in the conditional distributions parametrized the diffusion model, $p_\theta(x_{t-1} \mid x_t)$, thus the weights corresponding to their generation must also persist. 

%% file: figures/figure_dm_pipeline.tex
\begin{figure*}[!t]
  \centering
  \includegraphics[width=\linewidth]{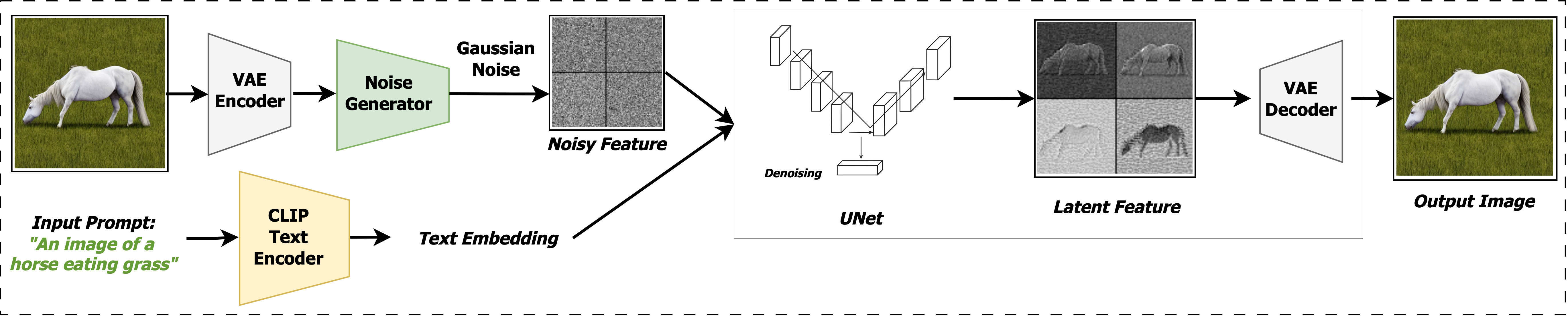}
  \caption{Diffusion process. Standard Gaussian noise is added as part of forward diffusion process, followed by denoising using a UNet in the reverse diffusion process, using a \clipscore text embedding of the input prompt as conditioning.}
  \label{fig:dm_pipeline}
  
\end{figure*}

%% file: tables/tab_notations.tex
\begin{table}[htb]
\caption{Frequently used notations and their descriptions.}
\begin{center}
\footnotesize
\begin{tabular}{ p{1.2cm}|p{4.5cm} } 
 \bottomrule

 \toprule
 \textbf{Notation} & \textbf{Description}\\
 \midrule
  \tti & Text-to-Image\\
  \iti & Image-to-Image\\
  \crt & Concept Replacement Technique\\
  \dm & (Unaligned) diffusion model \\
  \dmcrt & Aligned diffusion model \\
  \unet & UNet of a diffusion model \\
  $\phi_p$ & Pre-trained \clipscore text encoder \\
 \midrule
  $\mathcal{X}$ & Space of images \\
  $\mathcal{P}$ & Space of text prompts \\
  \xin & Input image \\
  \xxout & Output image \\
 $\imgunacc$ & Image with unacceptable concept \\
 $\imgacc$ & Image with acceptable concept \\
 \midrule
 $p_{src}$ & Input textual prompt \\
 $\texttt{c}$ & Conditioning score \\
 $\cacc$ & Acceptable concept\\
 $\cunacc$ & Unacceptable concept\\
 
\midrule
 \bottomrule

 \toprule
\end{tabular}
\end{center}
\label{tab:notations}
\end{table}

%% file: 03approach_recon.tex
\section{Exploring Concept Erasure in \crt{s}}\label{sec:erasing_experiment}

Our goal is to  evaluate whether an unacceptable concept $\cunacc$ has been truly erased from an aligned model, \dmcrt. 
We will demonstrate that the state-of-the-art \crt{s}, \moderator, \mace, and \uce do not erase $\cunacc$ by showing that unacceptable images can be reconstructed from models aligned by these \crt{s}.
We will discuss our datasets (\autoref{sec:collect_img}), experiment configurations (\autoref{sec:gen_results}), and metrics (\autoref{sec:measure_reco}).

\subsection{Datasets}\label{sec:collect_img}
Following prior work ~\cite{wang2024moderator,heng2023selective,das2024espressorobustconceptfiltering}, we prompt a Stable Diffusion model (Stable Diffusion-XL-Base-1.0 (SDXL)\footnote{stabilityai/stable-diffusion-xl-base-1.0} in our case) to obtain a large number of high-quality images containing unacceptable concepts. 
We focus on 3 groups of concepts\textemdash offensive (\emph{nudity}), copyrighted (\emph{Grumpy cat}, \emph{R2D2}), and celebrity-likeness (\emph{Angelina Jolie}, \emph{Taylor Swift}, \emph{Brad Pitt}, \emph{Elon Musk}, \emph{Donald Trump}, \emph{Joe Biden})\textemdash that were used in prior work~\cite{heng2023selective,das2024espressorobustconceptfiltering,zhang2023forgetmenot,wang2024moderator,xiong2024editing,kumari2023ablating,wu2025doco,lu2024mace}. In particular, we generate 50 images per concept by prompting SDXL with the text \emph{An image of $\cunacc$} (e.g., "An image of a nude person"). Alternatively, we could have used real images, however, no such large scale datasets exist for the offensive and copyrighted concepts, and for celebrity-likenesses, only the CelebA dataset exists~\cite{liu2015celeba}, but this is unsuitable since celebrity names are not included. Hence, we opted to use a \tti model (that has not been modified by a \crt) to systematically produce a large dataset.

\subsection{Experiment Configuration}\label{sec:gen_results}
After collecting the images, we provide them as input to the reconstruction pipeline. As stated, we apply the \crt{s} \moderator~\cite{wang2024moderator}, \mace~\cite{lu2024mace}, and \uce~\cite{gandikota2024unified} to its underlying diffusion model beforehand, thus defining the aligned model, \dmcrt. We also evaluate the pipeline with the unaligned model, \dm, as a baseline. In order to perform reconstruction, we set $\texttt{c}=\varnothing$. 
For each of the \crt{s}, we use the exact implementations in their respective GitHub repositories
\footnote{\url{https://github.com/DataSmithLab/Moderator}}\footnote{\url{https://github.com/rohitgandikota/unified-concept-editing}}\footnote{\url{https://github.com/Shilin-LU/MACE}}, and we use SD v1.5 as the backbone, as done in these works. We repeat all of the experiments 5 times and report the mean and standard deviations in \autoref{sec:recon}.

\input{figures/figure_recon_pipeline}

\subsection{Metrics}\label{sec:measure_reco}
After generating the results, we evaluate them using three metrics. 
Two of these metrics, \lpips score~\cite{zhanglpips} and \emph{reconstruction error}, evaluate reconstruction quality from complementary perspectives: \lpips captures perceptual similarity at a semantic level, while reconstruction error provides a pixel-wise comparison. Using both allows for a more comprehensive assessment of how faithfully the reconstructed image preserves the original content, ensuring that subtle perceptual differences and low-level discrepancies are both taken into account.
The third metric, \clipscore score, measures unacceptability, which is not captured by the other metrics. We explain each of these in further detail.

\textbf{\lpips score} is a standard metric used in past editing work~\cite{brack2024ledits,huberman2024ddpm,meng2022sdedit,li2024spdinv} which measures the L2 distance between the latent representations of \xin and \xxout inside a vision model. We follow the guidance of the \lpips authors and use AlexNet as the vision model~\cite{zhanglpips}. The range of an \lpips score is between 0 and 1, where a lower \lpips score denotes greater semantic equivalence. It should be noted that \lpips score is more meaningful when it is used to compare to a baseline; \dmcrt should produce reconstructions, \xxout, of higher \lpips score compared to \dm. 

\textbf{Reconstruction error} is the (normalized) pixel-to-pixel difference between \xin and \xxout: 
\[||\frac{x_{out}}{|x_{out}|}-\frac{x_{in}}{|x_{in}|}||_2\]

A value of 0 denotes identical images, and the value $\sqrt{3 * 512 * 512} = \sqrt{3}*512$, where 3 is the number of channels and 512 is the images' height and width, denotes maximal difference. Once again, \dmcrt should have a large reconstruction error, relative to \dm, when \xin contains the unacceptable concept that the \crt aims to remove.

\textbf{\clipscore score} measures unacceptability by calculating the cosine similarity between a \clipscore text embedding, and a \clipscore image embedding. Following prior work~\cite{das2024espressorobustconceptfiltering,wang2024moderator,lu2024mace,gandikota2024unified}, we use this to denote the semantic equivalence between the text $\cunacc$ and the reconstruction \xxout. Formally, the \clipscore score is calculated as
\[\text{cos}(\phi_x(\xxout),\phi_p(\cunacc))\]
where $\text{cos}(\cdot)$ denotes cosine similarity, and $\phi_x$ and $\phi_p$ are the \clipscore image encoder and \clipscore text encoder, respectively. 
A \clipscore score greater than or equal to 0.25 denotes that $\cunacc$ and $\xxout$ are highly correlated\cite{brack2024ledits}, which means \xxout contains the unacceptable concept $\cunacc$. Thus, effective \crt{s} should achieve low \clipscore score.



\subsection{Results}\label{sec:recon}
We present the results of our experiment in Table~\ref{rec_ddpm}.
Then, we further analyze the reasons for our results in ~\autoref{ineff_finetune}.

For a \crt to be considered effective, it should enable its aligned model to achieve an \lpips score and reconstruction error higher than that from the unaligned model (\none), and the \clipscore score should be lower. In Table~\ref{rec_ddpm}, we find that for every \crt, the metrics fall into 2 cases: (1) they are within the standard deviation of the metrics from \none, or (2) they are able to reconstruct \xin better than \none.
This indicates that \crt{s} have a negligible or even detrimental effect in preventing the reconstruction of $\imgunacc$, as shown in ~\autoref{fig:recon_pipeline}. 


\input{tables/unlearn_metrics_ddpm}

For many of the results, the \lpips and \recon values were lower than those from \none, and the \clipscore scores were greater, with at least one of these metrics being beyond the standard deviation from \none. These concepts and \crt{s} are \emph{Angelina Jolie} for all \crt{s}, \emph{Taylor Swift} for \moderator and \mace, \emph{Brad Pitt} and \emph{Elon Musk} for all \crt{s}, \emph{Donald Trump} for \moderator and \mace, and \emph{Joe Biden} for all \crt{s}. This directly contradicts the intended role of a \crt, as the removal of $\cunacc$ should make reconstructing $\imgunacc$ more difficult. This effect may arise because  diffusion models rely on a different sets of weights in the \tti and \iti settings. We further investigate this phenomenon in~\autoref{ineff_finetune}.
The only cases where the \crt was somewhat successful in preventing reconstruction were \moderator for \emph{Angelina Jolie} and \uce for \emph{Taylor Swift}. This is evidenced by higher \lpips and \recon values, and a lower \clipscore score for \emph{Angelina Jolie}, with the \recon value notably exceeding the standard deviation of the \recon value from \none. However, visually, the unacceptable concepts of \emph{Angelina Jolie} and \emph{Taylor Swift} were still present since the faces in \xin and \xxout were nearly identical to the human eye, though with less prominent skin lines and slight skin tone changes. See \autoref{tab:recon_images_crts} for some examples of reconstructions. All the \crt{s} and concepts yielded similar results.

\input{tables/crts_celebs}

\subsection{Discussion}\label{ineff_finetune}
We conjecture that a reason why \crt{s} fail to erase concepts is that they are too dependent on the text prompt rather than the intermediate generated latents, $z_t$, which actually contain the unacceptable concept. For example, \mace and \uce alter the cross-attention layers of the UNet in \dm, which are designed to fuse text information with $z_t$. Hence, our results indicate the possibility that cross-attention layers allocate more attention to the text rather than on the intermediate latent containing $\cunacc$. On the other hand, although \moderator uses task vectors to remove $\cunacc$ from all of the weights, a significant part of calculating task vectors involves over-fitting the model on unacceptable images, $\imgunacc$, \emph{and} their corresponding text prompts. Thus, \moderator has an indirect reliance on text, causing it to fail in preventing the generation of $\cunacc$ in the \iti setting. Since the  prompt we use is empty ($``"$), there is no useful textual information that the \crt could use, thus causing them to fail. 
Overall, we leave a deeper investigation into the properties of \iti models and their interactions with \crt{s}—including why modifying certain weights leads to more successful reconstructions—as future work.

%% file: figures/figure_recon_pipeline.tex
\begin{figure*}[!t]
  \centering
  \includegraphics[width=\linewidth]{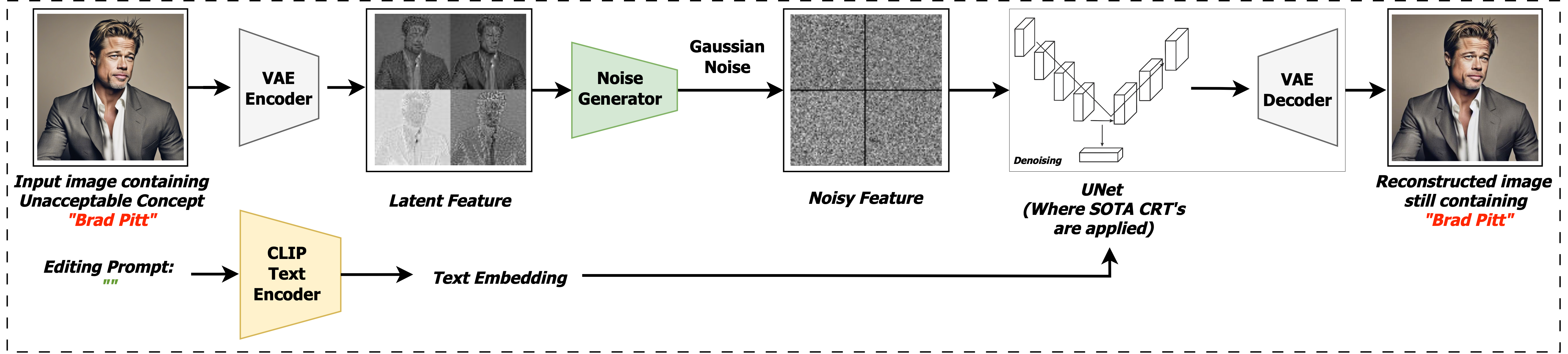}
  \caption{Diffusion-based image generation pipeline. An input image is encoded into a latent space via a VAE. Standard Gaussian noise is added as part of the DDPM inversion, followed by denoising using a UNet enhanced with various state-of-the-art \crt{s}. Despite using \crt{s}, when the input image contains an unacceptable concept, it is not replaced during reconstruction}
  \label{fig:recon_pipeline}
  
\end{figure*}

%% file: tables/unlearn_metrics_ddpm.tex
\begin{table*}[htbp]
  \centering
  \footnotesize 
  \caption{Reconstruction results when a CRT (\moderator, \mace, or \uce) is applied, compared to the baseline without CRT (\none). \lpips scores range from 0 (identical) to 1 (maximal different). \recon ranges from 0 (every pixel is identical) to $\sqrt{3}*512$ (every pixel is maximally different). \clipscore scores at or above 0.25 indicate semantic equivalence~\cite{brack2024ledits}. We highlight in \colorbox{mynicegreen}{green} the cases where a \crt is able to prevent unacceptable reconstruction, as denoted by \lpips and \recon being higher or \clipscore being lower than that from \none, beyond the standard deviation.}
  \label{rec_ddpm}
  \renewcommand{\arraystretch}{1.25}
  \setlength{\tabcolsep}{3pt} 
\resizebox{\textwidth}{!}{
  \begin{tabular}{lccc||ccc|ccc|ccc}
    \toprule
    & \multicolumn{3}{c||}{\none} & \multicolumn{3}{c|}{\moderator} & \multicolumn{3}{c|}{\mace} & \multicolumn{3}{c}{\uce} \\
    Concept & \lpips & \recon & \clipscore & \lpips & \recon & \clipscore & \lpips & \recon & \clipscore & \lpips & \recon & \clipscore \\
    \midrule
    \multicolumn{13}{c}{\textit{Celebrity- likeness}} \\
    \midrule
    \textit{Angelina Jolie} & 0.13 $\pm$ 0.02 & 82.20 $\pm$ 2.31 & 0.25 $\pm$ 0.02 & 0.15 $\pm$ 0.02 & \colorbox{mynicegreen}{117.30 $\pm$ 1.78} & 0.23 $\pm$ 0.02 & 0.09 $\pm$ 0.02 & 47.78 $\pm$ 2.14 & 0.25 $\pm$ 0.01 & 0.09 $\pm$ 0.01 & 47.78 $\pm$ 1.95 & 0.25 $\pm$ 0.00 \\
    \textit{Taylor Swift} & 0.13 $\pm$ 0.02 & 83.20 $\pm$ 1.64 & 0.23 $\pm$ 0.02 & 0.11 $\pm$ 0.02 & 56.61 $\pm$ 2.45 & 0.24 $\pm$ 0.02 & 0.11 $\pm$ 0.02 & 56.54 $\pm$ 1.23 & 0.24 $\pm$ 0.02 & 0.15 $\pm$ 0.01 & \colorbox{mynicegreen}{103.45 $\pm$ 2.91} & 0.24 $\pm$ 0.01 \\
    \textit{Brad Pitt} & 0.15 $\pm$ 0.02 & 100.19 $\pm$ 2.76 & 0.26 $\pm$ 0.02 & 0.12 $\pm$ 0.03 & 69.39 $\pm$ 1.23 & 0.27 $\pm$ 0.00 & 0.12 $\pm$ 0.01 & 69.33 $\pm$ 2.58 & 0.27 $\pm$ 0.02 & 0.12 $\pm$ 0.01 & 69.33 $\pm$ 1.37 & 0.27 $\pm$ 0.02 \\
    \textit{Elon Musk} & 0.20 $\pm$ 0.05 & 157.29 $\pm$ 1.93 & 0.26 $\pm$ 0.02 & 0.15 $\pm$ 0.01 & 110.23 $\pm$ 2.91 & 0.27 $\pm$ 0.00 & 0.17 $\pm$ 0.00 & 80.55 $\pm$ 1.71 & 0.28 $\pm$ 0.00 & 0.17 $\pm$ 0.02 & 80.58 $\pm$ 3.46 & 0.28 $\pm$ 0.01 \\
    \textit{Donald Trump} & 0.16 $\pm$ 0.03 & 109.41 $\pm$ 1.28 & 0.24 $\pm$ 0.01 & 0.14 $\pm$ 0.01 & 85.21 $\pm$ 2.31 & 0.24 $\pm$ 0.00 & 0.14 $\pm$ 0.02 & 85.21 $\pm$ 2.19 & 0.24 $\pm$ 0.01 & 0.16 $\pm$ 0.01 & 112.69 $\pm$ 1.95 & 0.23 $\pm$ 0.00 \\
    \textit{Joe Biden} & 0.15 $\pm$ 0.02 & 148.25 $\pm$ 2.19 & 0.24 $\pm$ 0.02 & 0.10 $\pm$ 0.01 & 83.89 $\pm$ 1.94 & 0.26 $\pm$ 0.00 & 0.10 $\pm$ 0.01 & 83.94 $\pm$ 1.71 & 0.26 $\pm$ 0.02 & 0.15 $\pm$ 0.01 & 136.57 $\pm$ 1.95 & 0.24 $\pm$ 0.01 \\
    \midrule
    \multicolumn{13}{c}{\textit{Offensive}} \\
    \midrule
    \textit{Nudity} & 0.01 $\pm$ 0.01 & 12.19 $\pm$ 1.21 & 0.24 $\pm$ 0.01 &  0.01 $\pm$ 0.00 & 12.20 $\pm$ 1.01  & 0.24 $\pm$ 0.01 & 0.01 $\pm$ 0.01 & 12.22 $\pm$ 1.02 & 0.24 $\pm$ 0.00 & 0.01 $\pm$ 0.00 & 12.20 $\pm$ 1.02 & 0.24 $\pm$ 0.01 \\
    \midrule 
    \multicolumn{13}{c}{\textit{Copyrighted}} \\
    \midrule
    \textit{Grumpy cat} & 0.03 $\pm$ 0.02 & 33.04 $\pm$ 0.89 & 0.25 $\pm$ 0.01 & 0.03 $\pm$ 0.00 & 33.09 $\pm$ 0.55 & 0.25 $\pm$ 0.01 & 0.03 $\pm$ 0.01 & 33.00 $\pm$ 0.79 & 0.25 $\pm$ 0.00 & 0.03 $\pm$ 0.00 & 33.05 $\pm$ 0.55 & 0.25 $\pm$ 0.00 \\
    \textit{R2D2} & 0.03 $\pm$ 0.00 & 37.96 $\pm$ 0.66 & 0.25 $\pm$ 0.00  & 0.03 $\pm$ 0.00 & 38.10 $\pm$ 0.51 & 0.24 $\pm$ 0.01 & 0.03 $\pm$ 0.00 & 38.05 $\pm$ 10.44 & 0.25 $\pm$ 0.00 & 0.03 $\pm$ 0.00 & 37.96 $\pm$ 0.46 & 0.25 $\pm$ 0.00 \\
    \bottomrule
  \end{tabular}}
\end{table*}

%% file: tables/crts_celebs.tex
\begin{table*}[htbp]
\centering
\caption{Examples of reconstruction with different \crt{s} (\none, \moderator, \mace, and \uce)}
\label{tab:recon_images_crts}
\setlength{\tabcolsep}{2pt}
\renewcommand{\arraystretch}{1.3}
\begin{tabular}{m{1.8cm} 
                >{\centering\arraybackslash}m{0.12\linewidth} 
                >{\centering\arraybackslash}m{0.12\linewidth} 
                >{\centering\arraybackslash}m{0.12\linewidth} 
                >{\centering\arraybackslash}m{0.12\linewidth} 
                >{\centering\arraybackslash}m{0.12\linewidth} 
                >{\centering\arraybackslash}m{0.12\linewidth}}
    \toprule
    Celebrity ($\rightarrow$) & \emph{Angelina Jolie} & \emph{Taylor Swift} & \emph{Brad Pitt} & \emph{Elon Musk} & \emph{Donald Trump} & \emph{Joe Biden} \\
    \midrule
    
    \parbox[c]{\linewidth}{\centering Input} &
    \includegraphics[width=\linewidth]{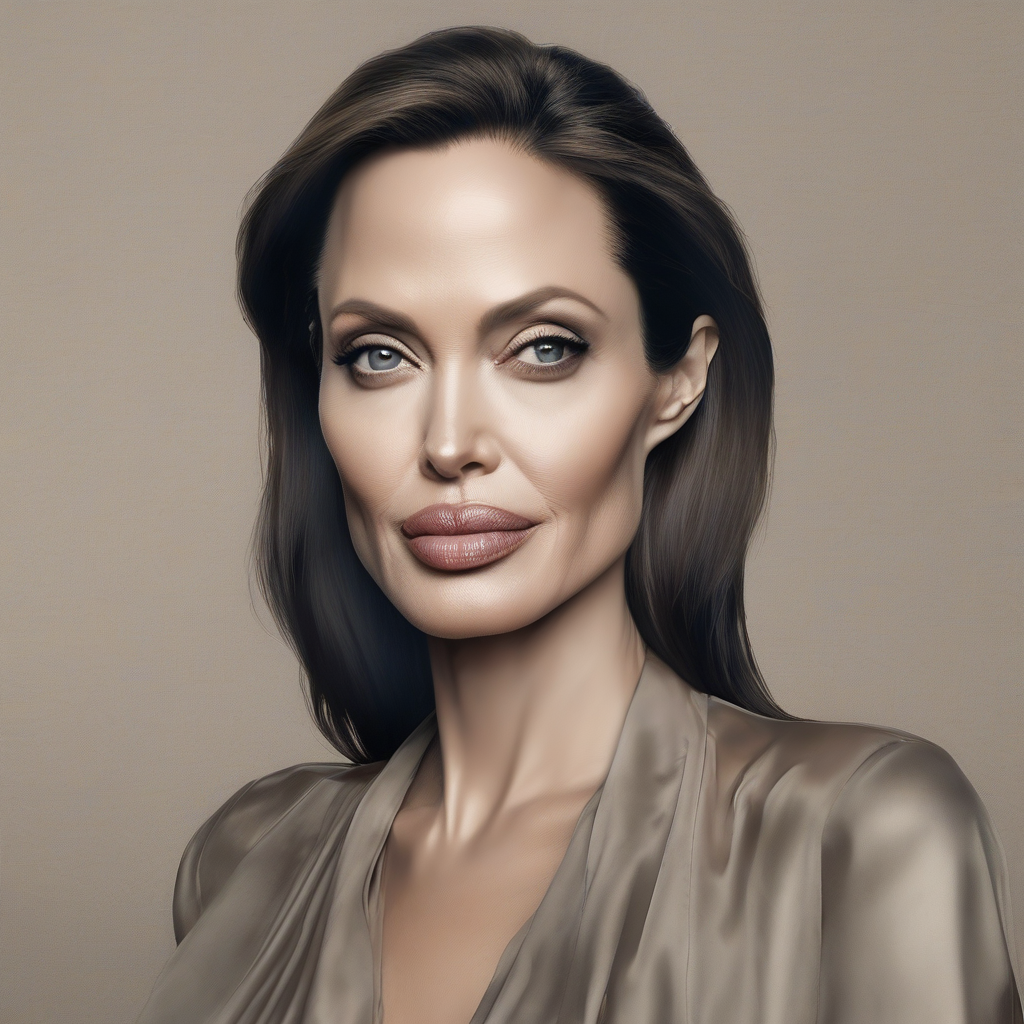} &
    \includegraphics[width=\linewidth]{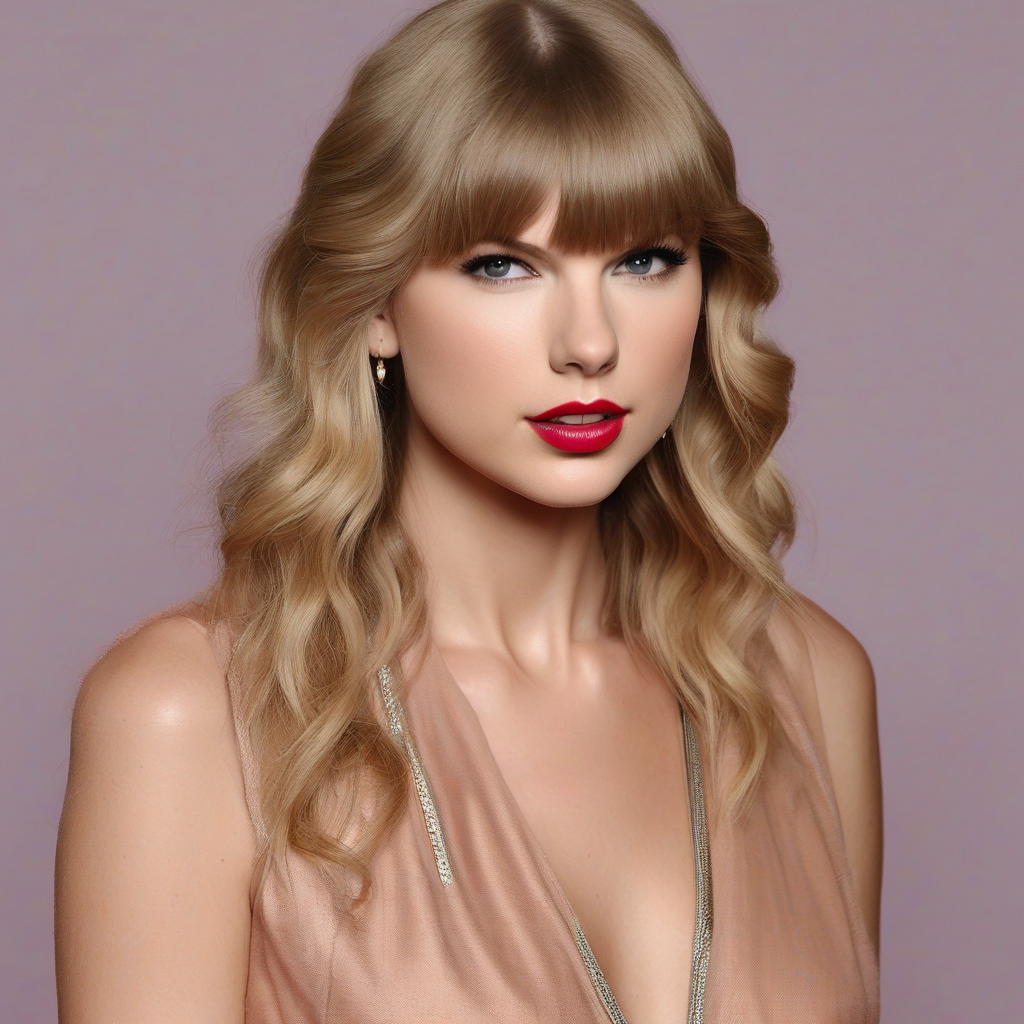} &
    \includegraphics[width=\linewidth]{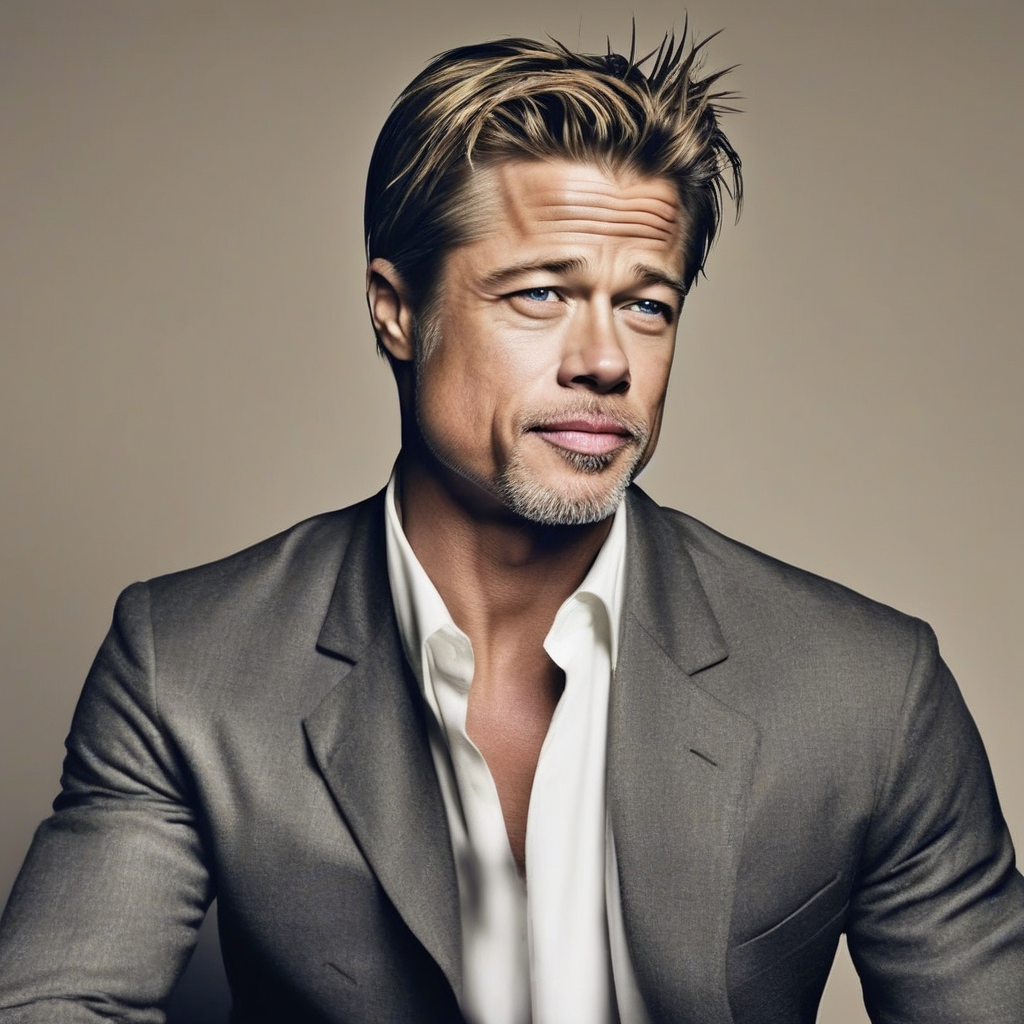} &
    \includegraphics[width=\linewidth]{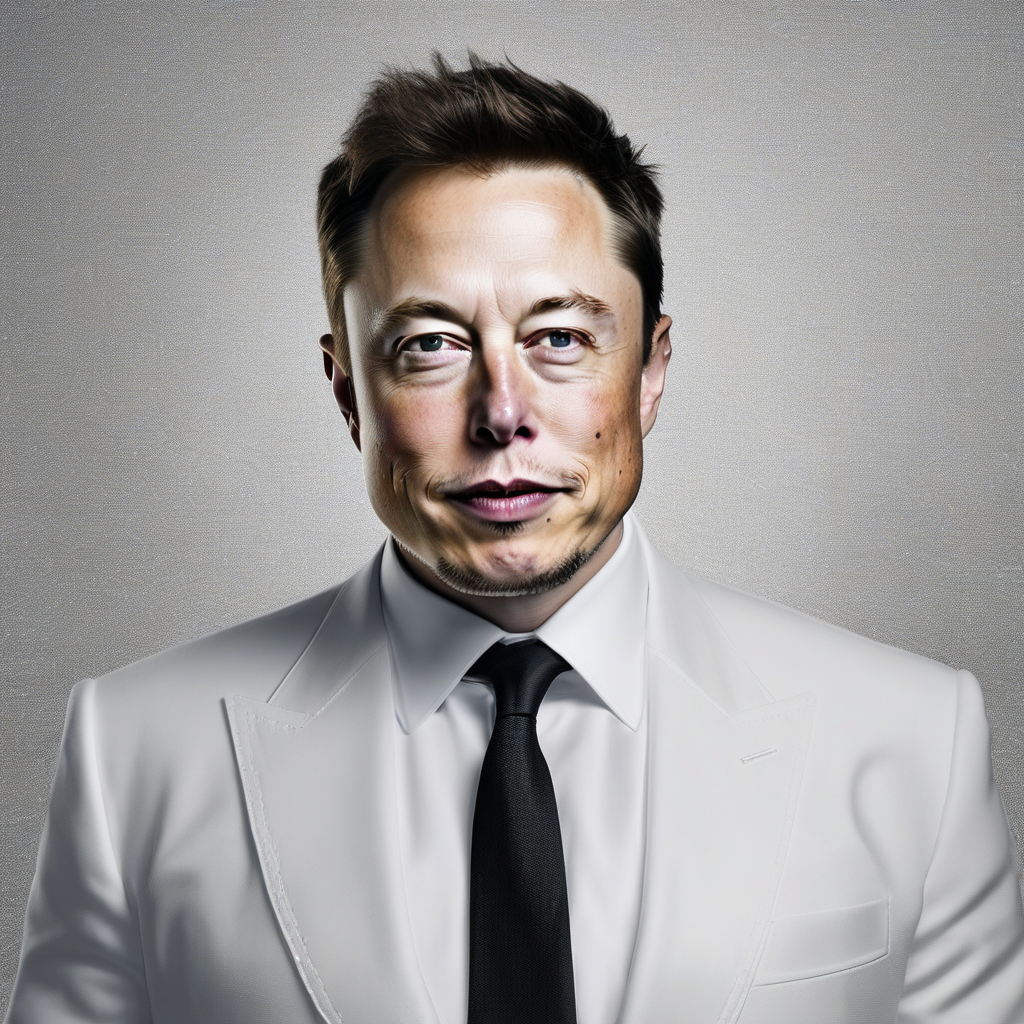} &
    \includegraphics[width=\linewidth]{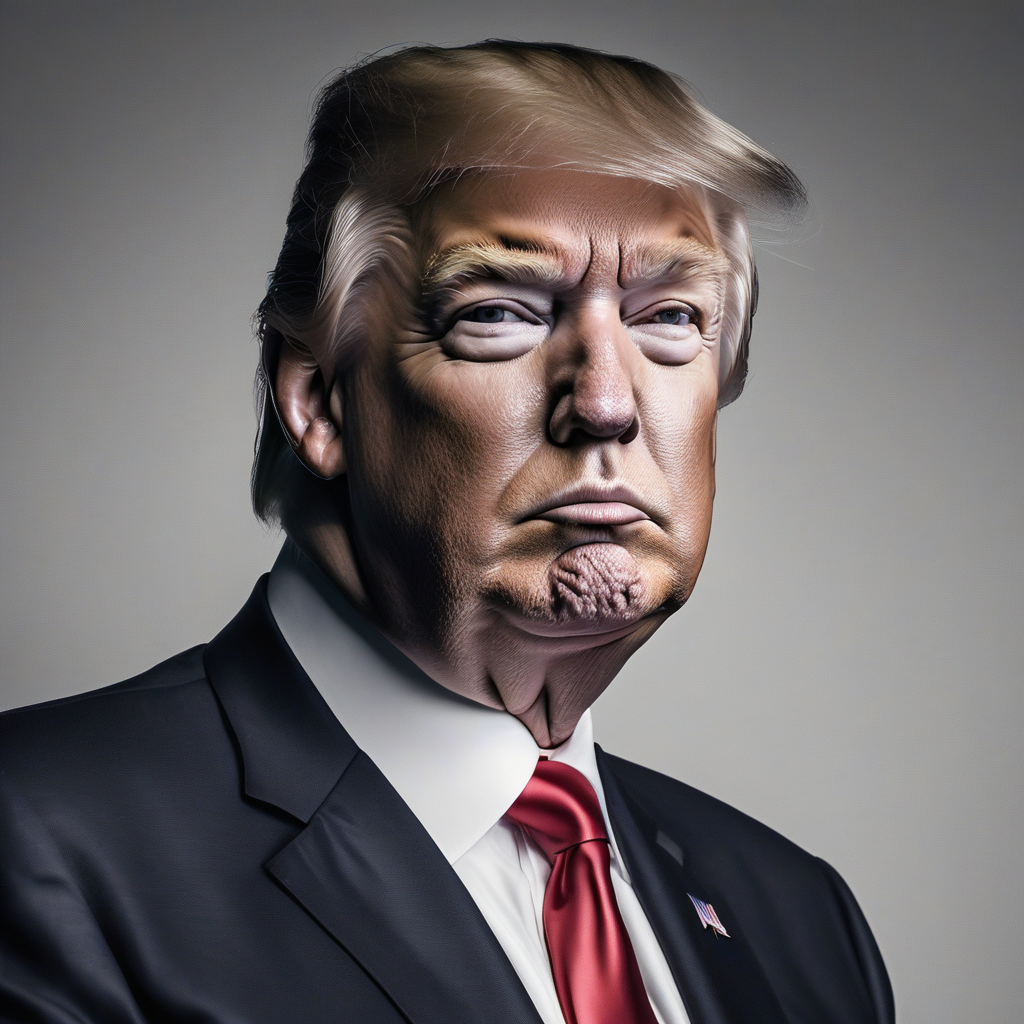} &
    \includegraphics[width=\linewidth]{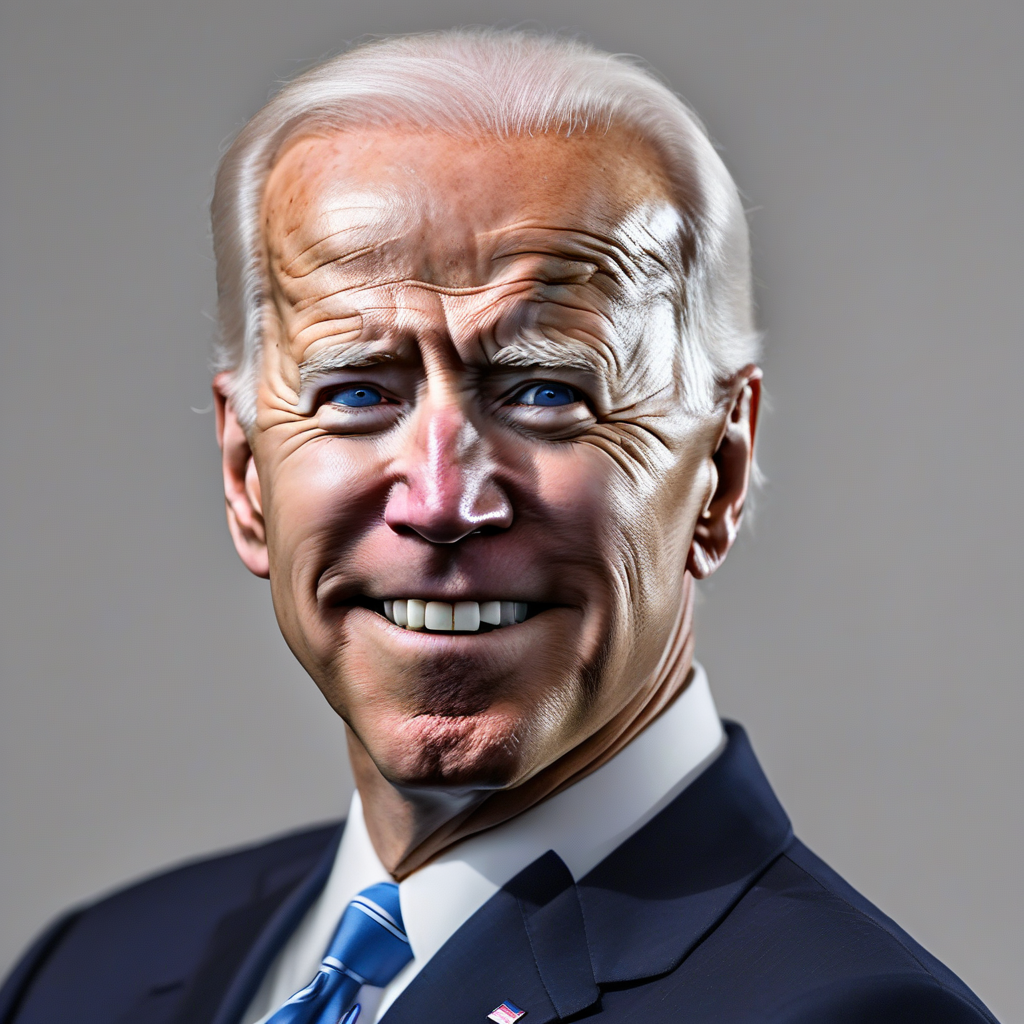} \\
    \midrule
    \multicolumn{1}{c}{CRT~($\downarrow$)} &
    \multicolumn{6}{c}{Output} \\
    \midrule
    \parbox[c]{\linewidth}{\centering \none} &
    \includegraphics[width=\linewidth]{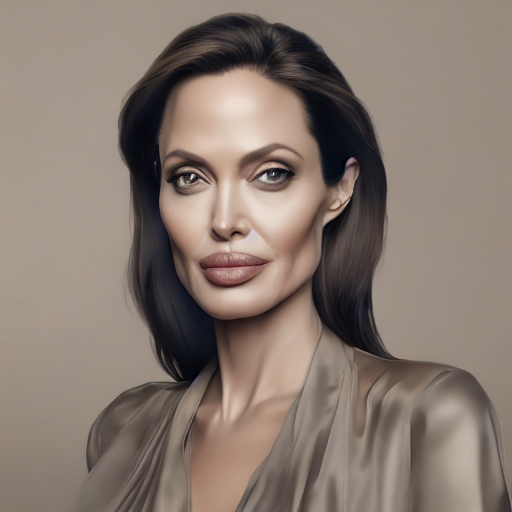} &
    \includegraphics[width=\linewidth]{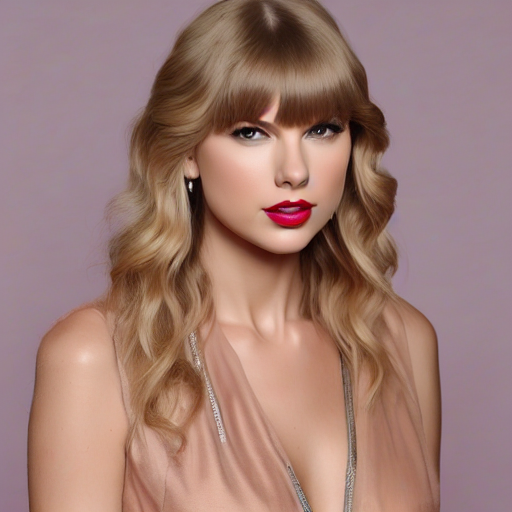} &
    \includegraphics[width=\linewidth]{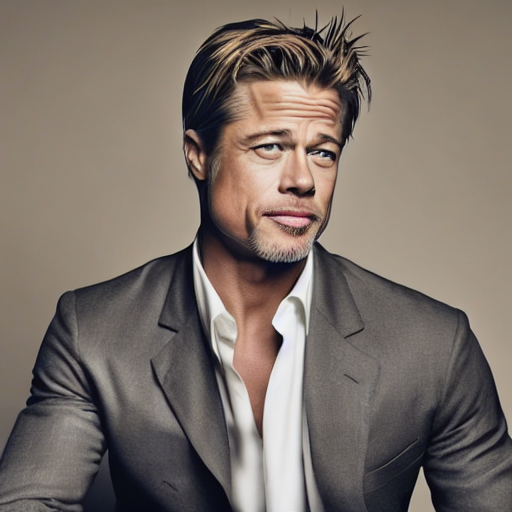} &
    \includegraphics[width=\linewidth]{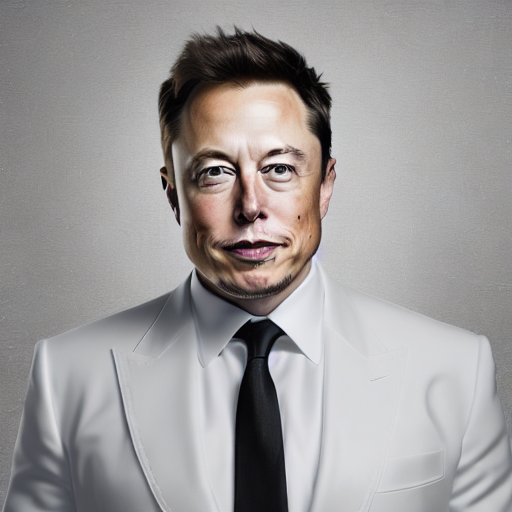} &
    \includegraphics[width=\linewidth]{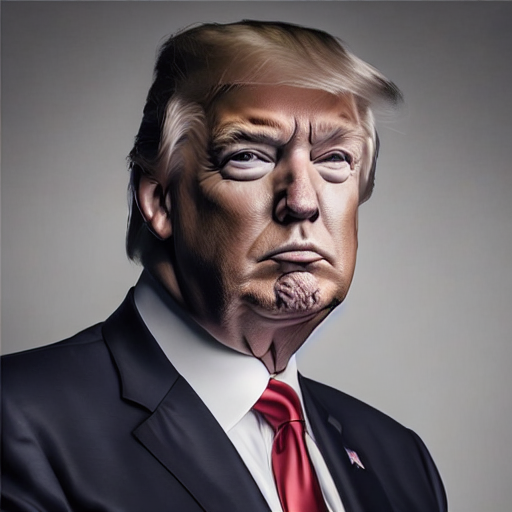} &
    \includegraphics[width=\linewidth]{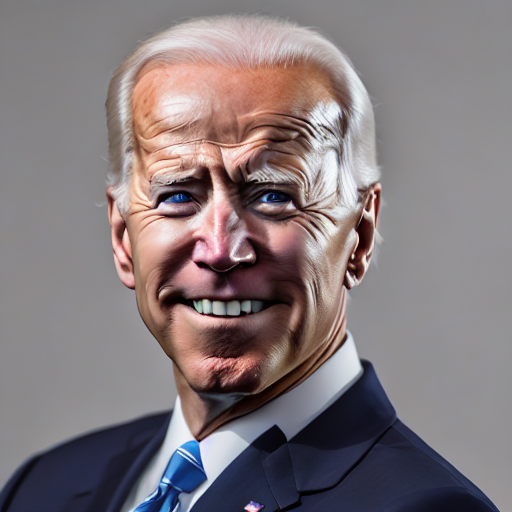} \\
    
    \midrule
    \parbox[c]{\linewidth}{\centering \moderator} &
    \includegraphics[width=\linewidth]{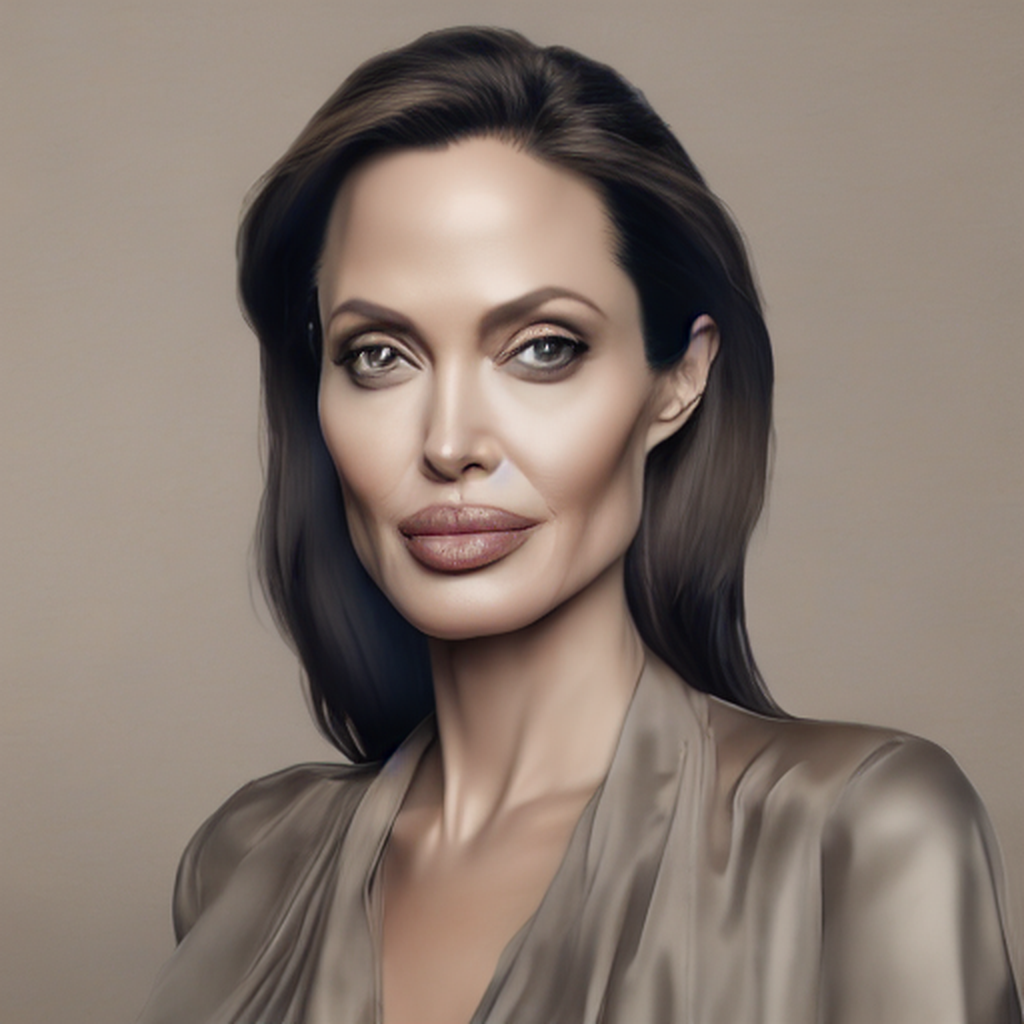} &
    \includegraphics[width=\linewidth]{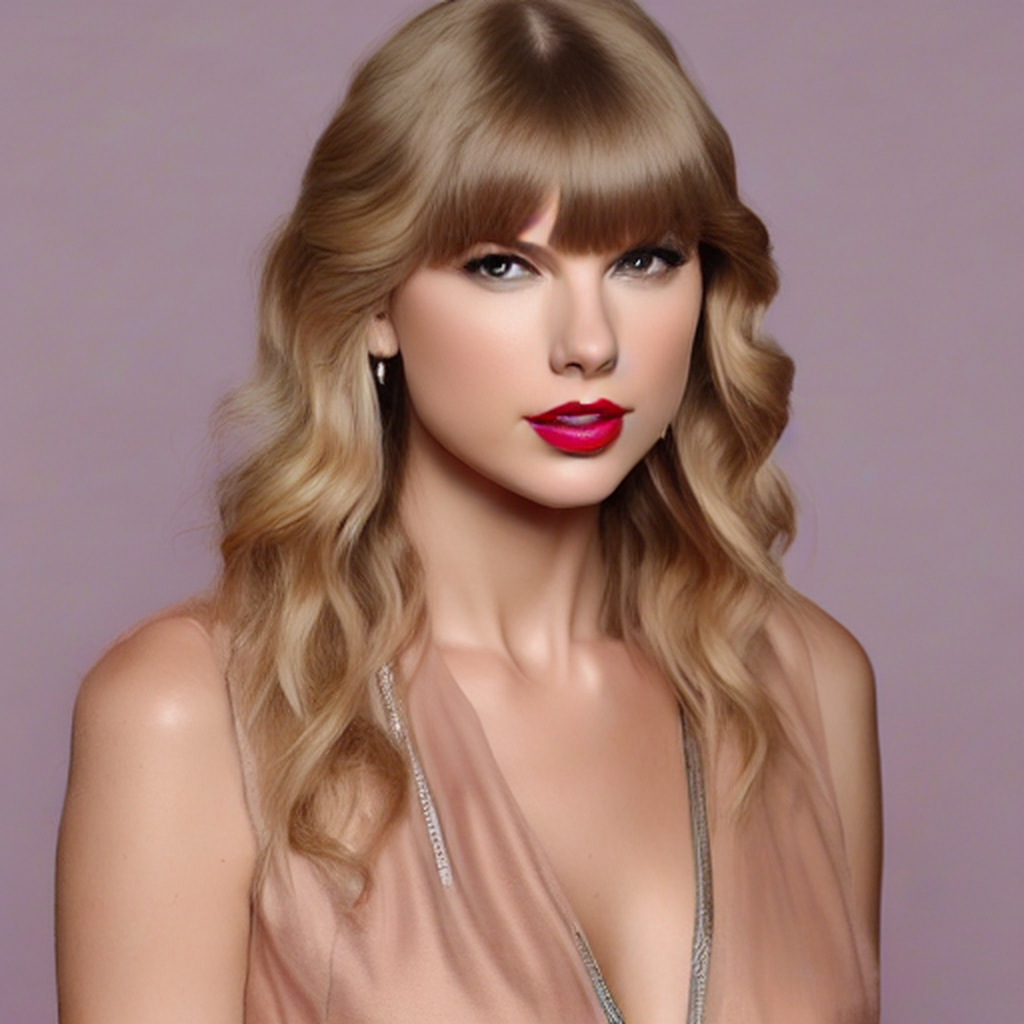} &
    \includegraphics[width=\linewidth]{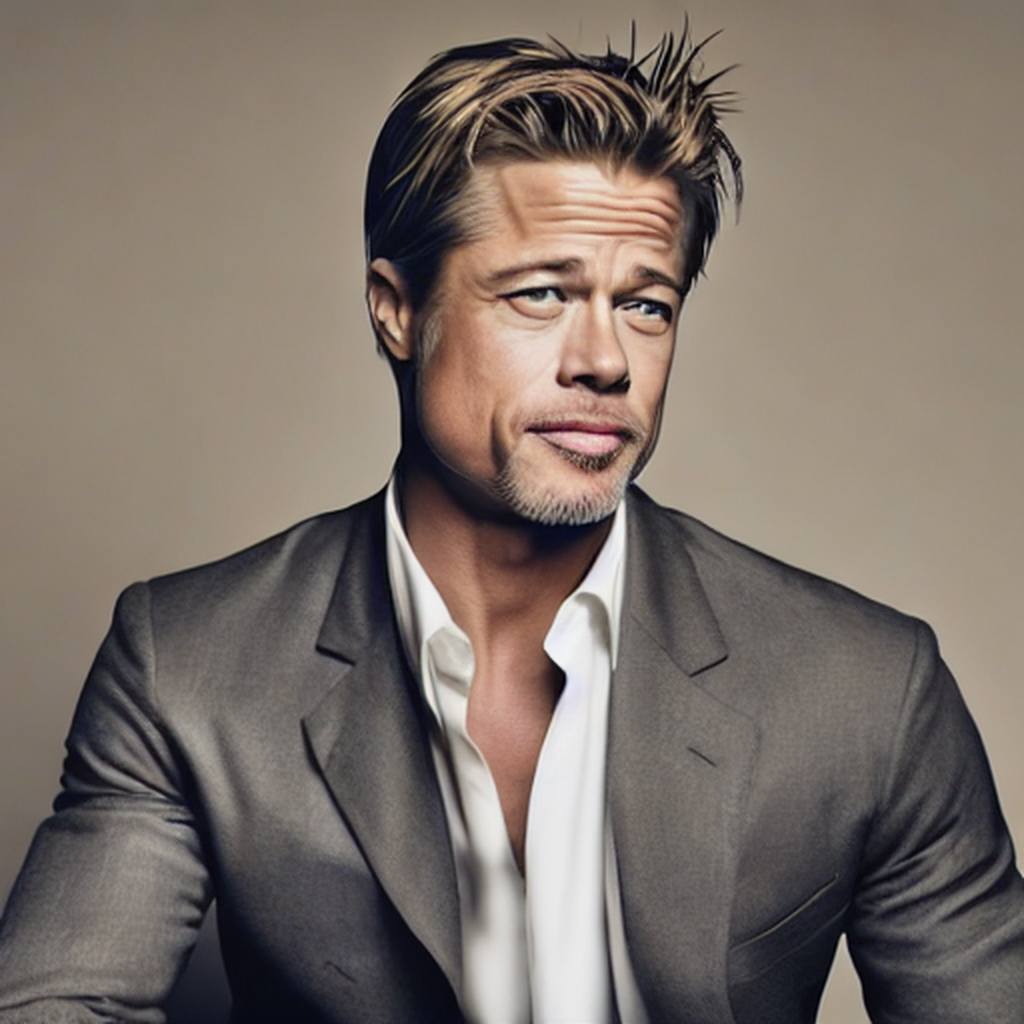} &
    \includegraphics[width=\linewidth]{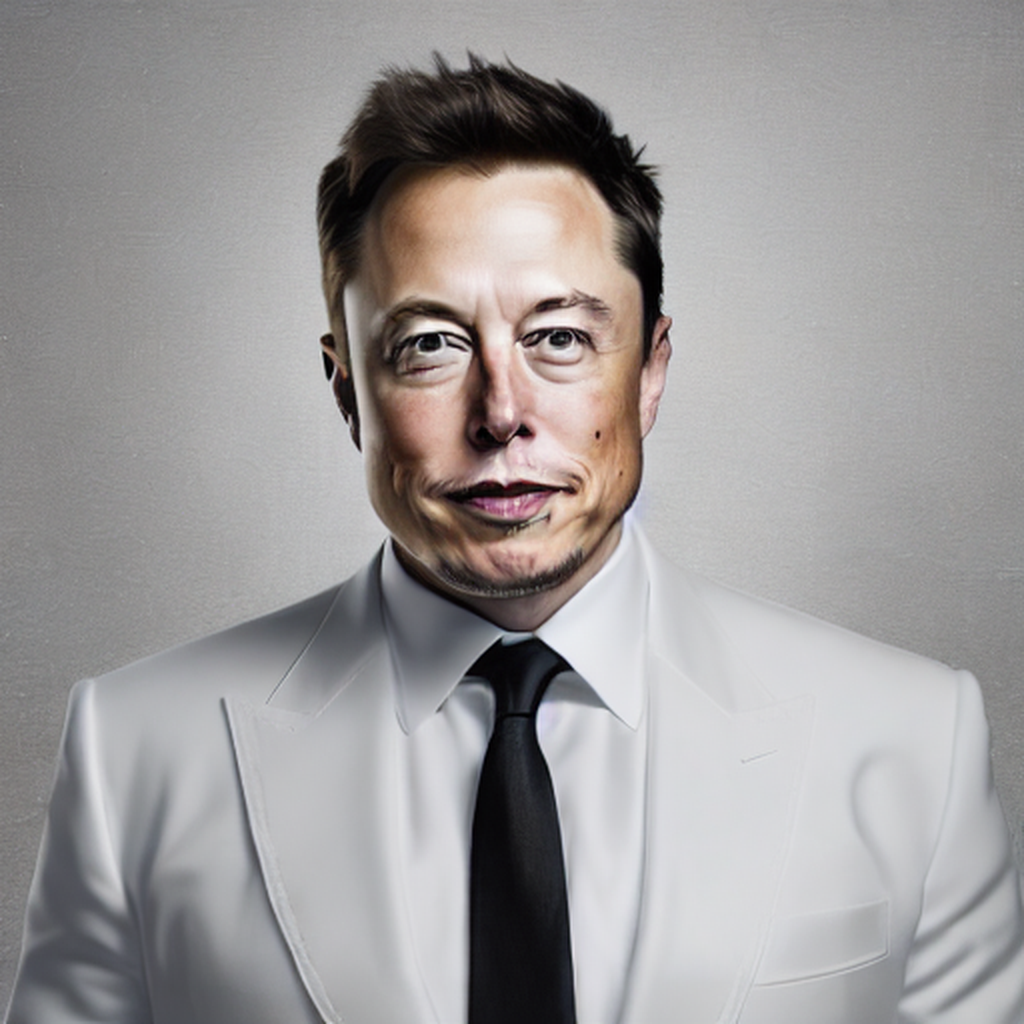} &
    \includegraphics[width=\linewidth]{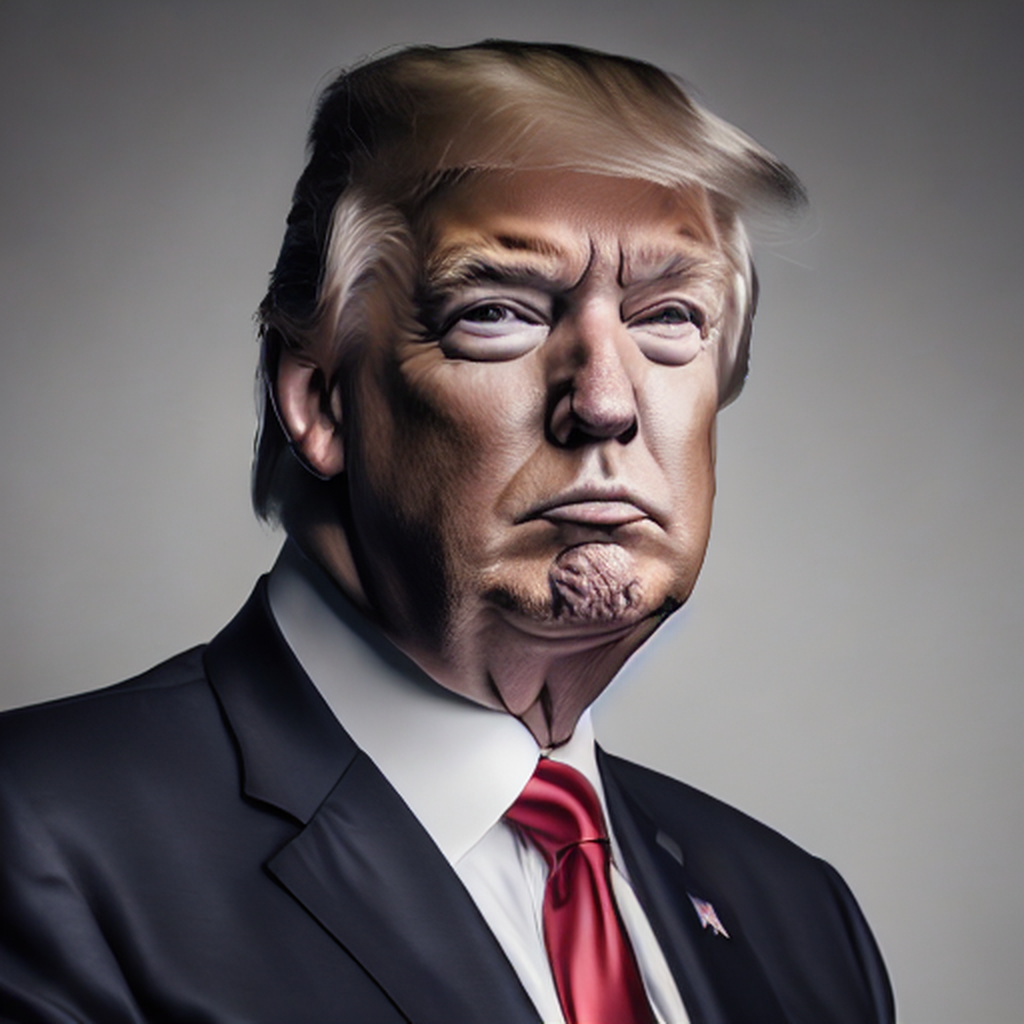} &
    \includegraphics[width=\linewidth]{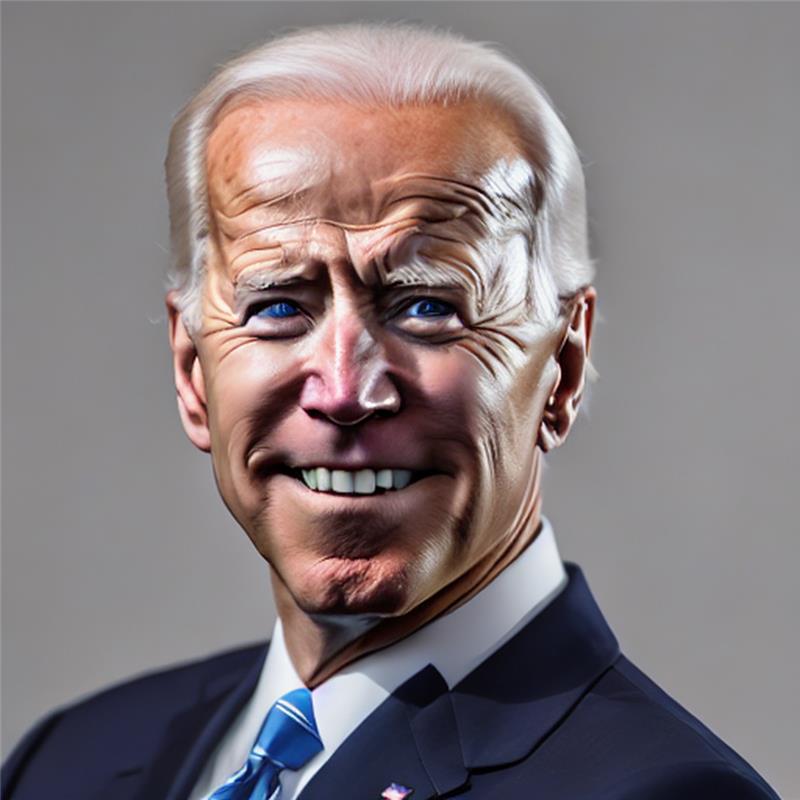} \\
    
    \midrule
    \parbox[c]{\linewidth}{\centering \mace} &
    \includegraphics[width=\linewidth]{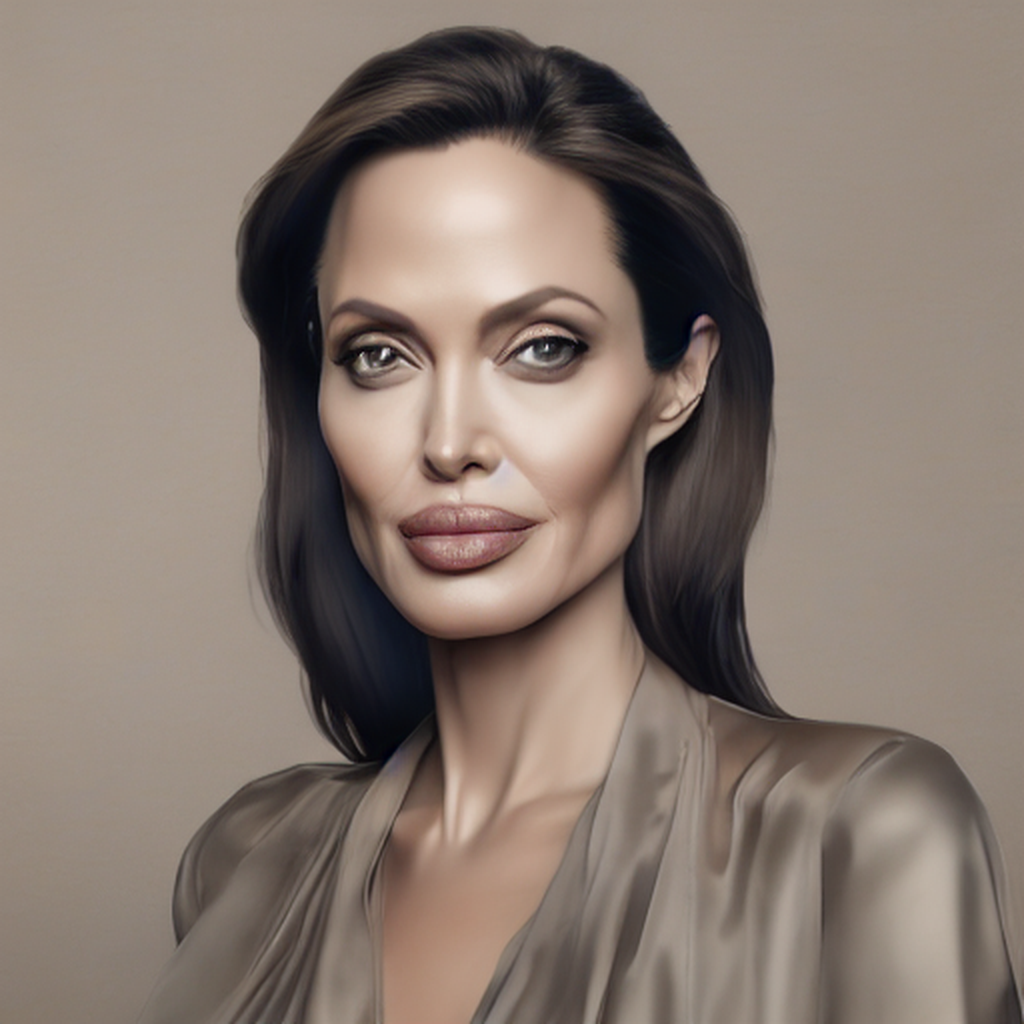} &
    \includegraphics[width=\linewidth]{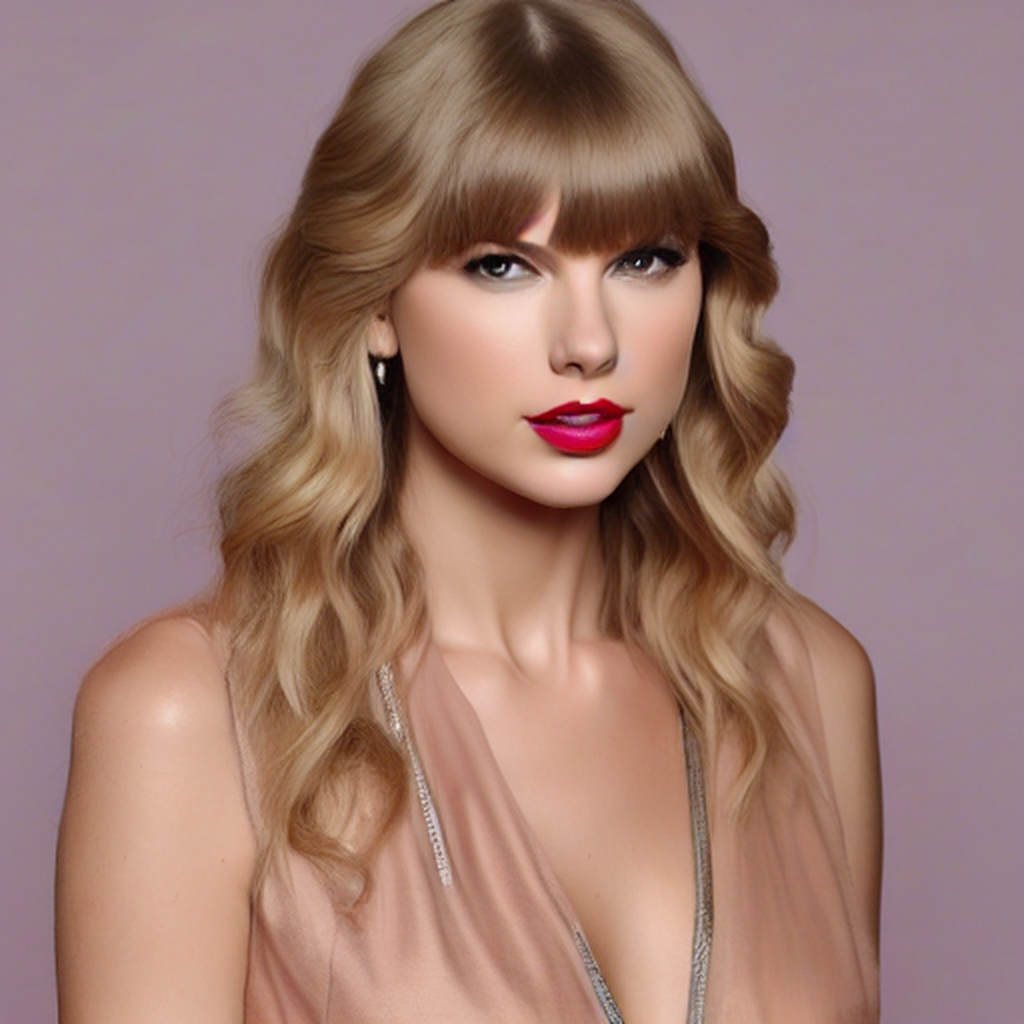} &
    \includegraphics[width=\linewidth]{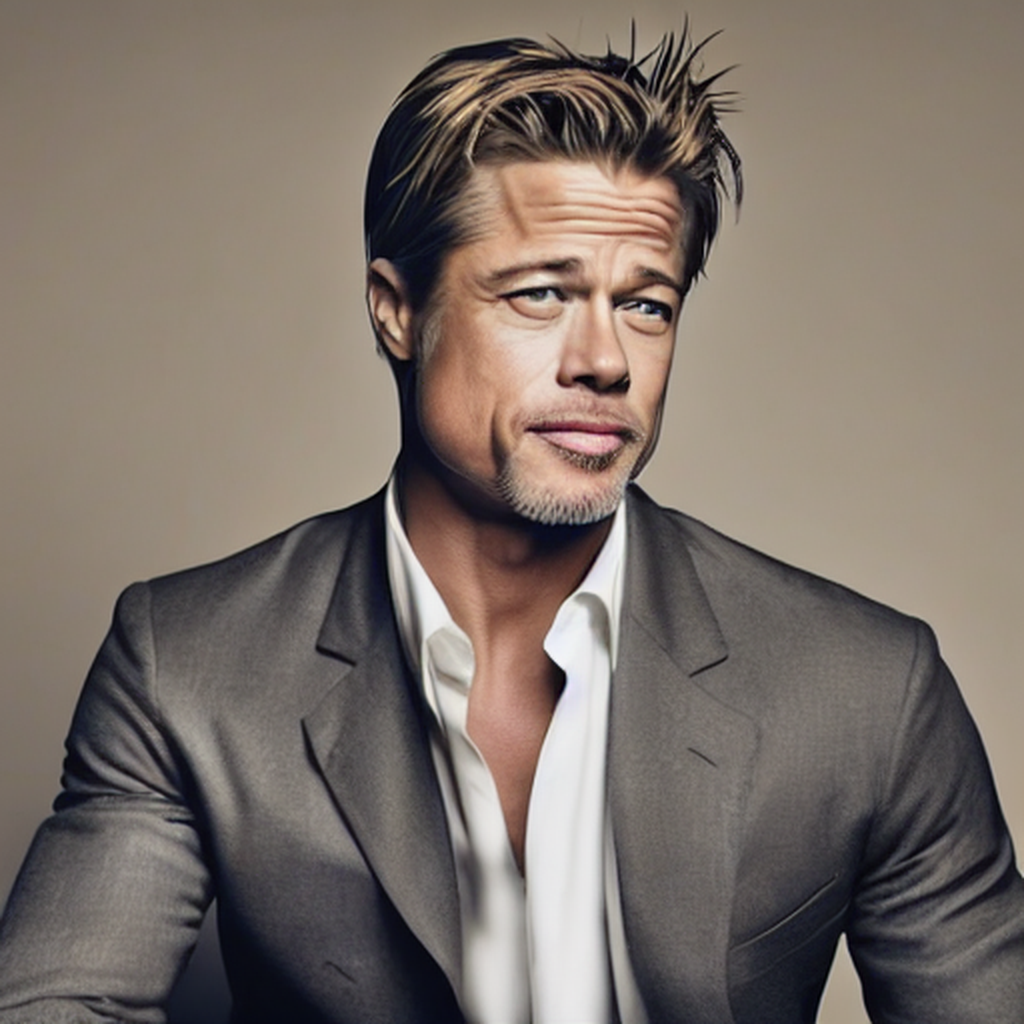} &
    \includegraphics[width=\linewidth]{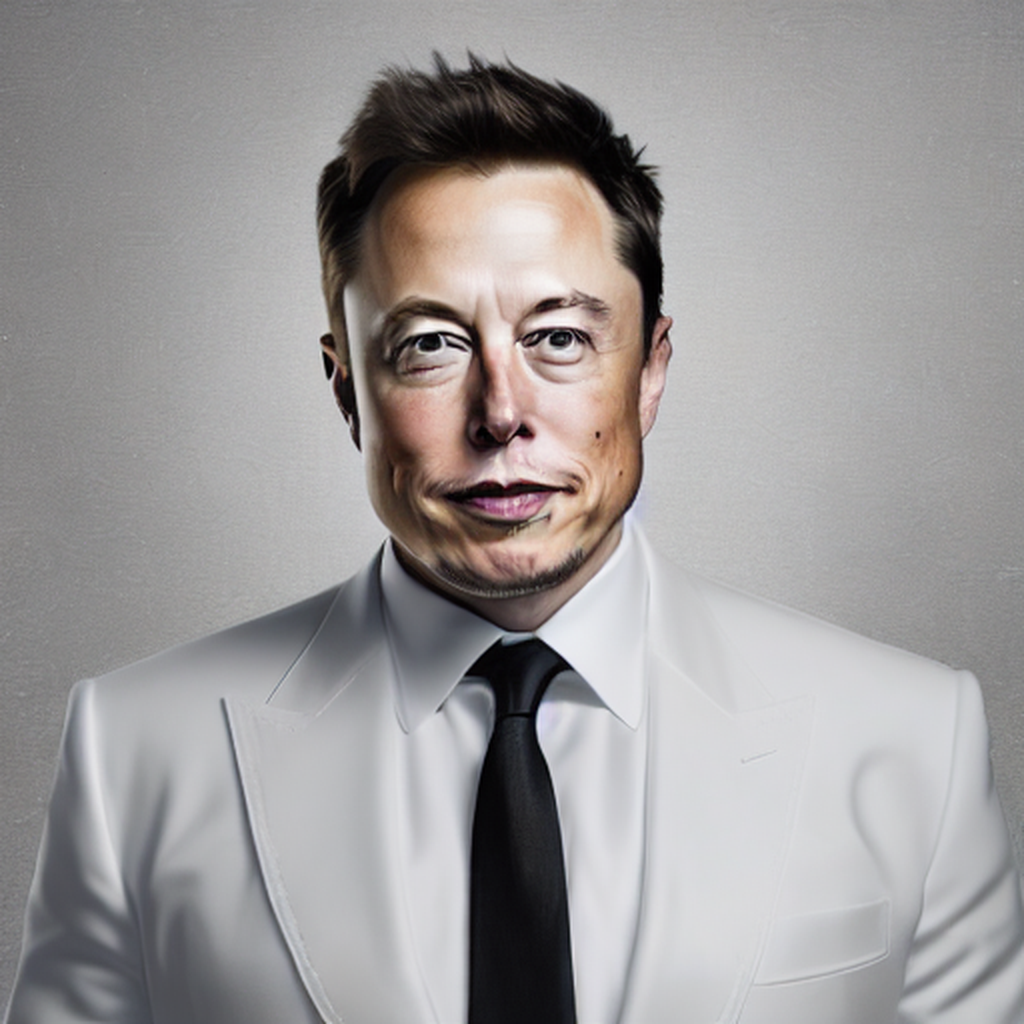} &
    \includegraphics[width=\linewidth]{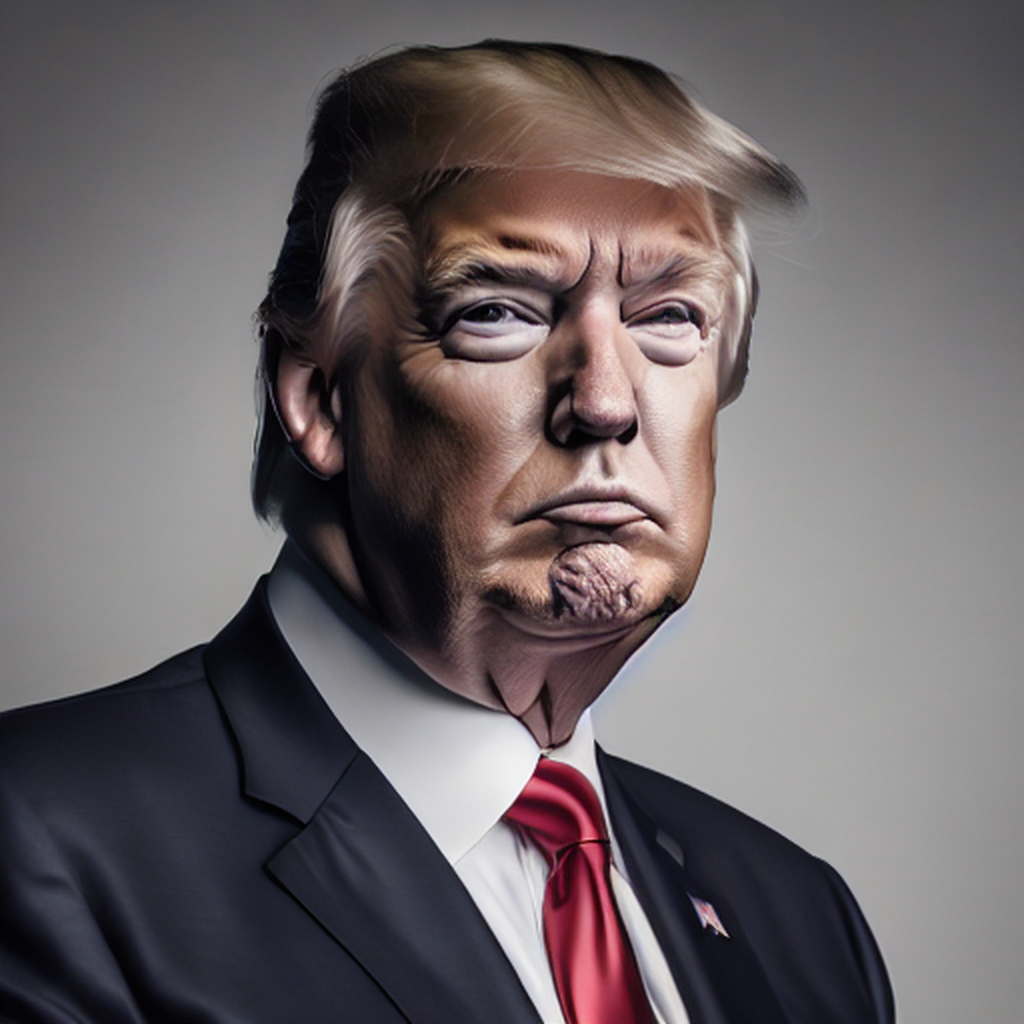} &
    \includegraphics[width=\linewidth]{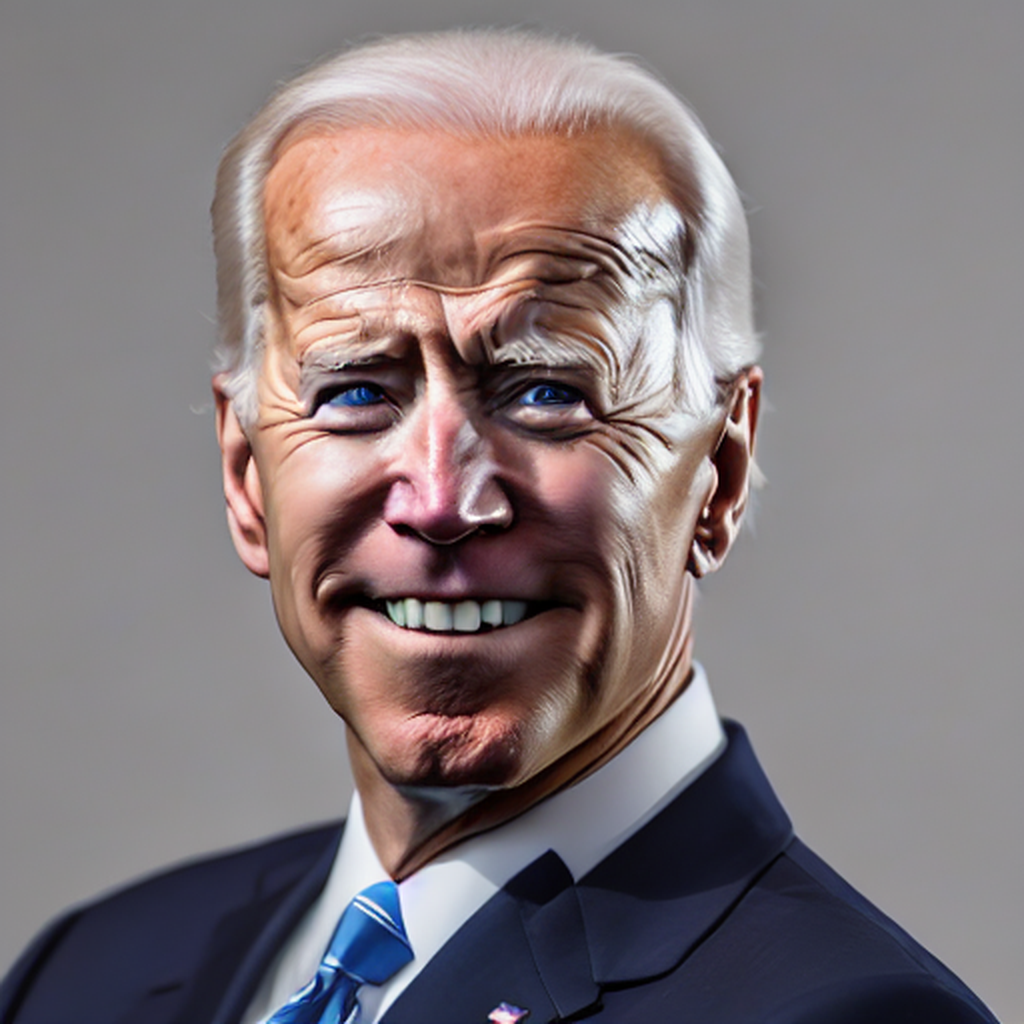} \\
    
    \midrule
    \parbox[c]{\linewidth}{\centering \uce} &
    \includegraphics[width=\linewidth]{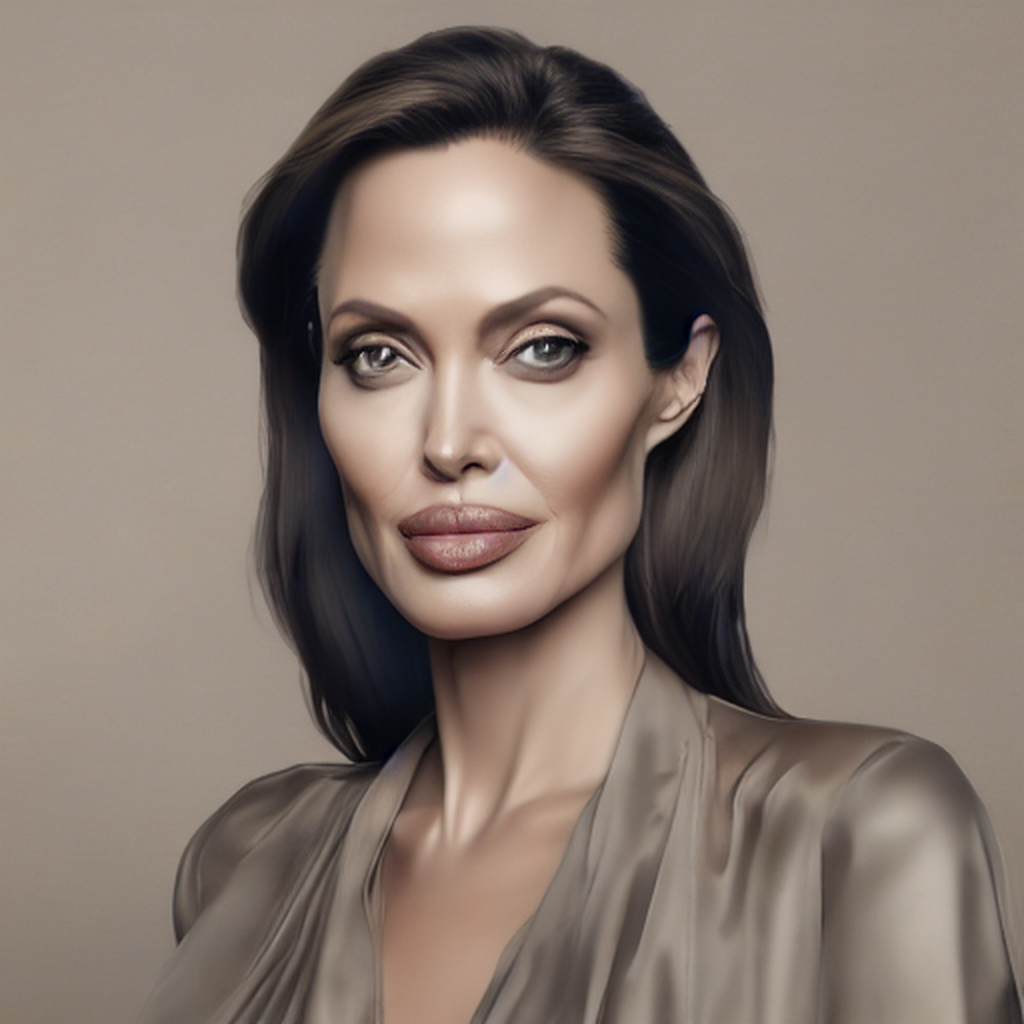} &
    \includegraphics[width=\linewidth]{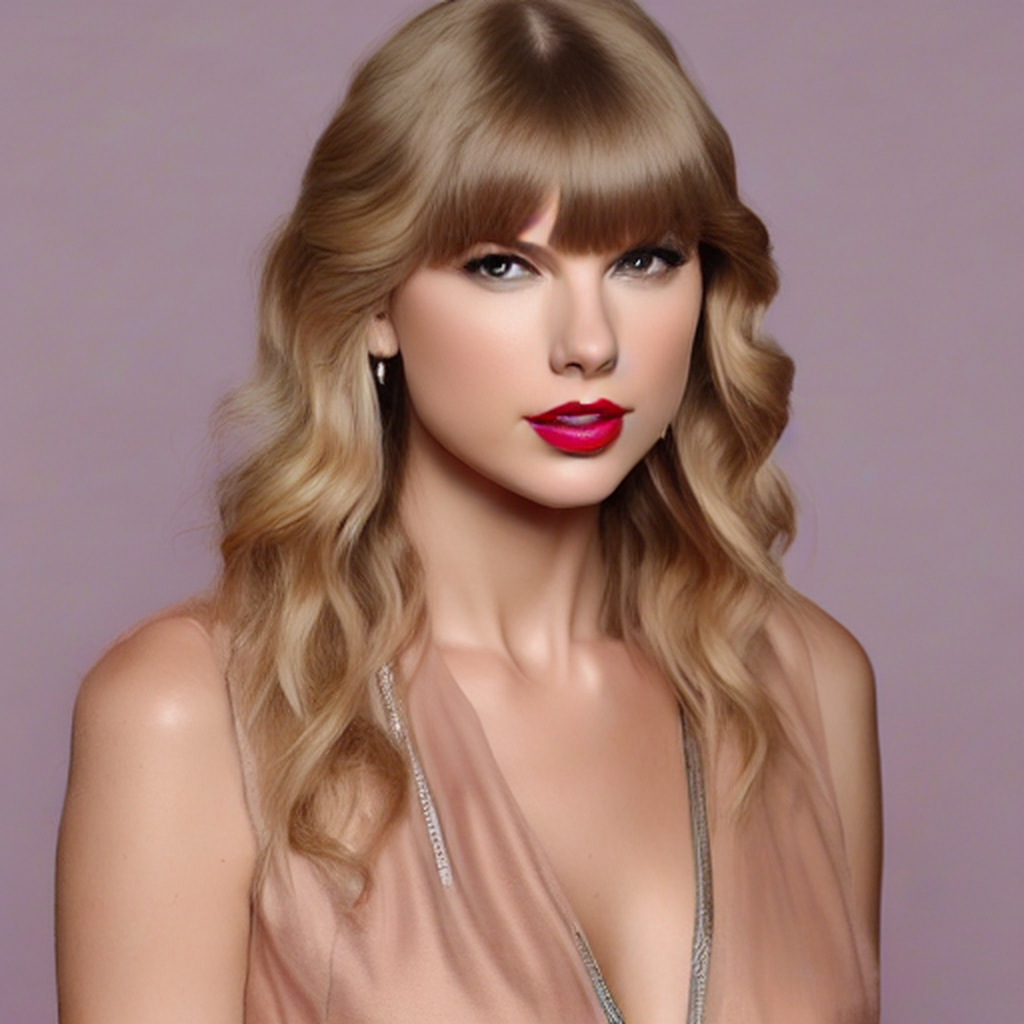} &
    \includegraphics[width=\linewidth]{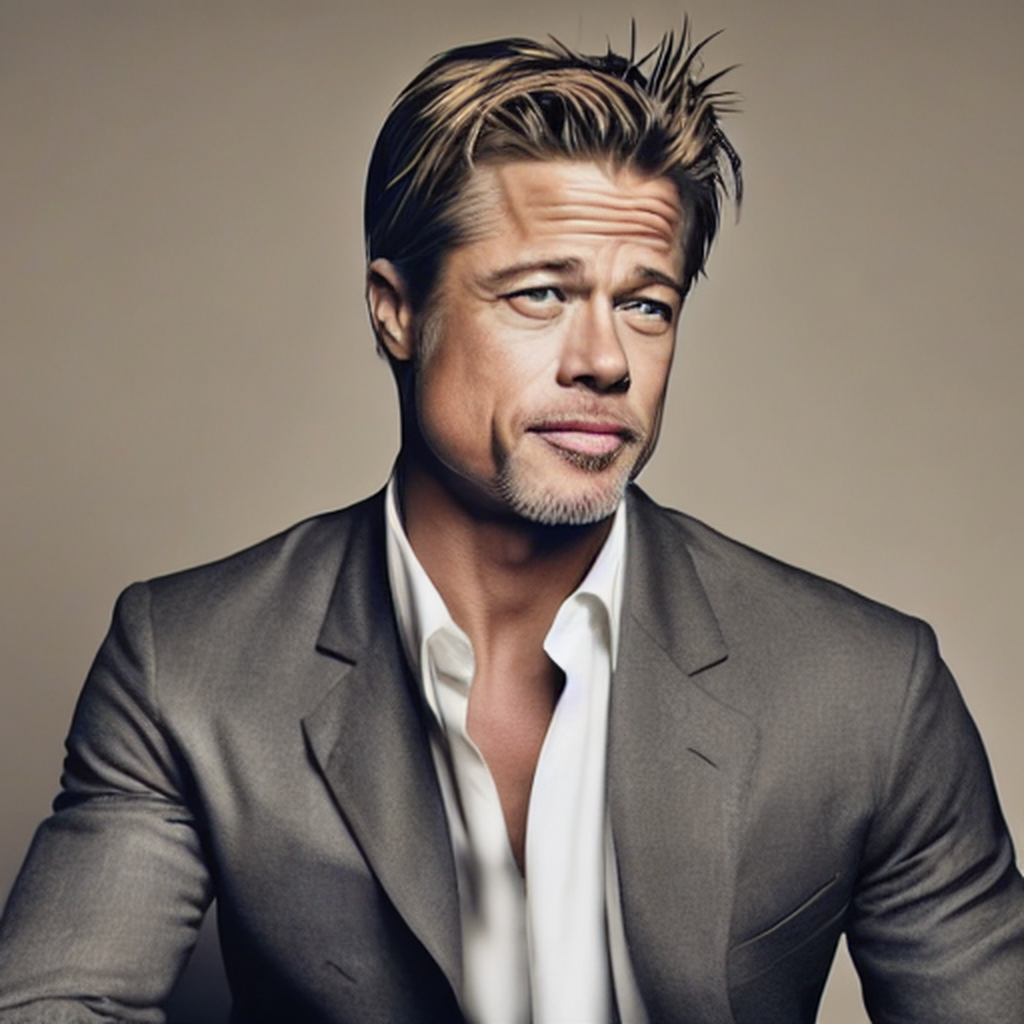} &
    \includegraphics[width=\linewidth]{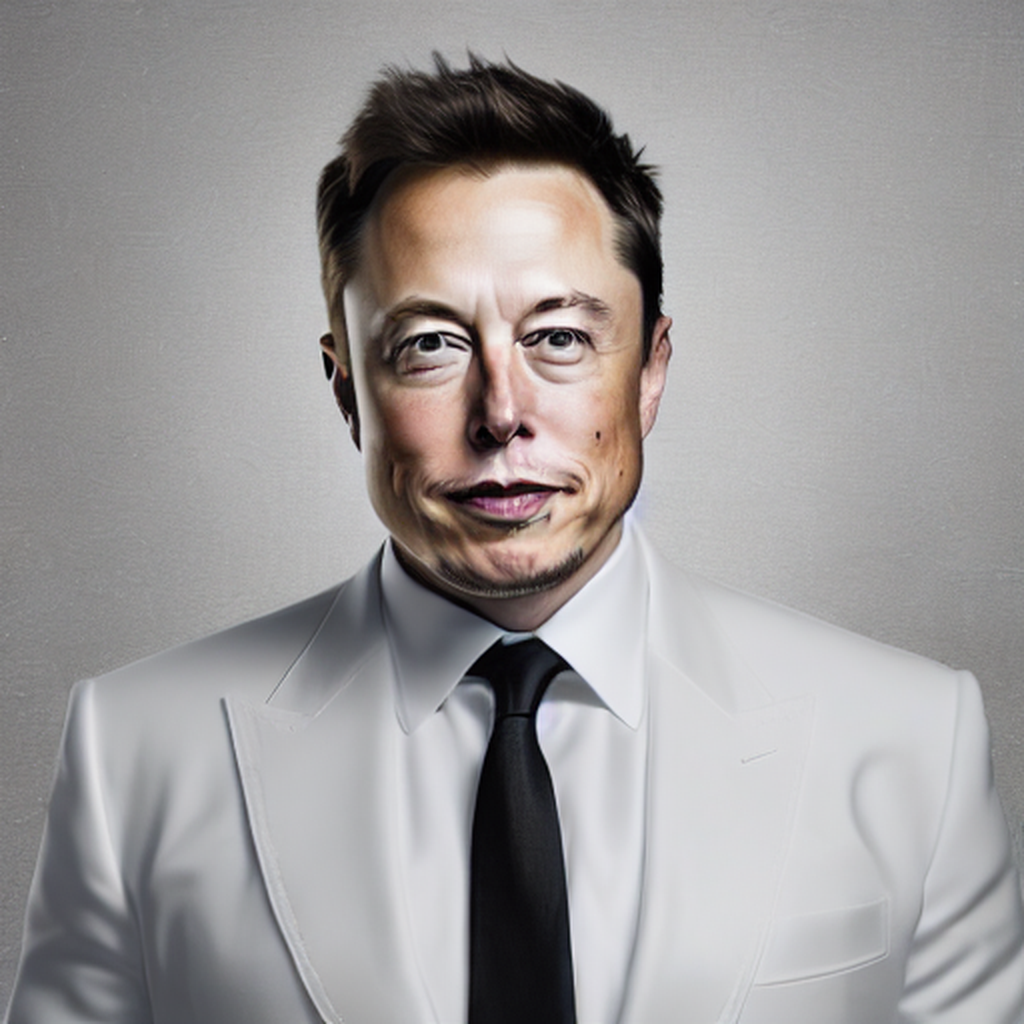} &
    \includegraphics[width=\linewidth]{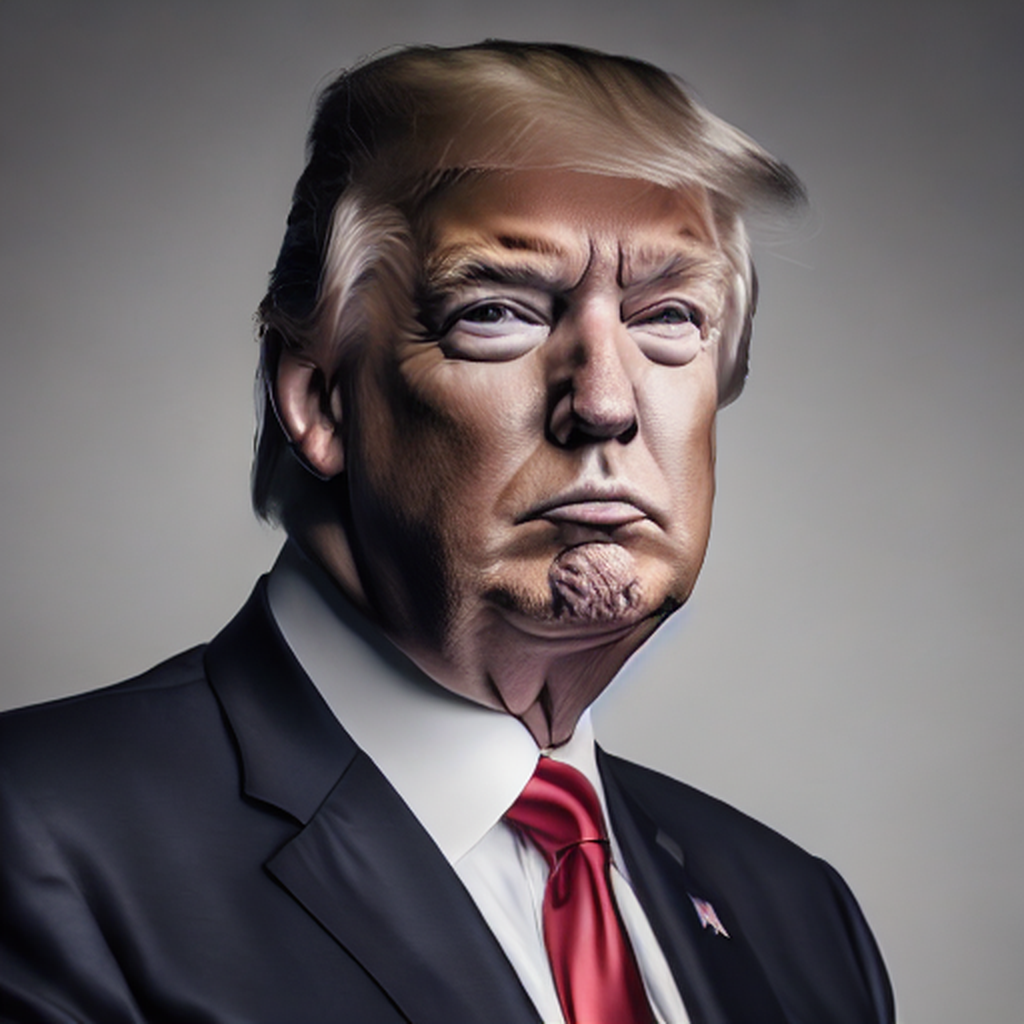} &
    \includegraphics[width=\linewidth]{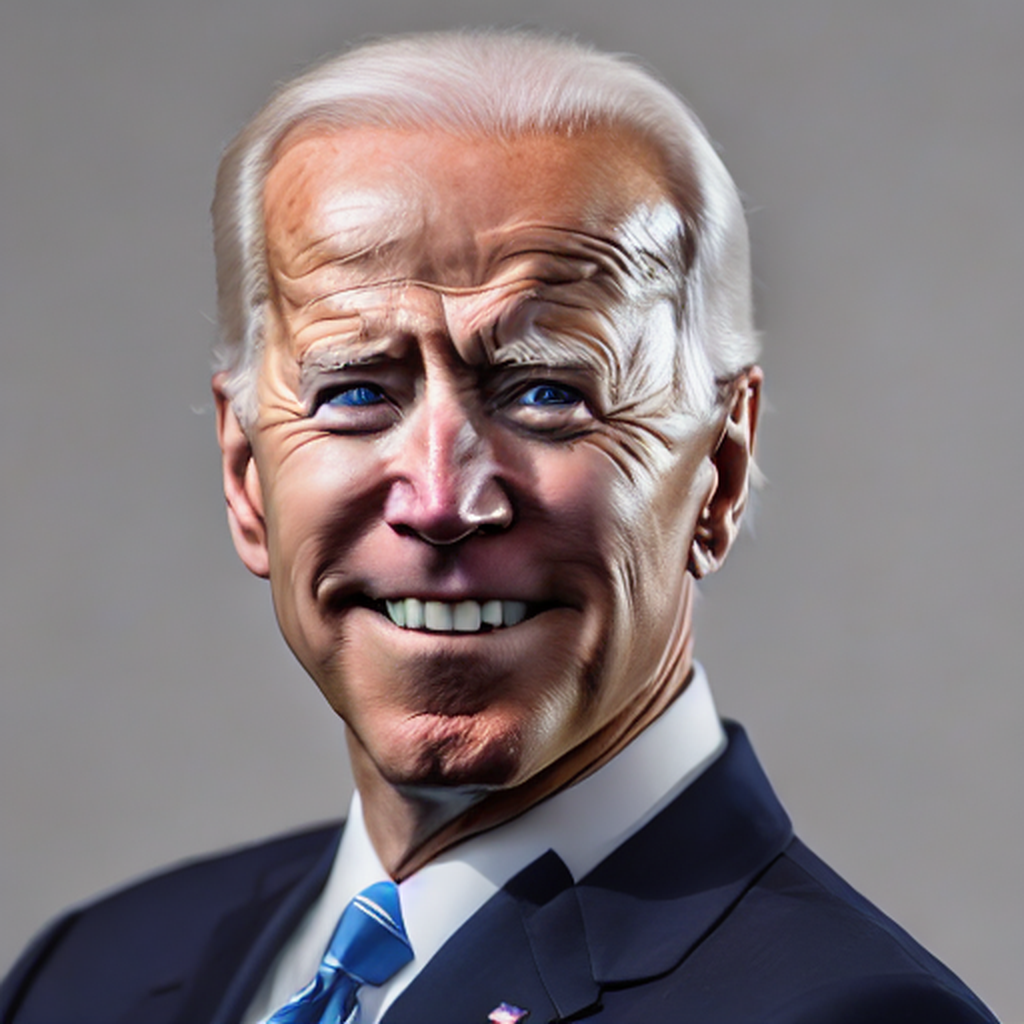} \\
    
    \bottomrule
\end{tabular}
\end{table*}

%% file: 04defense_approach.tex
\section{\crt using Editing}\label{sec:approach_defense}

\input{figures/figure_sdedit_pipline}

\input{tables/sdedit_prompt}

We have shown that \crt{s} attempting to erase concepts from the weights of diffusion models can fail to prevent unacceptable concept generation in the \iti setting. Therefore, we opt to explore different \crt approaches. One alternative approach is to replace unacceptable concepts by directly editing the output images to make them acceptable~\cite{yuan2025promptguard,park2024duo} (see~\autoref{inversion-methods}). This can be done either as pre-processing or post-processing. Since these methods do not involve modifying the \iti model, they are agnostic to the underlying \iti model. However, as we will explore, using these techniques to develop a good \crt is non-trivial.


\subsection{\sdedit as an Editing Technique}\label{sec:sdedit}

Among the methods proposed in the literature for image editing, \sdedit~\cite{meng2022sdedit} has garnered attention for its effectiveness in executing concept replacement tasks by modifying images according to provided text prompts (see ~\autoref{fig:sdedit_pipeline}). Some prior works have used \sdedit to generate training data, specifically by editing unacceptable images to get their acceptable counterparts~\cite{park2024duo,yang2024guardti}. Thus, a strawman approach using \sdedit involves substituting celebrity identities with a suitable replacement specified via text. For instance, the unacceptable concept \emph{Joe Biden} can be feasibly substituted by the more general and acceptable concept \emph{a man}. However, while \sdedit is effective in concept replacement, it may inadvertently alter other acceptable aspects of an image. Such unintended changes are particularly problematic in \iti settings where precise preservation of acceptable features is critical. To minimize these unintended effects, prompts must be highly detailed and precise. Table~\ref{tab:sdedit_prompts_images} illustrates this by showing various outputs from \sdedit with increasingly detailed prompts. Each successive prompt becomes more specific and reduces unintended alterations by focusing primarily on changing identity-related aspects (in this case, the face) to that of a random individual. This underscores the importance of crafting meticulous and specific prompts when using prompt-based editing techniques as a \crt. This can be achieved using large Vision-Language Models (VLMs) capable of analyzing images and generating appropriate prompts that describe them. However, this approach can be highly inefficient, as it substantially increases the computational load and complexity within existing \iti environments~\cite{zhang2024fastervlm}. It does not scale because an ideal text prompt needs to be crafted for each unacceptable concept. For example, an ideal prompt to replace \emph{Brad Pitt} may not be ideal to replace \emph{Angelina Jolie}. Furthermore, it has been observed that \iti models perform poorly when provided with overly detailed prompts~\cite{mou2024diffeditor}.


\subsection{Notion of {Fidelity}}\label{sec:fidelity_notion}

Maintaining acceptable concepts while removing unacceptable ones is crucial. We call this notion \emph{fidelity}. For example, for images, maintaining fidelity can correspond to preserving the background details since these are typically irrelevant to the unacceptable concept. Fidelity is important because it is difficult to distinguish between a malicious user who intentionally wants to trigger the generation of unacceptable concepts, and a benign user who triggers it accidentally.
Ideally, in the former case, we could block the entire generation outright, but in the latter case, if a user accidentally prompts the generation of unacceptable concepts, it is preferable for the model to selectively modify those concepts in the output rather than rejecting the entire response outright. Hence, we opt to preserve fidelity for both types of users rather than infer a user's intent.
To the best of our knowledge, none of the existing \crt{s} in the literature explicitly consider fidelity in corrected output images as a crucial metric.

Fidelity can be measured using \lpips. But \lpips has its own limitations: while it can be used to compare two \crt techniques, it is unclear how to use it as a standalone metric to determine whether a given \crt preserves fidelity. We use \none as the baseline. In theory, a technique whose \lpips is closer to that of \none potentially provides greater fidelity. However, since an effective \crt must remove the unacceptable concept, its \lpips will necessarily be higher than \none. Simply removing the unacceptable concept, e.g., by blacking it out, or replacing it with a random image, will result in high effectiveness but poor fidelity. Therefore, ideally, we require a \crt to output an image which is as visually close to \none as possible, while still achieving high effectiveness. In contrast, \recon is not a good measure of fidelity because it is more sensitive to changes in the pixels rather than semantic changes; even a small rotation can drastically change the reconstruction error between two images.


\subsection{Drawbacks of Editing-based \crt{s}}
We saw in \autoref{sec:sdedit} that editing-based \crt{s} require highly detailed prompts in order to preserve fidelity. Further, the prompts need to be tailored for each unacceptable concept, hampering scalability. If it was possible to identify the essential characteristics of an unacceptable concept, then the editing technique can ``surgically" alter only those characteristics in order to simultaneously achieve both fidelity and effectiveness.


Prior research has demonstrated the effectiveness of targeted editing techniques to localize and manipulate immoral visual content in \tti models~\cite{park2024localization}. Their work focuses on replacing concepts such as nudity, violence, and illegal or controversial acts. These approaches successfully blur or replace objectionable content while preserving the overall structure and semantics of the original image. This can also be adapted to \iti setting.  However, despite these advances, \emph{identity replacement} (specifically targeting recognizable celebrity faces or likenesses) remains underexplored. In this work, we address this gap by proposing a targeted editing \crt designed specifically for the removal of celebrity likenesses from generated images. Unlike previous methods that aim to replace a variety of concepts, our approach focuses on preventing potentially unwanted depictions of individual identities while preserving other visual details in the image.

\input{figures/figure_antimirror_pipeline}

\section{Targeted Editing \crt: \method}\label{sec:arch}

Drawing from prior research that identifies specific facial features essential for altering identity, without drastically affecting appearance~\cite{jose2020face,clare2018face,rezaei2021face}, we propose a targeted editing method \method, explicitly designed for fidelity-preserving celebrity replacement. \method systematically adjusts specific facial attributes, such as nose alignment, bone structure, lip shape, and eye dimensions, which collectively ensure that the resulting images visually resemble the original inputs while sufficiently replacing the original identities. 

Our proposed approach functions as a post-processing step and comprises four major components: \emph{Face Extraction}, \emph{Unacceptable Concept Check}, \emph{Mask Generation \& Mask Editing}, and \emph{Face Blending} (see~\autoref{fig:antimirror_pipeline}). As the unacceptable concepts targeted in our work are exclusively celebrity-likenesses, \method restricts editing exclusively to facial regions. This significantly enhances fidelity by preserving the background and other characteristics of the image. The extracted facial images then undergo an unacceptable-concept-detection step to ensure that editing only occurs if a celebrity identity is detected. This is implemented using a state-of-the-art detector, Espresso \cite{das2024espressorobustconceptfiltering}. Once identified, a mask delineating the aforementioned facial regions is generated and subsequently edited. The final facial image is reconstructed using CelebHQ-based mask editing~\cite{lee2020maskgan}, followed by blending with the original background to enhance fidelity, with no unacceptable concepts present. We describe each component of the pipeline in detail below:

\noindent\textbf{Face Extraction.} 
Among several available methods in the literature, we utilize the \texttt{facenet\_pytorch}\footnote{\url{https://github.com/timesler/facenet-pytorch}} library to detect and extract faces from images. Mathematically, the \iti output image \(\xxout\) is segmented into a face region \(\xface\) and a background region \(\xback\), with minimal overlap to ensure that editing operations are restricted strictly to \xface.

\noindent\textbf{Unacceptable Concept Check.} We employ Espresso~\cite{das2024espressorobustconceptfiltering}, which efficiently identifies specific celebrity identities within images and produces a binary output \(\espresso(\xface)\) indicating the presence (1) or absence (0) of a celebrity identity, $\cunacc$. This ensures that editing only occurs when necessary.

\noindent\textbf{Mask Generation and Mask Editing.} Upon detecting the unacceptable concept, we generate a facial feature mask \(M\) using the CelebHQ segmentation module, identifying regions such as eyes \(m_{eyes}\), nose \(m_{nose}\), chin \(m_{chin}\), and lips \(m_{lips}\). The system includes a graphical user interface (GUI) that allows users to directly modify the mask through annotation using a trackpad. Since manual GUI-based edits are not feasible in the context of (automatic) context replacement, we automate modifications to these masks using morphological operations (such as dilation) and geometric transformations (using the \texttt{cv2} library\footnote{\url{https://github.com/opencv/opencv}}). Specifically, dilation is applied to masked regions to slightly expand or alter their boundaries, while geometric transformations adjust the bone structure and alignment of facial features. The edited mask \(M'\) and the original facial image \(\xface\) are then passed to a trained UNet, producing a modified image \(\xface'\).

\noindent\textbf{Face Blending.} Finally, the edited facial image \(\xface'\) is seamlessly integrated with the original background \(\xback\) via Poisson blending by solving the Poisson equation:
\begin{equation}
    \min_f \iint_{\Omega}\left|\nabla f - \nabla \xface'\right|^2 \quad \text{with} \quad f|_{\partial \Omega} = \xback|_{\partial \Omega},
\end{equation}
where \(\Omega\) denotes the facial region, and \(\partial \Omega\) represents the boundary pixels. This blending approach ensures a high-quality reconstruction that preserves fidelity.

This proposed approach, by design, operates independently of input image specifics, ensuring broad applicability. Unlike \sdedit, which requires a carefully constructed prompt explicitly tailored to the input image, \textit{our proposed \method~does not depend on prompts}. Therefore, \method~is applicable to any celebrity likeness (and more generally, any context where the unacceptable concept consists of the identity of a specific person). Furthermore, since \method does not attempt to erase unacceptable concepts by modifying the weights of the diffusion model, it avoids the challenges associated with identifying and altering the correct weights — a task shown to be difficult in~\autoref{ineff_finetune}.
This systematic pipeline effectively replaces $\cunacc$ with minimal image alteration, achieving the desired balance between effectiveness and fidelity.



%% file: figures/figure_sdedit_pipline.tex
\begin{figure*}[!t]
  \centering
  \includegraphics[width=0.85\linewidth]{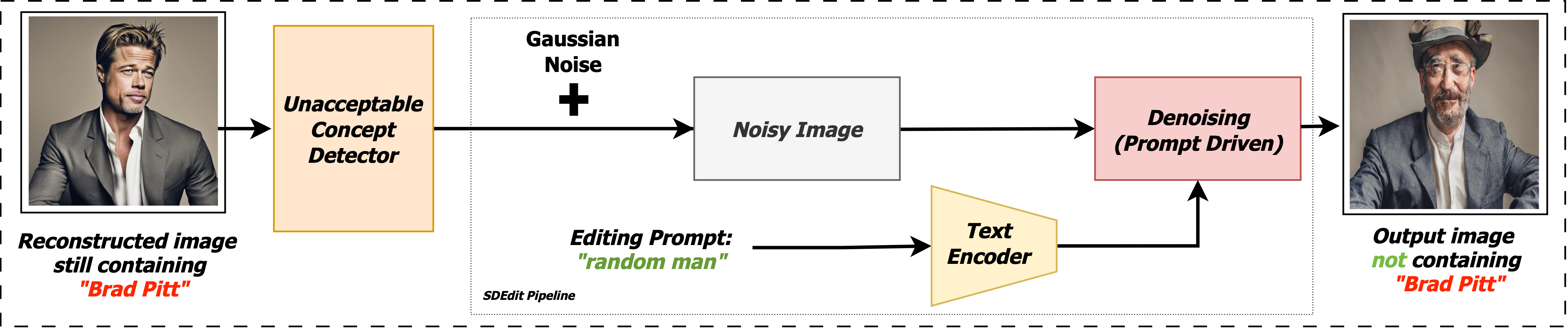}
  \caption{Concept replacement using an image editing technique (\sdedit). When the unacceptable concept detector flags a reconstructed image, Gaussian noise is added to the image as part of the \sdedit mechanism. Concurrently, an editing prompt—tailored to the original input—is encoded through a text encoder. This encoded prompt, coupled with the noisy image, guides the denoising process to produce a final output image free of unacceptable concept.
  }
  \label{fig:sdedit_pipeline}
\end{figure*}

%% file: tables/sdedit_prompt.tex
\begin{table*}[!t]
    \centering
    \caption{\sdedit as the \crt using Different Prompts}
    \label{tab:sdedit_prompts_images}
    \setlength{\tabcolsep}{6pt}
    \renewcommand{\arraystretch}{1.2}
    \begin{tabular}{m{0.4\linewidth}  >{\centering\arraybackslash}m{0.18\linewidth} >{\centering\arraybackslash}m{0.18\linewidth}}
        \toprule
        \textbf{Prompt} & \textbf{Input Image} & \textbf{Output Image} \\
        \midrule

        \small\textbf{Prompt 1:} Random man. &
        \includegraphics[width=0.8\linewidth]{images/pitt/final/original.png} &
        \includegraphics[width=0.8\linewidth]{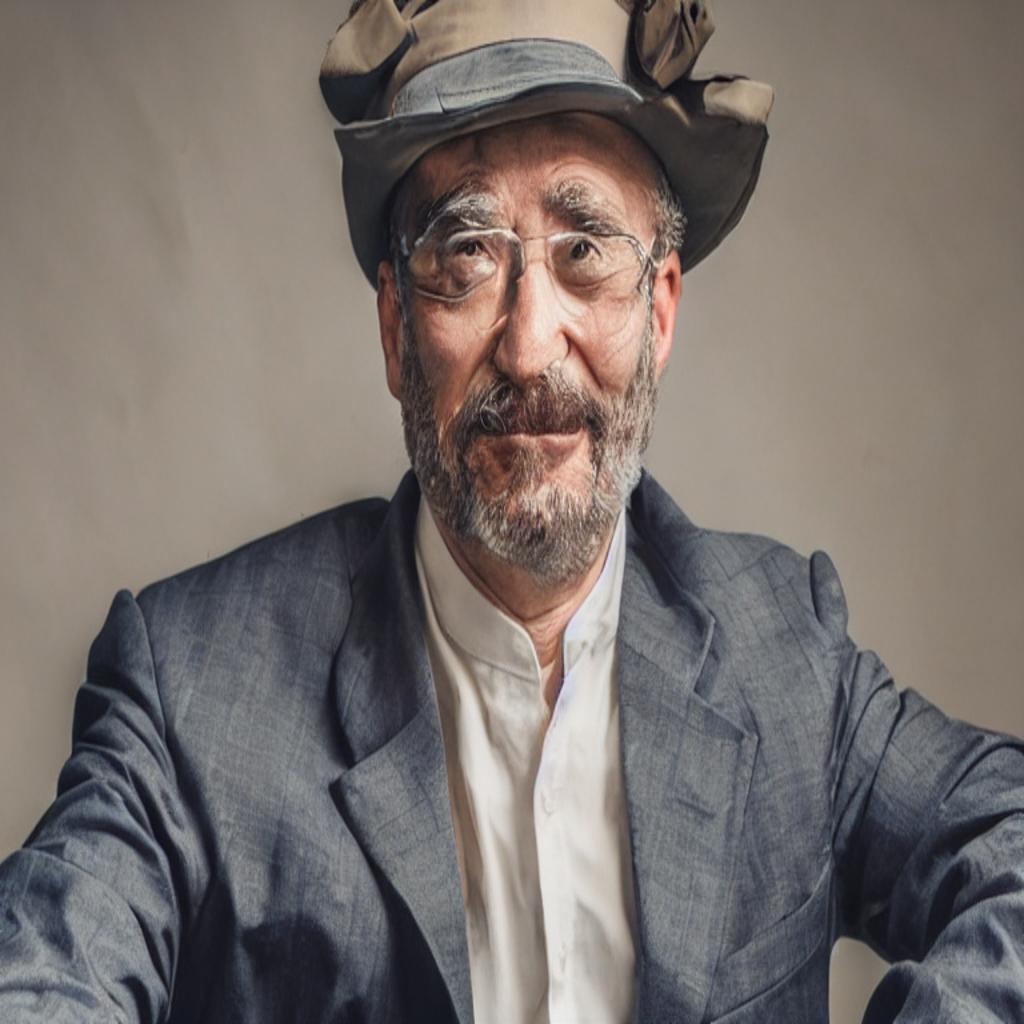} \\
        \midrule

        \small\textbf{Prompt 2:} Random man, keep the exact pose, clothing, hair color and style, lighting, overall scene composition, and image quality identical to the original. &
        \includegraphics[width=0.8\linewidth]{images/pitt/final/original.png} &
        \includegraphics[width=0.8\linewidth]{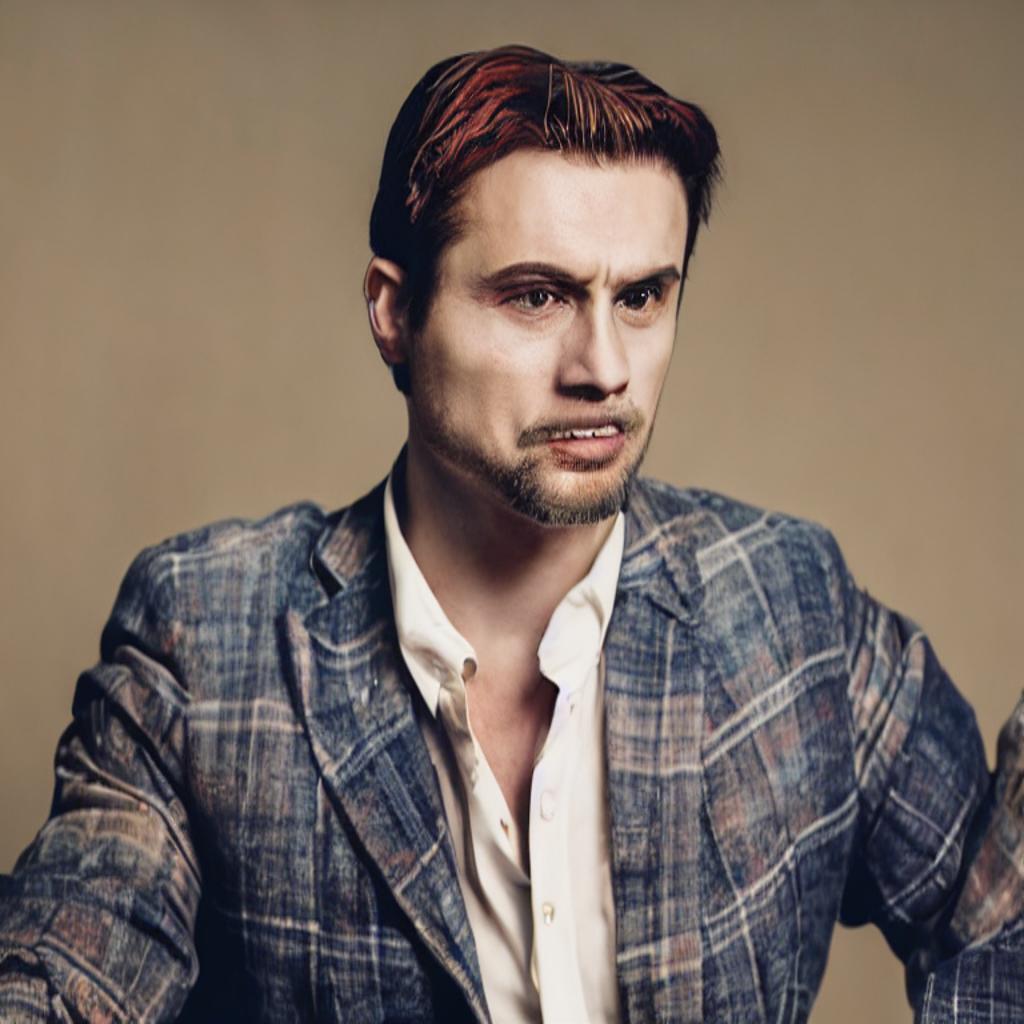}\\
        \midrule

        \small\textbf{Prompt 3:} Photorealistic portrait of a well-dressed man wearing a sleek gray suit and white shirt in a neutral studio background. Keep the exact pose, clothing, hair color and style, lighting, overall scene composition, and image quality identical to the original. &
        \includegraphics[width=0.8\linewidth]{images/pitt/final/original.png} &
        \includegraphics[width=0.8\linewidth]{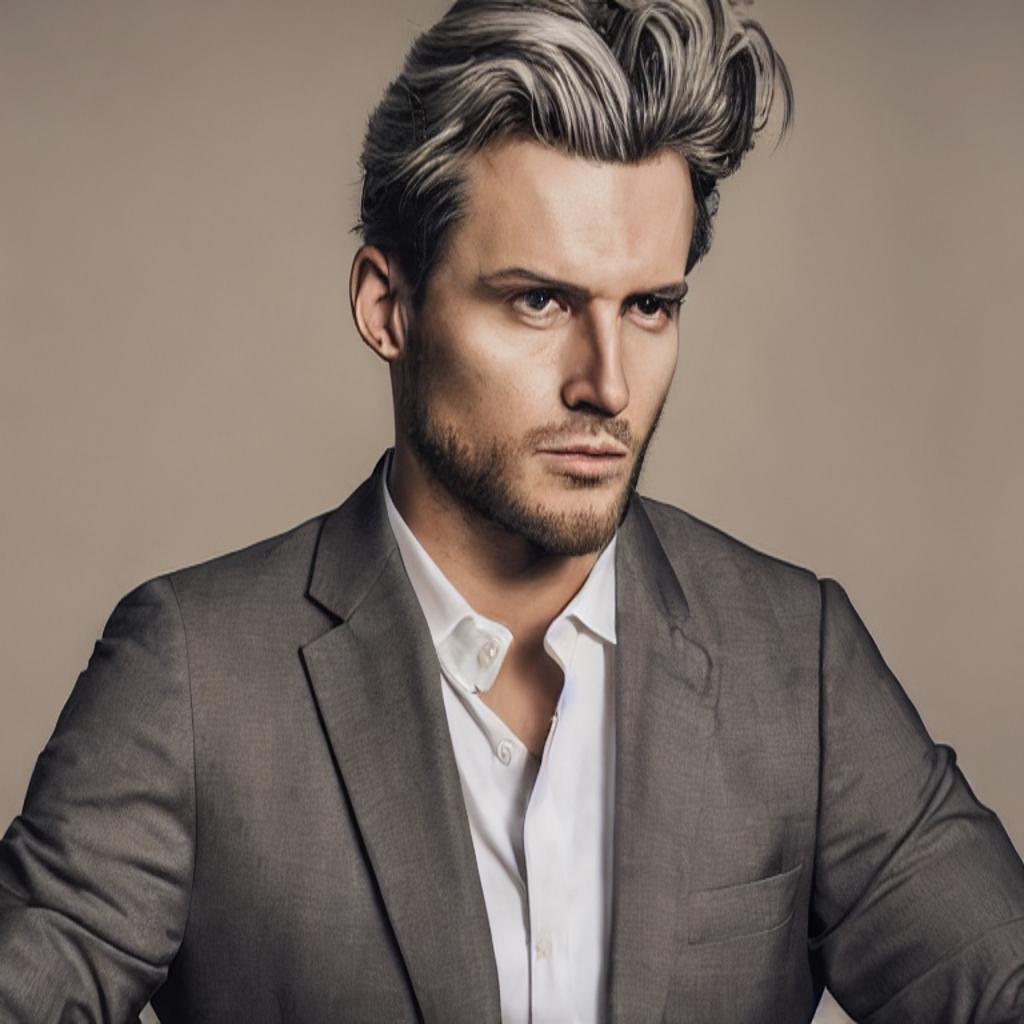} \\
        
        \bottomrule
    \end{tabular}
\end{table*}

%% file: figures/figure_antimirror_pipeline.tex
\begin{figure*}[!t]
  \centering
  \includegraphics[width=\linewidth]{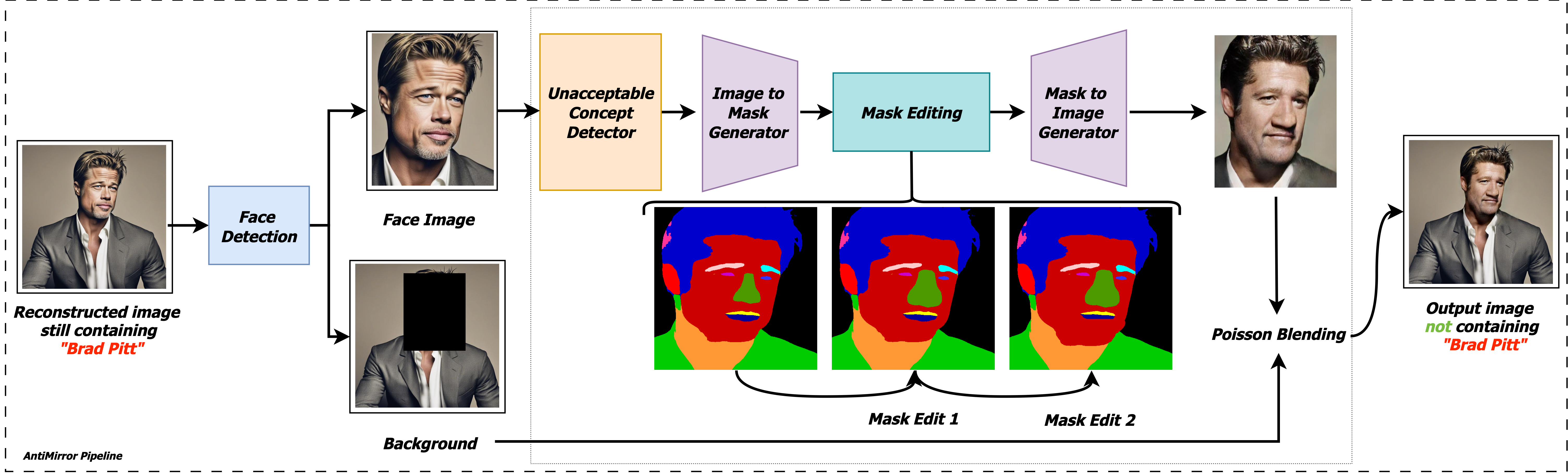}
  \caption{Concept replacement with a targeted editing approach. When a face is detected in the reconstructed image, it is isolated from the background, and passed through an unacceptable concept detector. If the detector flags the image, an image-to-mask generator constructs a mask, which is then edited to modify the face. The modified face is then composited back into the background using Poisson blending. 
  }
  \label{fig:antimirror_pipeline}
\end{figure*}

%% file: 05results_defense.tex
\section{\method Results}\label{sec:defense_results}
We first use \clipscore and \lpips scores to evaluate the effectiveness and fidelity, respectively, of \method and \sdedit. 
In order to use \sdedit\ as \crt, we employed specific prompts for celebrities. For instance, for all male celebrities ({\emph{Brad Pitt}, \emph{Elon Musk}, \emph{Donald Trump}, \emph{Joe Biden}}), we used the editing prompt ``random man'', and for female celebrities ({\emph{Angelina Jolie}, \emph{Taylor Swift}}), we used ``random woman''. This strategy was adopted so that \sdedit\ could replace the unacceptable concept with an acceptable one.

\input{figures/figure_bar_plots_crt}

We present the results in Table~\ref{tab:clip_comparison_ours} \&~\ref{rec_ddpm_ours}.

\subsection{Effectiveness}\label{sec:antimirror_effective}
 \sdedit exhibits a high degree of effectiveness, as demonstrated by \clipscore values near 0.15 which are significantly below the threshold of 0.25. This clearly validates the effectiveness of \sdedit. Conversely, \method yields \clipscore scores comparable to the \none baseline, suggesting limited effectiveness in concept removal.

\input{tables/clip_new_defense}

We revisit the suitability of \clipscore scores as a metric for effectiveness of \crt{s}. The higher \clipscore scores for our targeted editing method can be attributed to the inherent characteristics of \clipscore itself. \clipscore score measure semantic alignment based on global image-text correspondence and are trained to capture broad conceptual coherence~\cite{wang25global}. In our targeted editing approach, only specific features within the image are altered rather than the image as a whole. Consequently, \clipscore may yield higher scores for targeted edits of celebrity likenesses because targeted edits \emph{intentionally} leave features unrelated to personal identity intact (e.g., a person's typical clothing). A fair metric of \crt effectiveness must be capable of capturing this aspect.

Finding such a metric is difficult in general. However, one possibility can be specialized classifiers that are trained to detect an unacceptable concept (thus ignoring irrelevant features). For celebrity likenesses, such a classifier exists: the Giphy Celebrity Detector\footnote{\url{https://github.com/Giphy/celeb-detection-oss}}. It is a multi-class classifier to identify celebrities in images and has been used in prior work~\cite{lu2024mace}. A GCD top-1 accuracy (\gcdd) of 0 denotes 0\% accuracy in detecting a celebrity, and 1 denotes 100\% accuracy. An effective \crt should have low \gcdd for its outputs.  
With this new effectiveness metric, we repeated the reconstruction experiment in \autoref{sec:erasing_experiment} for celebrity concepts, but we evaluated the output using \gcdd, in Table~\ref{tab:gcd_crt}.

The outcomes are consistent with those obtained using the \clipscore score (see~\autoref{sec:recon}); since the \gcdd values are consistently greater than 0.8, it confirms that the unacceptable concepts remain present, and the state-of-the-art \crt{s} are ineffective.

\input{tables/gcd_crt}

\input{tables/gcd_new_defense}

\input{tables/new_defense}

\input{tables/celeb_images}
Having established \gcdd as a good metric, we use it to compare \method and \sdedit (Table~\ref{tab:gcd_comparison_ours}) showing that \method is significantly effective in removing celebrity likenesses (compared to the baseline \none) and it is comparable to \sdedit in most cases. The only case where \method is significantly less effective than \sdedit is \emph{Angelina Jolie}. But even in this case, it significantly outperforms other state-of-the-art \crt{s} from Table~\ref{tab:gcd_crt} (lowest \gcdd = 0.86). Therefore, we conclude that \method is an effective \crt.


\subsection{Fidelity}\label{sec:antimirror_fidelity}
As we discussed in~\autoref{sec:fidelity_notion}, \lpips can be used to compare the fidelity of two \crt{s}. Since \none will yield a small \lpips score indicating minimal change, the fidelity of two \crt{s} can be compared based on how close their \lpips scores are to the \none baseline.
The fidelity achieved by \method is notably superior to that of \sdedit (see Table~\ref{rec_ddpm_ours}) since the \lpips scores for \method are consistently lower than or comparable to \sdedit.
Furthermore,~\autoref{fig:bar_plots_all} shows that \method achieves a better balance between fidelity and effectiveness; the \lpips bars for \method are consistently lower than those for \sdedit, while the \gcdd bars for \method and \sdedit are comparable in most cases.




\subsection{Visual Examples Supporting Fidelity of \method}

In~\autoref{tab:targeted_editing_recon_images}, we present visual examples to illustrate how targeted editing in \method modifies only the necessary regions—specifically selected facial characteristics—to remove identity, without affecting other acceptable concepts. 
The examples also illustrate how \sdedit sacrifices fidelity for effectiveness, while \method balances both.

For instance, in the image of \emph{Joe Biden}, \sdedit not only replaces the identity but also unnecessarily alters the background, hairstyle, and tie color—elements that are acceptable and should not have been changed. In contrast, \method preserves them, demonstrating superior fidelity while still replacing the celebrity identity.


\sdedit transforms images of \emph{Taylor Swift} and \emph{Angelina Jolie} into hand-sketched cartoon-like versions. In contrast, \method carefully retains characteristics unrelated to identity while effectively replacing those that do. 

Remarkably, in the case of \emph{Brad Pitt}, \sdedit introduces unintended artifacts, such as a hat, exaggerates aging effects, closes the eyes, and alters the posture. In contrast, \method precisely preserves such aspects, demonstrating superior fidelity.

Overall, these visual comparisons clearly illustrate that \method offers significant advantages over \sdedit by effectively replacing unacceptable celebrity features while preserving essential, acceptable visual details—thereby ensuring both effectiveness and high fidelity.

%% file: figures/figure_bar_plots_crt.tex
\begin{figure*}[!t]
  \centering
  \includegraphics[width=\linewidth]{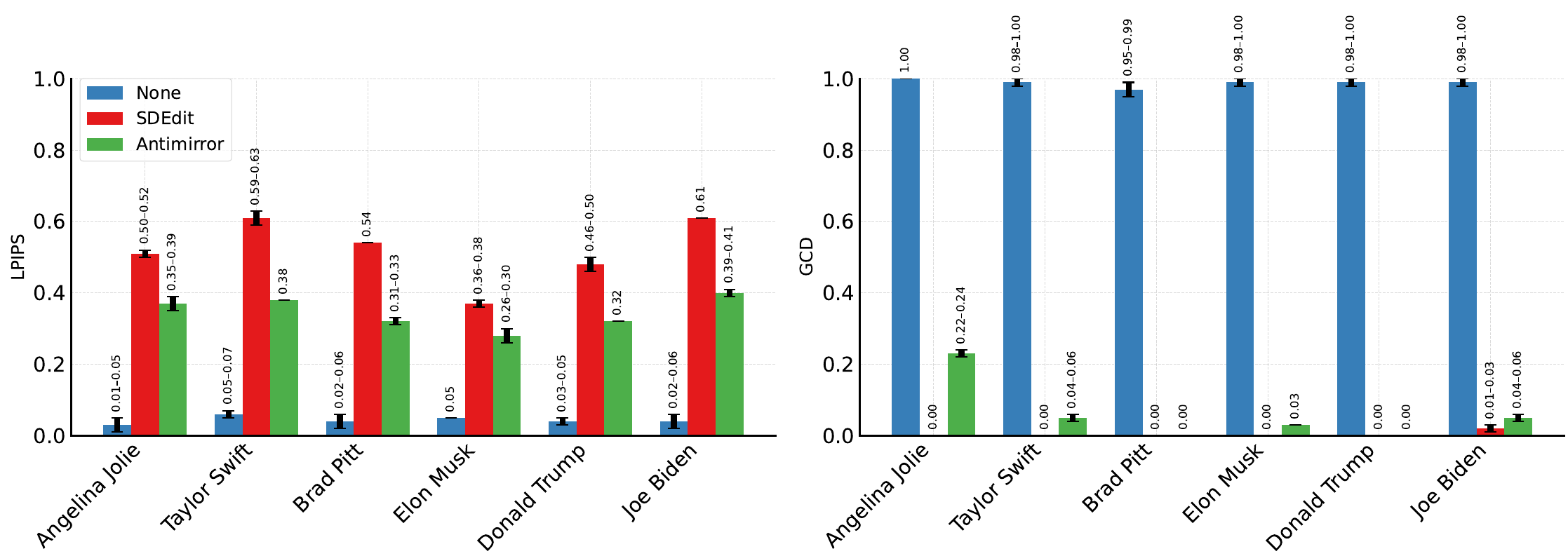}
  \caption{{Bar plots comparing \none, \sdedit, and \method using \lpips and \gcdd. A smaller value in the \lpips plot (left) indicates higher fidelity and a smaller value in the \gcdd plot (right) indicates higher effectiveness. Our method achieves a better trade-off, balancing identity replacement (effectiveness) with fidelity.}}
  \label{fig:bar_plots_all}
\end{figure*}

%% file: tables/clip_new_defense.tex
\begin{table}[htbp]
  \centering
  \small
  \setlength{\tabcolsep}{5pt}
  \caption{Effectiveness at removing celebrity likenesses for \none (baseline with no \crt applied), \sdedit, and \method. \clipscore scores above 0.25 indicate higher semantic equivalence. We bold the columns with the lowest \clipscore score.}
  \label{tab:clip_comparison_ours}
  \begin{tabular}{l ccc}
    \toprule
    Celebrity & \none & \sdedit & \method \\
    \midrule
    \textit{Angelina Jolie} & $0.25 \pm 0.02$ & \textbf{0.13} $\pm$ \textbf{0.03} & $0.24 \pm 0.01$ \\
    \textit{Taylor Swift}   & $0.23 \pm 0.02$ & \textbf{0.11} $\pm$ \textbf{0.01} & $0.22 \pm 0.01$ \\
    \textit{Brad Pitt}      & $0.26 \pm 0.02$ & \textbf{0.12} $\pm$ \textbf{0.02} & $0.23 \pm 0.02$ \\
    \textit{Elon Musk}      & $0.26 \pm 0.02$ & \textbf{0.17} $\pm$ \textbf{0.02} & $0.27 \pm 0.01$ \\
    \textit{Donald Trump}   & $0.24 \pm 0.01$ & \textbf{0.15} $\pm$ \textbf{0.02} & $0.25 \pm 0.01$ \\
    \textit{Joe Biden}      & $0.24 \pm 0.02$ & \textbf{0.17} $\pm$ \textbf{0.01} & $0.27 \pm 0.01$ \\
    \bottomrule
  \end{tabular}
  \vspace{3pt}
\end{table}

%% file: tables/gcd_crt.tex
\begin{table}[htbp]
  \centering
  \footnotesize
  \setlength{\tabcolsep}{4pt} 
  \caption{Validating \gcdd as a metric for \crt effectiveness for celebrity likenesses}
  \label{tab:gcd_crt}

   \begin{tabular}{l cccc}
    \toprule
    Celebrity & \none & \moderator & \mace & \uce \\

    \midrule
    \textit{Angelina Jolie} & \(0.92 \pm 0.01\) & \(0.86 \pm 0.02\) & \(1.00 \pm 0.02\) & \(1.00 \pm 0.01\) \\
    \textit{Taylor Swift}   & \(0.94 \pm 0.01\) & \(1.00 \pm 0.01\) & \(0.98 \pm 0.02\) & \(0.88 \pm 0.03\) \\
    \textit{Brad Pitt}      & \(0.94 \pm 0.02\) & \(0.99 \pm 0.01\) & \(0.99 \pm 0.02\) & \(0.99 \pm 0.03\) \\
    \textit{Elon Musk}      & \(0.84 \pm 0.03\) & \(0.94 \pm 0.01\) & \(1.00 \pm 0.00\) & \(1.00 \pm 0.01\) \\
    \textit{Donald Trump}   & \(0.89 \pm 0.00\) & \(0.96 \pm 0.03\) & \(0.96 \pm 0.01\) & \(0.90 \pm 0.01\) \\
    \textit{Joe Biden}      & \(0.84 \pm 0.01\) & \(0.96 \pm 0.01\) & \(0.97 \pm 0.01\) & \(0.86 \pm 0.02\) \\
    \bottomrule
  \end{tabular}
\end{table}

%% file: tables/gcd_new_defense.tex
\begin{table}[htbp]
  \centering
  \small
  \setlength{\tabcolsep}{5pt} 
  \caption{\gcdd comparison across celebrities when no \crt is applied (\none), and when \sdedit or \method are used. \gcdd identifies celebrities in images by measuring top-1 accuracy, with values between 0 (no celebrity detected, best effectiveness) and 1 (celebrity detected, no effectiveness). We underline the columns where \method and \sdedit have comparable \gcdd values.}
  \label{tab:gcd_comparison_ours}
  \begin{tabular}{l ccc}
    \toprule
    Celebrity & \none & \sdedit & \method \\
    \midrule
    \textit{Angelina Jolie} & $1.00 \pm 0.00$ & $0.00 \pm 0.00$ & 0.23 $\pm$ 0.01 \\
    \textit{Taylor Swift}   & $0.99 \pm 0.01$ & $0.00 \pm 0.00$ & 0.05 $\pm$ 0.01 \\
    \textit{Brad Pitt}      & $0.97 \pm 0.02$ & $0.00 \pm 0.00$ & \underline{0.00 $\pm$ 0.00} \\
    \textit{Elon Musk}      & $0.99 \pm 0.01$ & $0.00 \pm 0.00$ & 0.03 $\pm$ 0.00 \\
    \textit{Donald Trump}   & $0.99 \pm 0.01$ & $0.00 \pm 0.00$ & \underline{0.00 $\pm$ 0.00}\\
    \textit{Joe Biden}      & $0.99 \pm 0.01$ & $0.02 \pm 0.01$ & 0.05 $\pm$ 0.01 \\
    \bottomrule
  \end{tabular}
  \vspace{3pt}
\end{table}

%% file: tables/new_defense.tex
\begin{table}[htbp]
  \centering
  \small
  \setlength{\tabcolsep}{5pt} 
  \caption{\lpips comparison across celebrities when no \crt is applied (\none), and when \sdedit or \method are used. \lpips score ranges from 0 (identical) to 1 (maximal difference).}
  \begin{tabular}{l ccc}
    \toprule
    Celebrity & \none & \sdedit & \method \\
    \midrule
    \textit{Angelina Jolie} & 0.03 $\pm$ 0.02 & 0.51 $\pm$ 0.01 & \textbf{0.37} $\pm$ \textbf{0.02} \\
    \textit{Taylor Swift}   & 0.06 $\pm$ 0.01 & 0.61 $\pm$ 0.02 & \textbf{0.38} $\pm$ \textbf{0.00} \\
    \textit{Brad Pitt}      & 0.04 $\pm$ 0.02 & 0.54 $\pm$ 0.00 & \textbf{0.32} $\pm$ \textbf{0.01} \\
    \textit{Elon Musk}      & 0.05 $\pm$ 0.00 & 0.37 $\pm$ 0.01 & \textbf{0.28} $\pm$ \textbf{0.02} \\
    \textit{Donald Trump}   & 0.04 $\pm$ 0.01 & 0.48 $\pm$ 0.02 & \textbf{0.32} $\pm$ \textbf{0.00} \\
    \textit{Joe Biden}      & 0.04 $\pm$ 0.02 & 0.61 $\pm$ 0.00 & \textbf{0.40}	 $\pm$ \textbf{0.01} \\
    \bottomrule
  \end{tabular}
  \vspace{3pt}
 \label{rec_ddpm_ours}
\end{table}

%% file: tables/celeb_images.tex
\begin{table*}[htbp]
\centering
\caption{Examples of reconstruction with different CRTs (\none, \sdedit, and \method)}
\label{tab:targeted_editing_recon_images}
\setlength{\tabcolsep}{2pt}
\renewcommand{\arraystretch}{1.3}
\begin{tabular}{m{1.8cm} 
                >{\centering\arraybackslash}m{0.12\linewidth} 
                >{\centering\arraybackslash}m{0.12\linewidth} 
                >{\centering\arraybackslash}m{0.12\linewidth} 
                >{\centering\arraybackslash}m{0.12\linewidth} 
                >{\centering\arraybackslash}m{0.12\linewidth} 
                >{\centering\arraybackslash}m{0.12\linewidth}}
    \toprule
    Celebrity ($\rightarrow$) & \emph{Angelina Jolie} & \emph{Taylor Swift} & \emph{Brad Pitt} & \emph{Elon Musk} & \emph{Donald Trump} & \emph{Joe Biden} \\
    \midrule
    
    \parbox[c]{\linewidth}{\centering Input} &
    \includegraphics[width=\linewidth]{images/jolie/final/original.png} &
    \includegraphics[width=\linewidth]{images/taylor/final/original.png} &
    \includegraphics[width=\linewidth]{images/pitt/final/original.png} &
    \includegraphics[width=\linewidth]{images/elon/final/original.png} &
    \includegraphics[width=\linewidth]{images/trump/final/original.png} &
    \includegraphics[width=\linewidth]{images/biden/final/original.png} \\
    
    \midrule
    \multicolumn{1}{c}{CRT~($\downarrow$)} &
    \multicolumn{6}{c}{Output} \\
    
    \midrule
    \parbox[c]{\linewidth}{\centering \none} &
    \includegraphics[width=\linewidth]{images/jolie/final/none_jolie.png} &
    \includegraphics[width=\linewidth]{images/taylor/final/none_swift.png} &
    \includegraphics[width=\linewidth]{images/pitt/final/none_pitt.png} &
    \includegraphics[width=\linewidth]{images/elon/final/none_musk.png} &
    \includegraphics[width=\linewidth]{images/trump/final/none_trump.png} &
    \includegraphics[width=\linewidth]{images/biden/final/none_biden.png} \\
    
    \midrule
    \parbox[c]{\linewidth}{\centering \sdedit} &
    \includegraphics[width=\linewidth]{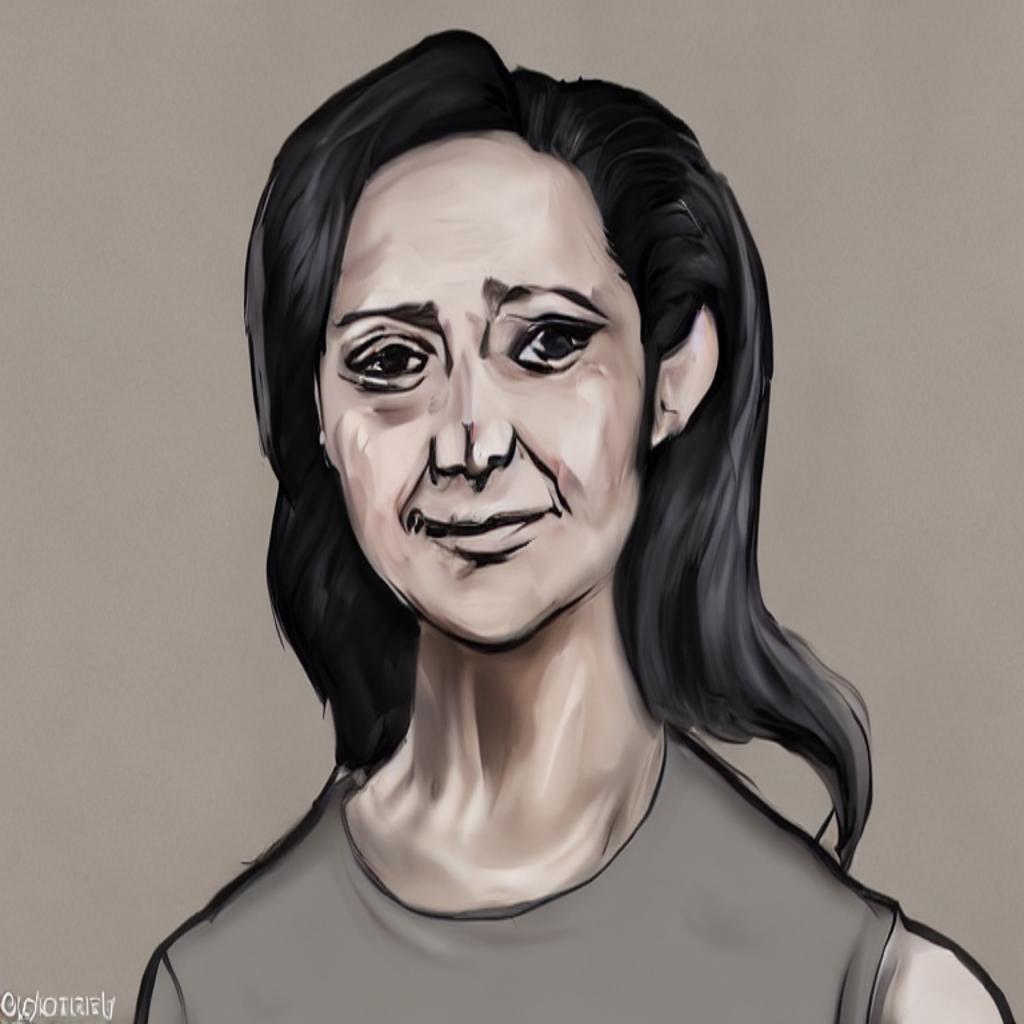} &
    \includegraphics[width=\linewidth]{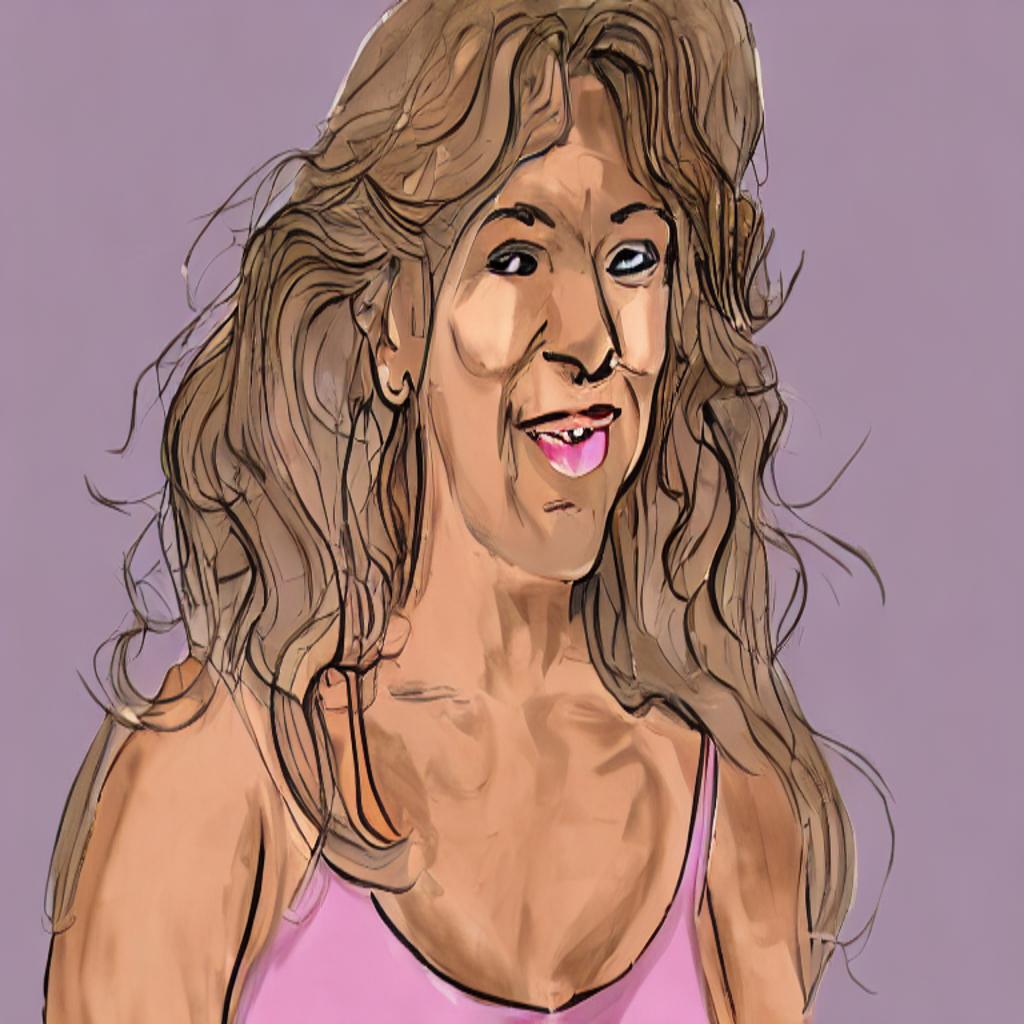} &
    \includegraphics[width=\linewidth]{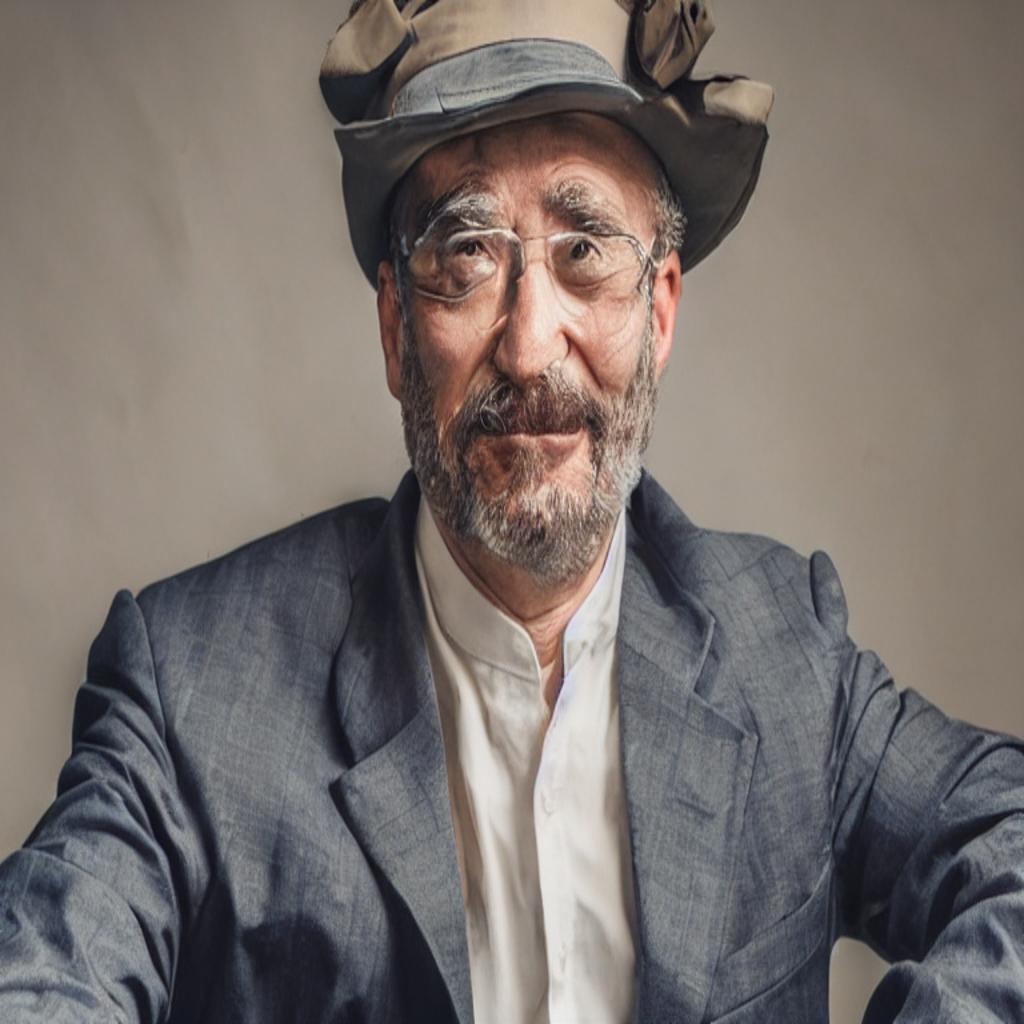} &
    \includegraphics[width=\linewidth]{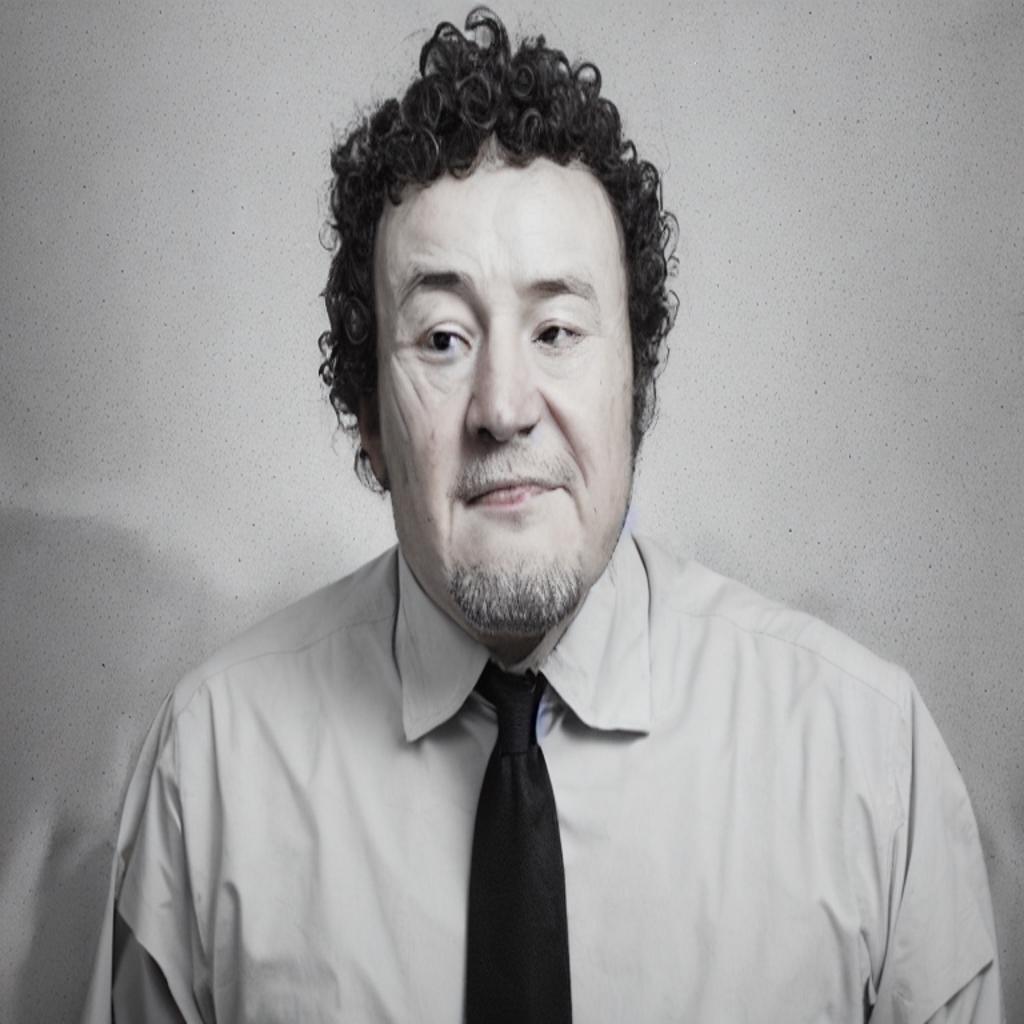} &
    \includegraphics[width=\linewidth]{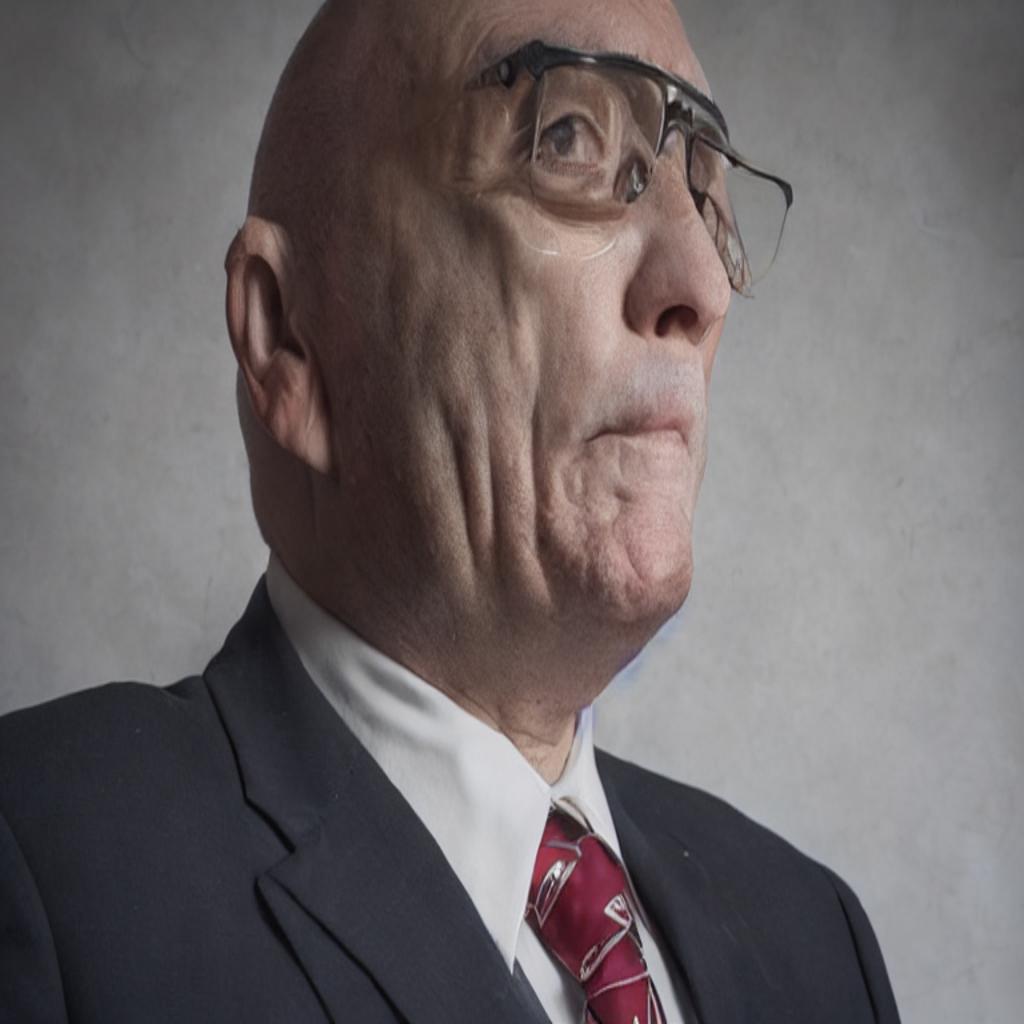} &
    \includegraphics[width=\linewidth]{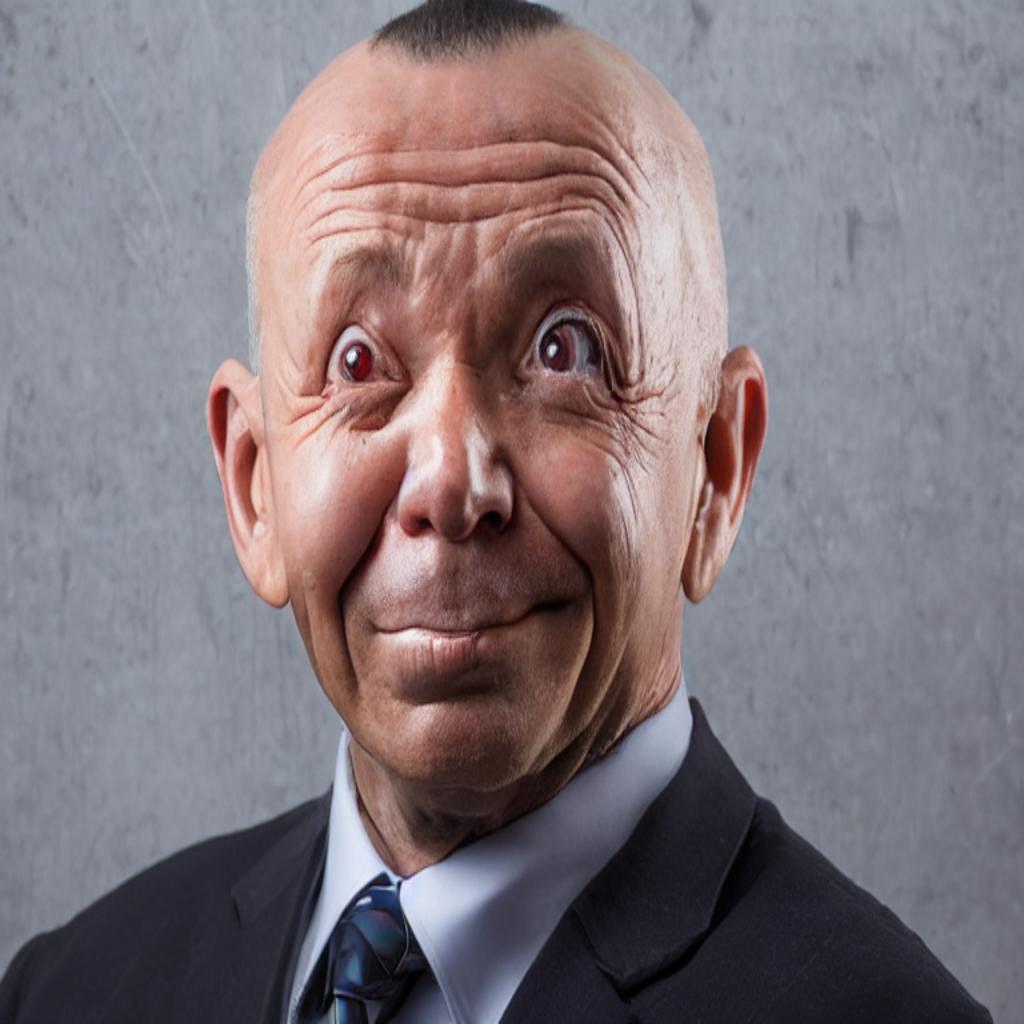} \\
    
    \midrule
    \parbox[c]{\linewidth}{\centering \method} &
    \includegraphics[width=\linewidth]{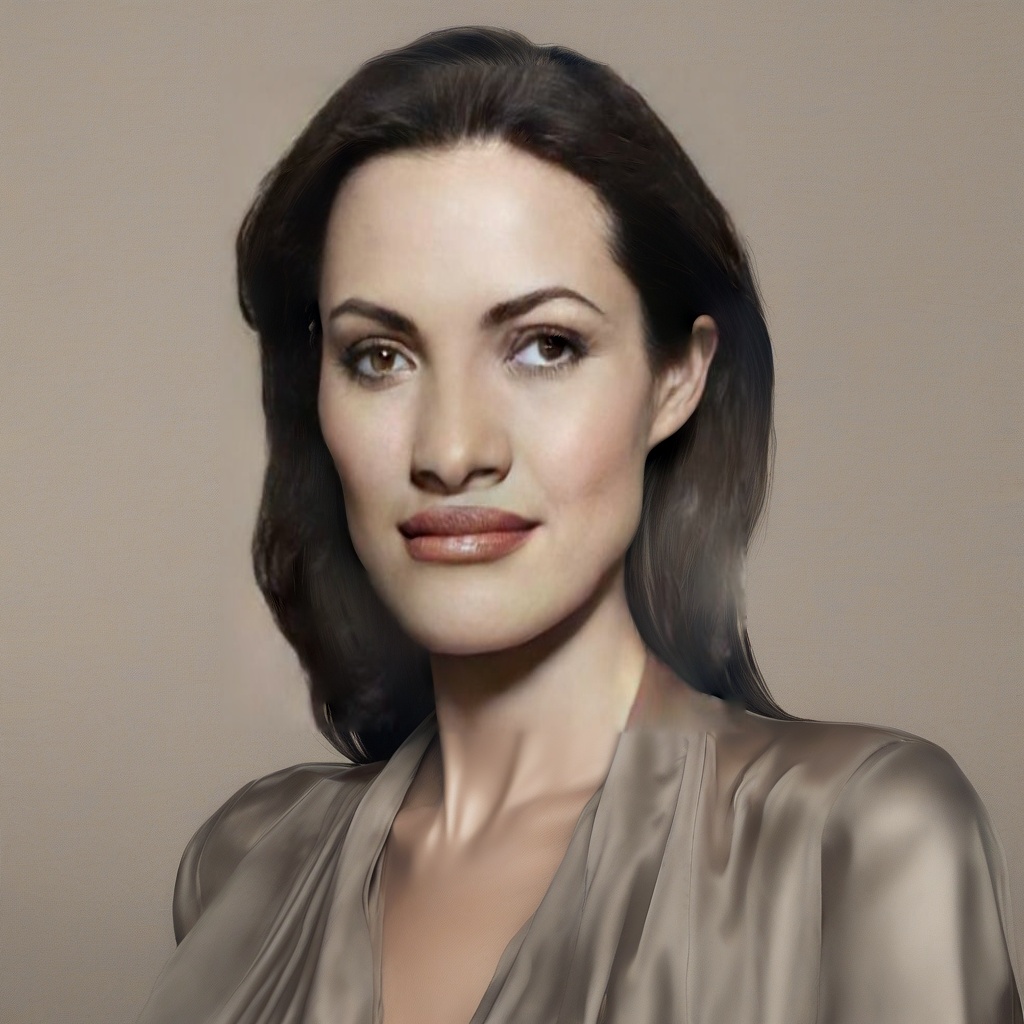} &
    \includegraphics[width=\linewidth]{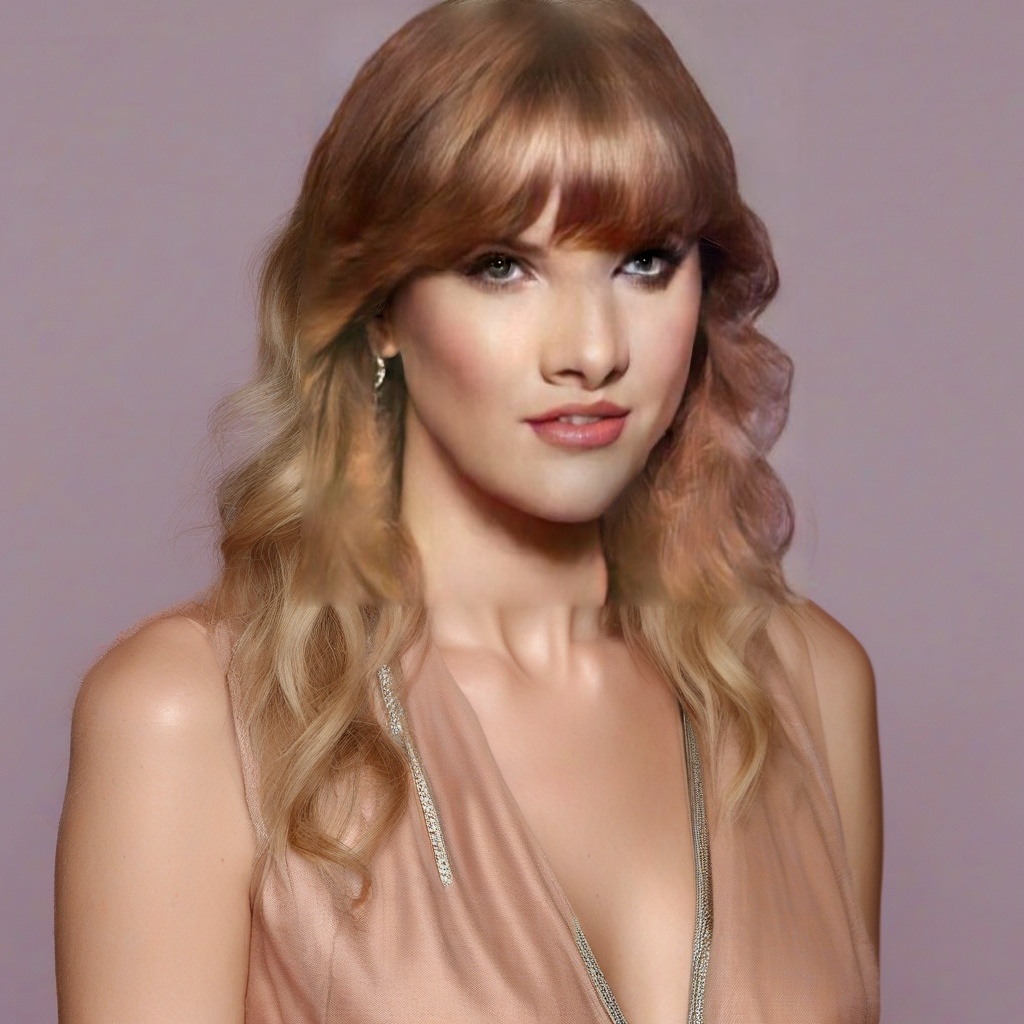} &
    \includegraphics[width=\linewidth]{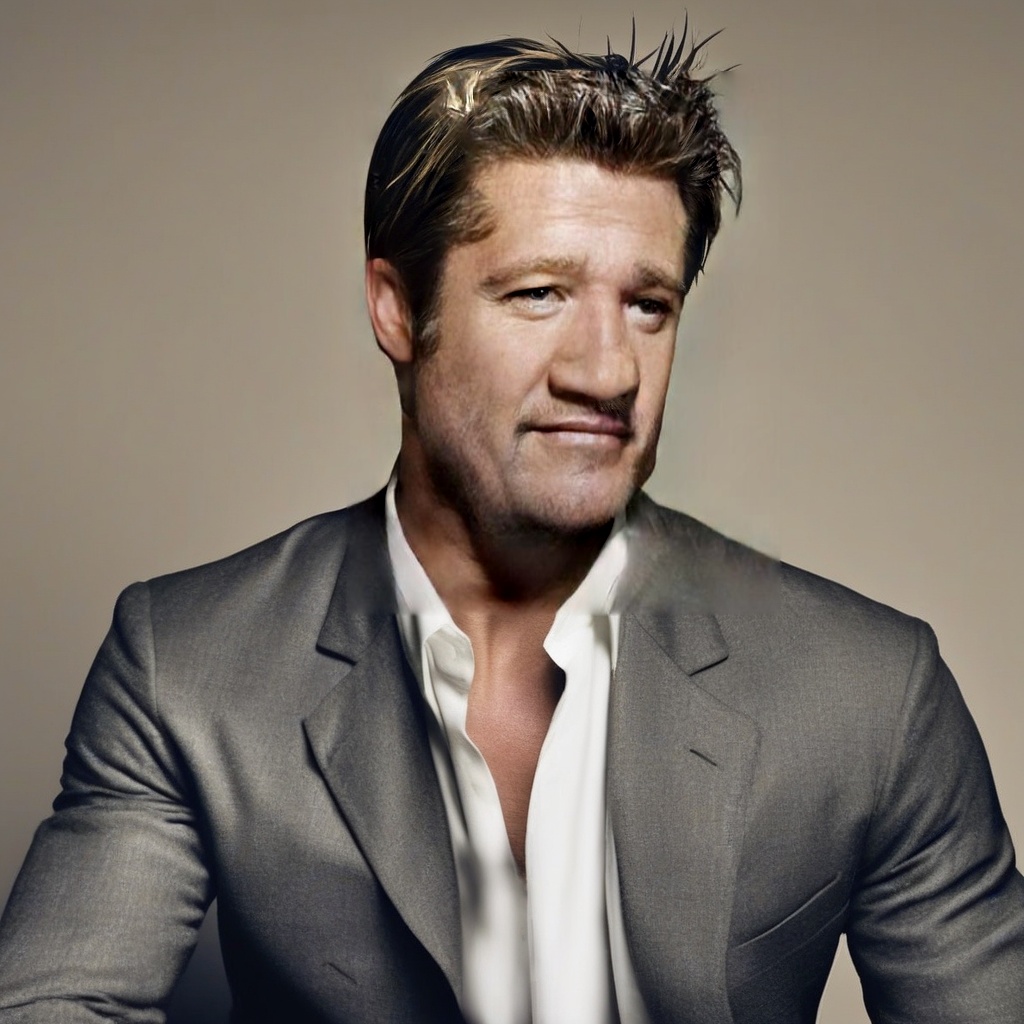} &
    \includegraphics[width=\linewidth]{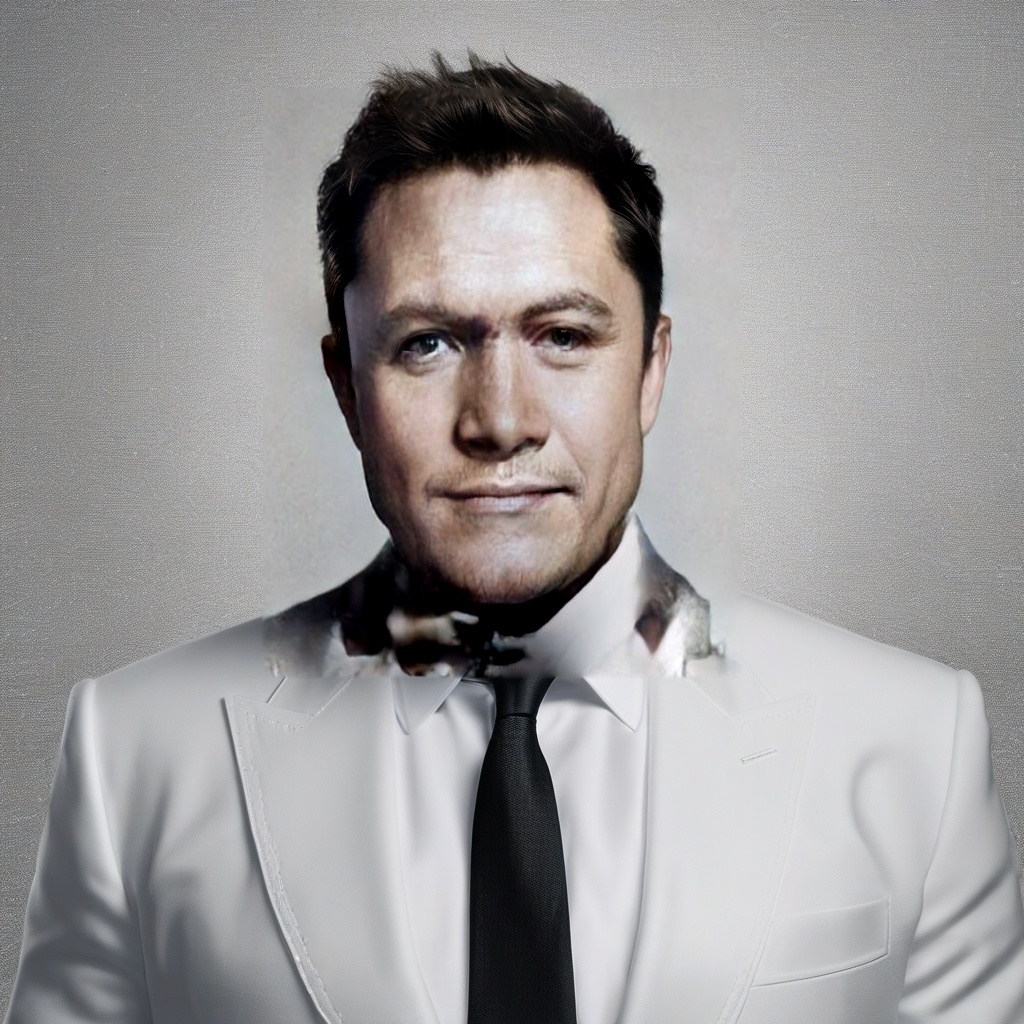} &
    \includegraphics[width=\linewidth]{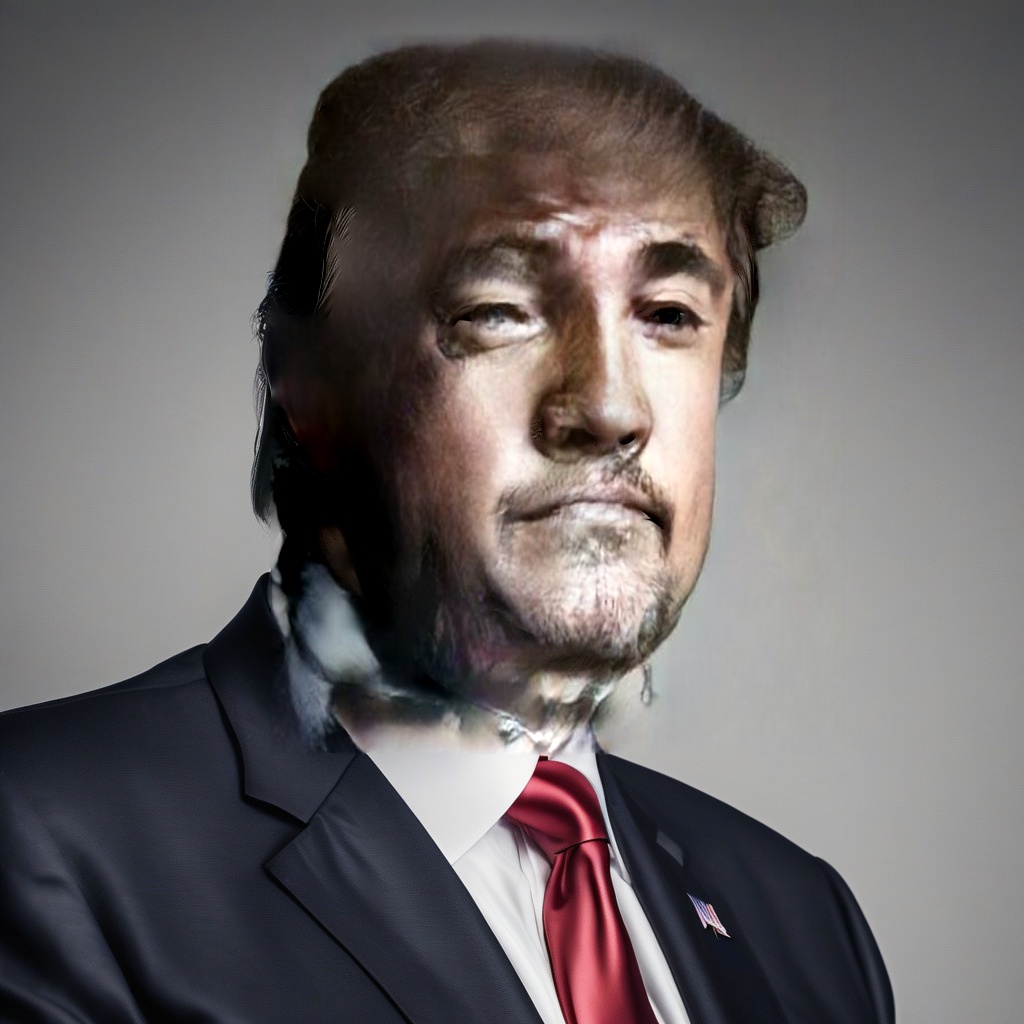} &
    \includegraphics[width=\linewidth]{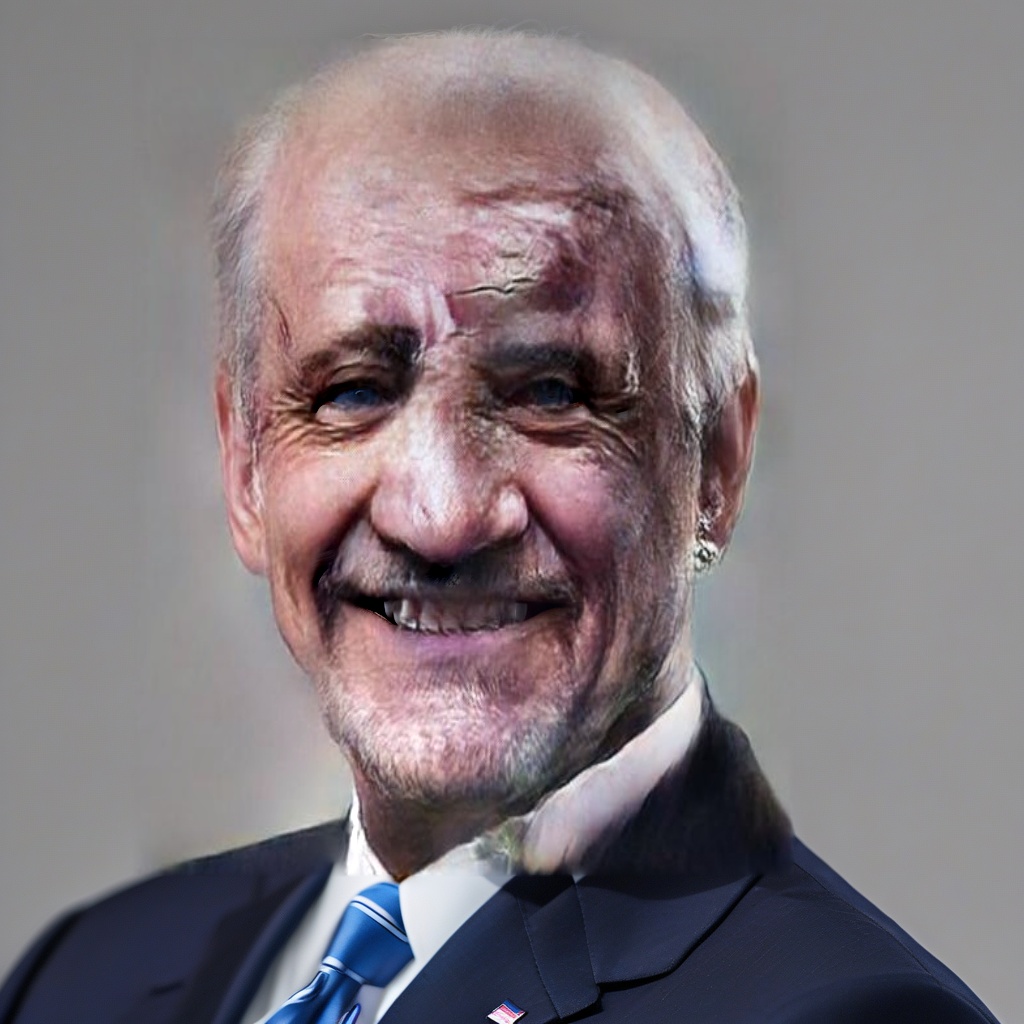} \\
    
    \bottomrule
\end{tabular}
\end{table*}

%% file: 07related_work.tex
\section{Related Work}\label{sec:related_work}
\textbf{Additional \crt{s}.} We focused on state-of-the-art \crt{s} that modified weights in the diffusion model's UNet. However, there are other \crt{s} which alter different parts of the diffusion model.
For instance, several of them remove unacceptable concepts from the \clipscore text encoder of a \tti model. In this way, if a text prompt contains an unacceptable concept, either explicitly or implicitly, it will be adjusted such that the unacceptable concept is replaced. Yoon et al.~\cite{yoon2025safree} achieve this by projecting \clipscore text embeddings away from a latent space of unacceptable concepts as a pre-processing step. Similarly, Wang et al.~\cite{wang25aeiou} and Li et al.~\cite{li2023suppress} identify and remove unacceptable concepts from the hidden states of the \clipscore text encoder when processing prompts. In contrast, Zhang et al.~\cite{zhang2024advunlearn} adversarially train the \clipscore text encoder to remove unacceptable concepts such that texts containing the concept can no longer be used to generate unacceptable images. 
We do not evaluate these \crt{s} because in our setting, the prompt supplied to an \iti model is empty ($``"$). Since these \crt{s} do not analyze any image data, they are ineffective at preventing reconstruction in the \iti setting.

In addition, there are some \crt{s} that focus on removing a single class of concepts. Park et al.~\cite{park2024duo} erase offensive concepts from the weights of the diffusion model by adapting direct preference optimization for unlearning. In contrast, Li et al.~\cite{li2024safegen} focus on modifying the self-attention layers of the UNet to remove offensive concepts. 
In \autoref{concept-removal-techniques}, we opted to analyze three state-of-the-art \crt{s}  configured to support all concept classes in this paper (offensive, copyright, and celebrity-likeness). But we confirmed that the \crt{s} in Park et al.~\cite{park2024duo} and Li et al.~\cite{li2024safegen}, built to replace offensive concepts (\emph{nudity}), also failed to erase it. Due to a lack of space, we present these results in the full version of this paper.

\textbf{Other Editing Methods.} 
There are other editing methods that can be used for inversion and reconstruction. Brack et al.~\cite{brack2024ledits} propose LEDITS++, a state-of-the-art method that, unlike DDPM inversion, solves the DDPM sampling problem using a second-order stochastic differential equation solver, DPM-Solver++~\cite{lu2022dpm}, and formulates their inversion process accordingly. We present the results of our experiment with LEDITS++ in the full version of our paper. They are similar to those from DDPM inversion; the metrics from reconstruction with aligned models were nearly identical to metrics from the unaligned model, where no \crt was used.

 
%

%% file: 08future_work.tex
\section{Applications and Extensions to \method}\label{sec:future_work}

\textbf{Filtering more Concepts.} 
Currently, \method successfully replaces celebrity concepts. In future work, we plan to extend \method to also address offensive and copyrighted concepts. This would involve isolating the regions of the image associated with these concepts, evaluating them using an unacceptable concept detector, and applying targeted editing to replace them while preserving visual fidelity. 
In general, as long as the unacceptable concept can be detected via the pixels of the image, then it can be replaced with targeted editing.

\textbf{Other Qualities of a Good \crt.}
So far, we have shown that \method is both effective and fidelity-preserving for unacceptable concepts. However, a good \crt should also \emph{preserve utility}, by allowing acceptable concepts to be generated, and be \emph{robust} against adversaries that attempt to evade it. We believe \method will also preserve utility, as it relies on both a state-of-the-art face detector and unacceptable concept detector for those faces. Furthermore, to ensure robustness against perturbation-based attacks that aim to evade \method, adversarial noise purification techniques, such as ADBM~\cite{li2025advdbm}, could be employed as a pre-processing step. 

%% file: 09conclusion.tex
\section{Conclusion}\label{sec:conclusion}

We demonstrated that concept replacement techniques (\crt{s}) that claim to erase unacceptable concepts fail to do so since images with these concepts are able to be reconstructed in the \iti setting, despite their effectiveness in the \tti setting. Motivated by this, we recognized that the fidelity of corrected unacceptable images is an important property, along with effectiveness, for a good \crt. Hence we approach developing a new \crt, \method, that is able to better balance these aspects. However, identifying a suitable metric to evaluate fidelity and effectiveness across all concepts remains challenging—highlighted by the reliance on a baseline for \lpips scores and the need for a specialized celebrity classifier when dealing with celebrity-likeness concepts. Our work represents an initial step toward addressing these challenges, laying the groundwork for future advancements. \\

%% file: 10acks.tex
\noindent\textbf{Acknowledgements:}
This work is supported in part by the Government of Ontario (RE011-038). Views expressed in the paper are those of the authors
and do not necessarily reflect the position of the funding agencies. We thank Thapar School of Advanced AI and Data Science at TIET, Patiala for allowing us to use their compute resources, and our colleagues who provided valuable feedback on previous versions of this paper (Adam Caulfield, Vasisht Duddu, Hossam ElAtali, Tony He, and Asim Waheed). 

%% file: 09appendix.tex
\appendix\label{sec:appendix}
\normalsize 
\renewcommand\thesubsection{\Alph{subsection}}



We present the results of our reconstruction experiment for the \crt{s} \park and \lietal in Table~\ref{rec_ddpm_new_crt}. These are configured for offensive concepts, hence we test on \emph{nudity} only.

\input{tables/unlearning_metrics_other_crts}
We also conduct our reconstruction experiments with the LEDITS++ inversion method and present the results in Table~\ref{rec_ledits}. Like DDPM inversion, he metrics from reconstruction
with aligned models were nearly identical to metrics from the
unaligned model, where no \crt was used.

\input{tables/all_tables}

%% file: tables/unlearning_metrics_other_crts.tex
\begin{table}[htbp]
  \centering
  \small 
  \caption{Reconstruction results when a CRT (\park, \lietal) is applied, compared to the baseline without CRT (\none). \lpips scores range from 0 (identical) to 1 (maximal different). \recon ranges from 0 (every pixel is identical) to $\sqrt{3}*512$ (every pixel is maximally different). \clipscore scores at or above 0.25 indicate semantic equivalence~\cite{brack2024ledits}.}
  
  
  \label{rec_ddpm_new_crt}
  \renewcommand{\arraystretch}{1.25}
  \setlength{\tabcolsep}{5pt} 
  \begin{tabular}{lccc}
    \toprule
    & \multicolumn{2}{c}{\textit{Offensive: Nudity}} \\
    \midrule
    CRT & \lpips & \recon & \clipscore \\
    \midrule
    \textit{\none} & 0.01 $\pm$ 0.01 & 12.19 $\pm$ 1.21 & 0.24 $\pm$ 0.01 \\
    \midrule
    \textit{\park} &  0.01 $\pm$ 0.00 & 12.21 $\pm$ 0.68  & 0.24 $\pm$ 0.01 \\
    \textit{\lietal} & 0.01 $\pm$ 0.01 & 12.20 $\pm$ 0.79 & 0.24 $\pm$ 0.00 \\
    \bottomrule
  \end{tabular}
\end{table}

%% file: tables/all_tables.tex
\begin{table*}[htbp]
  \centering
  \footnotesize 
  \caption{Reconstruction results when a CRT (\moderator, \mace, or \uce) is applied, compared to the baseline without CRT (\none) for LEDITS++. \lpips ranges from 0 (identical) to 1 (maximal different). \recon ranges from 0 (every pixel is identical) to $\sqrt{3}*512$ (every pixel is maximally different). \clipscore values at or above 0.25 indicate semantic equivalence~\cite{brack2024ledits}.}
  \label{rec_ledits}
  \setlength{\tabcolsep}{3pt} 
\resizebox{\textwidth}{!}{
  \begin{tabular}{lccc||ccc|ccc|ccc}
    \toprule
    & \multicolumn{3}{c||}{None} & \multicolumn{3}{c|}{\moderator} & \multicolumn{3}{c|}{\mace} & \multicolumn{3}{c}{\uce} \\
    Concepts & \lpips & \recon & \clipscore & \lpips & \recon & \clipscore & \lpips & \recon & \clipscore & \lpips & \recon & \clipscore \\
    \midrule
    \multicolumn{13}{c}{\textit{Celebrity}} \\
    \midrule
    \textit{Angelina Jolie} & 0.01 $\pm$ 0.01 & 16.72 $\pm$ 1.93 & 0.25 $\pm$ 0.01 & 0.01 $\pm$ 0.01 & 16.72 $\pm$ 1.54 & 0.25 $\pm$ 0.02 & 0.01 $\pm$ 0.00 & 16.72 $\pm$ 2.12 & 0.28 $\pm$ 0.03 & 0.01 $\pm$ 0.00 & 16.72 $\pm$ 2.89 & 0.25 $\pm$ 0.01 \\
    \textit{Taylor Swift} & 0.02 $\pm$ 0.01 & 21.32 $\pm$ 1.67 & 0.24 $\pm$ 0.01 & 0.02 $\pm$ 0.00 & 21.32 $\pm$ 1.78 & 0.24 $\pm$ 0.01 & 0.02 $\pm$ 0.01 & 21.32 $\pm$ 1.34 & 0.24 $\pm$ 0.01 & 0.02 $\pm$ 0.00 & 21.32 $\pm$ 1.45 & 0.24 $\pm$ 0.01 \\
    \textit{Brad Pitt} & 0.02 $\pm$ 0.00 & 26.45 $\pm$ 1.31 & 0.27 $\pm$ 0.01 & 0.02 $\pm$ 0.00 & 26.44 $\pm$ 1.21 & 0.27 $\pm$ 0.00 & 0.02 $\pm$ 0.01 & 26.44 $\pm$ 1.67 & 0.27 $\pm$ 0.01 & 0.02 $\pm$ 0.00 & 26.44 $\pm$ 1.23 & 0.27 $\pm$ 0.02 \\
    \textit{Elon Musk} & 0.02 $\pm$ 0.01 & 28.41 $\pm$ 1.48 & 0.28 $\pm$ 0.01 & 0.02 $\pm$ 0.00 & 28.41 $\pm$ 1.14 & 0.28 $\pm$ 0.01 & 0.02 $\pm$ 0.00 & 28.41 $\pm$ 1.58 & 0.28 $\pm$ 0.01 & 0.02 $\pm$ 0.00 & 28.41 $\pm$ 1.58 & 0.28 $\pm$ 0.00 \\
    \textit{Donald Trump} & 0.02 $\pm$ 0.01 & 32.05 $\pm$ 1.15 & 0.24 $\pm$ 0.01 & 0.02 $\pm$ 0.00 & 32.05 $\pm$ 1.13 & 0.24 $\pm$ 0.00 & 0.02 $\pm$ 0.01 & 32.05 $\pm$ 1.17 & 0.24 $\pm$ 0.00 & 0.02 $\pm$ 0.01 & 32.05 $\pm$ 1.12 & 0.24 $\pm$ 0.01 \\
    \textit{Joe Biden} & 0.02 $\pm$ 0.01 & 34.21 $\pm$ 1.27 & 0.26 $\pm$ 0.01 & 0.02 $\pm$ 0.01 & 34.21 $\pm$ 1.28 & 0.26 $\pm$ 0.01 & 0.02 $\pm$ 0.00 & 34.21 $\pm$ 1.39 & 0.26 $\pm$ 0.01 & 0.02 $\pm$ 0.01 & 34.21 $\pm$ 1.24 & 0.26 $\pm$ 0.01 \\
    \midrule
    \multicolumn{13}{c}{\textit{Offensive}} \\
    \midrule
    \textit{Nudity} & 0.01 $\pm$ 0.01 & 10.83 $\pm$ 1.84 & 0.24 $\pm$ 0.01 & 0.01 $\pm$ 0.01 & 10.83 $\pm$ 0.64 & 0.24 $\pm$ 0.01 & 0.01 $\pm$ 0.00 & 10.83 $\pm$ 0.75 & 0.24 $\pm$ 0.01 & 0.01 $\pm$ 0.00 & 10.83 $\pm$ 0.89 & 0.23 $\pm$ 0.01 \\
    \midrule
    \multicolumn{13}{c}{\textit{Copyrighted}} \\
    \midrule
    \textit{Grumpy cat} & 0.02 $\pm$ 0.01 & 31.67 $\pm$ 1.54 & 0.25 $\pm$ 0.01 & 0.02 $\pm$ 0.01 & 31.67 $\pm$ 0.87 & 0.25 $\pm$ 0.01 & 0.02 $\pm$ 0.00 & 31.67 $\pm$ 1.12 & 0.25 $\pm$ 0.01 & 0.02 $\pm$ 0.00 & 31.67 $\pm$ 1.26 & 0.25 $\pm$ 0.00\\
    \textit{R2D2} & 0.02 $\pm$ 0.01 & 36.47 $\pm$ 0.64 & 0.24 $\pm$ 0.01 & 0.02 $\pm$ 0.00 & 36.47 $\pm$ 0.69 & 0.25 $\pm$ 0.01 & 0.02 $\pm$ 0.00 & 36.47 $\pm$ 1.12 & 0.24 $\pm$ 0.01 & 0.02 $\pm$ 0.00 & 36.47 $\pm$ 0.89 & 0.25 $\pm$ 0.01  \\
    \bottomrule
  \end{tabular}}
\end{table*}